\documentclass[journal]{IEEEtran}
\usepackage{graphicx}
\usepackage{amsmath, amssymb}
\usepackage{caption, subcaption}
\usepackage{multirow}
\usepackage{soul,color}
\usepackage{cite}
\usepackage{hyperref}
\usepackage{tikz}
\usetikzlibrary{positioning,fit,shapes.geometric,calc,arrows.meta,backgrounds,shadows.blur,3d}
\usetikzlibrary{positioning, arrows.meta, shapes.geometric, fit, calc, decorations.pathreplacing,3d}
\usepackage{array}

\usepackage{algorithm}
\usepackage{algpseudocode}
\usepackage{pgfplots}
\pgfplotsset{compat=1.18}
\usepackage{adjustbox}
\usetikzlibrary{patterns}
\usepackage{booktabs}
\usepackage[table]{xcolor}
\usepackage{makecell}
\usepackage{threeparttable}
\usetikzlibrary{backgrounds}
\DeclareMathOperator*{\argmax}{arg\,max}
\begin{document}
\title{See the Future, Classify the Present: Stochastic Latent World Modeling for Hyperspectral Images}
\author{Muhammad Ahmad, Gemine~Vivone, Ronny~Hänsch, Ahmed Sohaib, Manuel~Mazzara, Danfeng~Hong
\thanks{M. Ahmad and A. Sohaib are with the Khanpur Institute of Technology, Khanpur 64100, Pakistan. (e-mail: mahmad00@gmail.com; engr.dr.ahmed.sohaib@gmail.com).}
\thanks{G. Vivone is with the National Research Council - Institute of Methodologies for Environmental Analysis (CNR-IMAA), 85050 Tito, Italy (e-mail: gemine.vivone@imaa.cnr.it).}
\thanks{R. Hänsch is with the Department of SAR Technology, German Aerospace Center (DLR), Germany. (e-mail: ronny.haensch@dlr.de).}
\thanks{M. Mazzara is with the Institute of Software Development and Engineering, Innopolis University, Innopolis, 420500, Russia. (e-mail: m.mazzara@innopolis.ru)}
\thanks{D. Hong is with the School of Automation, Southeast University, Nanjing, China. (e-mail: danfeng.hong@seu.edu.cn)}
}
\markboth{ArXiv}
{M.Ahmad \MakeLowercase{\textit{et al.}}}
\maketitle
\begin{abstract}
Hyperspectral image classification (HSIC) relies on static spatial-spectral representations learned from labeled samples, limiting its ability to model structured dependencies in the latent representation space. To address this limitation, this work proposes Stochastic Latent World Modeling (LWM), a trajectory-based representation-learning framework that enriches classification through stochastic latent-state prediction. Given a hyperspectral patch, LWM tokenizes and encodes spectral and spatial information separately using dedicated transformer branches, then fuses their outputs into a unified representation. This representation is further organized into an environment state that captures center, neighborhood, spectral, and global contextual information. A stochastic latent transition module then predicts multiple future latent states through Gaussian parameterization and reparameterization, with recurrent refinement used to model dependencies across successive states. The resulting latent trajectory is adaptively aggregated by a rollout planner to construct a planned representation, which is integrated with the base representation for final classification. By extending static spatial-spectral encoding with stochastic multi-step latent-state modeling, LWM enables the classifier to exploit structured dependencies in the representation space and provides richer features for HSIC. Across five benchmark datasets, LWM achieves overall accuracies ranging from 89.21\% to 99.52\%, demonstrating consistent classification performance under limited-label settings.
\end{abstract}
\IEEEpeerreviewmaketitle
\section{Introduction}

\IEEEPARstart{H}{yperspectral} image (HSI) classification (HSIC) aims to assign semantic labels to pixels by leveraging high-dimensional spectral signatures together with spatial context \cite{hong2026hyperspectral}. The rich spectral resolution of hyperspectral imagery enables fine-grained material discrimination \cite{11359013}; however, it also introduces significant challenges, including high dimensionality, strong inter-band correlation, and severe scarcity of labeled samples \cite{10529603}. Despite substantial advances, most existing HSIC approaches remain fundamentally based on static spatial-spectral representations \cite{hong2020joint}, where classification is performed from a single-stage encoding of local patches \cite{11090003}. This paradigm limits the ability of models to capture higher-order dependencies and structured relationships beyond immediate context, particularly in low-label regimes \cite{AHMAD2025130428}.

\textbf{Limitations of Existing Approaches.} Traditional HSIC methods, including hand-crafted feature pipelines and early deep models, represent each patch as a fixed embedding that is directly mapped to a class label \cite{Scheibenreif_2023_CVPR}. Convolutional neural networks (CNNs) and spatial-spectral hybrid models improve representation capacity through hierarchical feature extraction and local context aggregation \cite{LI2026104248, 9903062}. However, these models remain inherently feed-forward and static: the learned representation is a single point estimate in latent space \cite{11408874}. Such formulations do not explicitly encode relationships between latent states, nor do they model how representations could evolve under structured transformations \cite{11501276}. Consequently, they fail to exploit the continuity and smoothness inherent in hyperspectral data \cite{11482193, 11475204}.

\textbf{Attention-Based Context Modeling Without Structure.} Transformer-based models have been introduced to capture long-range dependencies across spectral bands and spatial locations \cite{ViTICLR2021dosovitskiy, li2024casformer, 11653218}. While multi-head self-attention (MHSA) improves global context aggregation \cite{vaswani2017attention}, it operates as a weighted averaging mechanism over tokens within a single forward pass \cite{11479589}. The resulting representation is still a one-step aggregation without explicit constraints on latent organization \cite{WASEEM2026104791, 11595698, li2026im2state}. In particular, attention does not impose transition structure or predictive consistency across representations, which limits its ability to enforce physically meaningful relationships such as spectral continuity and spatial coherence \cite{101145, li2026mambax, NTOUSKOS201310}.

\textbf{Self-Supervised Learning Without Latent Transitions.} Self-supervised approaches, including masked reconstruction \cite{Scheibenreif_2023_CVPR, hong2024spectralgpt}, contrastive learning \cite{9734031, 11361224}, and generative modeling \cite{10919111, 11222092}, aim to reduce dependence on labeled data by introducing auxiliary objectives \cite{10976430}. These methods encourage representations to preserve information or maintain invariances, but they largely operate at the level of reconstruction and similarity \cite{11226902}. They do not explicitly model dependencies between latent states, nor do they enforce how representations should transform under structured variations. As a result, they provide regularization but not structured predictive modeling \cite{10942442}.

\textbf{World Models and Their Limitations.} Recent advances in world models and latent dynamics learning \cite{zenodo1207631}, particularly in reinforcement learning and sequential modeling, aim to learn compact representations that evolve through predictive transitions \cite{MATSUO2022267, 11640768,hong2026foundation}. These methods typically rely on explicit temporal observations and learn state transition dynamics conditioned on actions or sequential inputs \cite{chung2026objectcentric, NEURIPS2025_8187faaf}. While effective in domains with well-defined temporal structure, such formulations are not directly applicable to hyperspectral imagery, which is inherently static and lacks explicit temporal supervision \cite{kingma2014autoencoding}.

Moreover, existing world models assume that transitions correspond to physically meaningful state changes observed in time \cite{NEURIPS2025_b63a24a1, he2026worldsimulator, 105281zenodo1207631, hafner2019planet, chung2015recurrent}, whereas hyperspectral data primarily exhibit structural correlations rather than temporal evolution. Applying standard world modeling frameworks to HSIC would therefore require artificial sequence construction or surrogate temporal assumptions, which may introduce bias and instability. In addition, world models focus on generative prediction in observation space, rather than enforcing structured organization within latent representations for discriminative tasks \cite{chung2014empirical, 1011453746449, 1011453450963}. Consequently, they do not directly address the need for latent predictive regularization in static spatial-spectral data \cite{NEURIPS2024_43ba0466, 101126sciadvaea4248}.

\textbf{Absence of Latent Predictive Structure.} A key limitation across these categories is the lack of latent predictive modeling tailored to HSIC. Hyperspectral data exhibit strong intrinsic structure, including smooth spectral variation across bands and spatial consistency across neighboring pixels. However, current models do not capture these properties as transitions in latent space. Without modeling such transitions, representations remain locally optimized but globally unstructured. This restricts the ability to generalize under limited supervision, as the model cannot reason beyond the immediate encoded observation.

\subsubsection{\textbf{Proposed Perspective}} To overcome these limitations, this work introduces stochastic latent world modeling (LWM), a trajectory-aware representation learning framework for HSIC. The central idea is to move beyond static encoding and instead learn representations that are structured through latent evolution. Rather than treating the representation of a patch as a fixed embedding, LWM models how this representation can transform under a learned stochastic process in latent space.

Unlike classical world models that rely on explicit temporal observations, the proposed formulation operates entirely in representation space. A latent environment state is first constructed from fused spatial-spectral features. A stochastic transition model then generates a sequence of latent states via multi-step rollouts, parameterized as Gaussian transitions with recurrent refinement. These rollouts represent plausible variations of the underlying representation, acting as a form of structured regularization rather than modeling physical time.

To extract decision-relevant information, a rollout planning mechanism adaptively aggregates latent states using learned importance weights. This produces a trajectory-aware representation that complements the base transformer encoding. The final prediction is obtained by combining the static and trajectory-based representations, enabling the model to leverage both immediate evidence and structured latent evolution.

The framework is trained with complementary objectives that jointly enforce reconstruction fidelity, latent dynamics consistency, stochastic regularization, and physically meaningful constraints. In particular, spectral and spatial dynamics losses explicitly model structured relationships within latent space, while stochastic rollout regularization enforces probabilistic coherence across predicted transitions. This unified formulation enables the model to learn representations that are not only discriminative but also structured and robust. The main contributions of this work are summarized as follows:

\begin{itemize}
    \item \textbf{Trajectory-Aware Representation Learning for HSIC:} We reformulate HSIC as a trajectory-aware latent representation problem, moving beyond static spatial-spectral encoding toward structured latent reasoning.

    \item \textbf{Stochastic Latent World Modeling:} We propose a stochastic latent transition framework that generates multi-step rollouts in representation space using Gaussian parameterization and recurrent refinement, enabling predictive structure without requiring temporal supervision.

    \item \textbf{Latent Rollout Planning Mechanism:} We introduce an adaptive rollout aggregation strategy that learns to weight latent states, producing a trajectory-aware representation for classification.

    \item \textbf{Structured Multi-Objective Learning:} We design a unified training objective combining masked reconstruction, spectral and spatial latent dynamics, stochastic rollout regularization, and physically motivated constraints, ensuring both discriminative power and structural consistency.

\end{itemize}

The remainder of this paper is organized as follows. Section~\ref{sec:method} introduces the proposed Stochastic Latent World Modeling (LWM) framework and its learning objectives. Section~\ref{sec:exp_setup} details the experimental setup. Sections~\ref{sec:branch}-\ref{sec:robustness} analyze latent rollout reasoning and robustness, while Section~\ref{sec:attention} examines model interpretability. Section~\ref{sec:ablation} presents ablation studies, and Section~\ref{sec:comparison} compares LWM with state-of-the-art methods. Section~\ref{sec:limitations} discusses limitations and failure cases. Finally, Section~\ref{sec:conclusion} concludes the paper and outlines future directions.

\begin{figure*}[!t]
\centering
\begin{adjustbox}{max totalsize={0.99\textwidth}{0.99\textheight},center}
\begin{tikzpicture}[
    scale=0.99,
    font=\scriptsize,
    >=Latex,
    panel/.style={draw=blue!45!black, rounded corners=7pt, thick, fill=white},
    subblue/.style={draw=blue!50!black, rounded corners=5pt, thick, fill=blue!3},
    subgreen/.style={draw=green!45!black, rounded corners=5pt, thick, fill=green!3},
    suborange/.style={draw=orange!70!black, rounded corners=5pt, thick, fill=orange!3},
    subpurple/.style={draw=purple!60!black, rounded corners=5pt, thick, fill=purple!3},
    subred/.style={draw=red!55!black, rounded corners=5pt, thick, fill=red!3},
    titleA/.style={font=\bfseries\footnotesize\color{purple!80!black}},
    smallbold/.style={font=\bfseries\footnotesize},
    smalltext/.style={font=\footnotesize},
    titleB/.style={font=\bfseries\color{blue!70!black}},
    titleC/.style={font=\bfseries\color{orange!90!black}},
    titleD/.style={font=\bfseries\color{red!75!black}},
    titleE/.style={font=\bfseries\color{purple!80!black}},
    arrow/.style={->, thick},
    thickarrow/.style={->, very thick},
    token/.style={draw=black!35, minimum width=0.23cm, minimum height=0.45cm},
    smallbox/.style={draw=black!50, rounded corners=2pt, fill=white, minimum width=.75cm, minimum height=.45cm, align=center},
    mhsa/.style={draw=orange!70!black, fill=orange!15, rounded corners=3pt, minimum width=.85cm, minimum height=.8cm, align=center},
    ln/.style={draw=cyan!70!black, fill=cyan!15, rounded corners=2pt, minimum width=.45cm, minimum height=.55cm, align=center},
    mlp/.style={draw=green!60!black, fill=green!15, rounded corners=3pt, minimum width=.65cm, minimum height=.55cm, align=center},
    latent/.style={circle, draw=orange!70!black, fill=orange!35, minimum size=.48cm},
    cls/.style={draw=purple!70!black, fill=purple!18, minimum width=.18cm, minimum height=.5cm},
    legendbox/.style={draw=black!35, rounded corners=4pt, fill=white}
]
\node[panel, minimum width=3.35cm, minimum height=6.6cm, anchor=north west] (A) at (0,0) {};
\node[titleA] at ($(A.north)+(0,-.25)$) {(A) Input \& Preprocessing};
\node[smallbold] at ($(A.north)+(0,-.55)$) {Hyperspectral Image};
\node[anchor=north] (cubeimg) at ($(A.north)+(0,-0.70)$)
{\includegraphics[width=2.45cm]{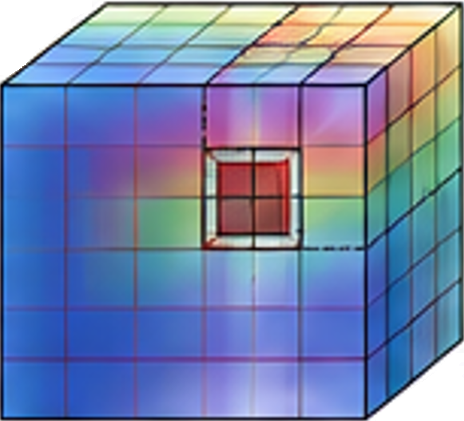}};
\node[subblue, minimum width=2.85cm, minimum height=3.25cm, anchor=north] (mask) at ($(A.north)+(0,-3.30)$) {};
\node[font=\bfseries \scriptsize, align=center] at ($(mask.north)+(0,-.25)$) {Masked Patch Modeling};
\node[anchor=west] at ($(mask.west)+(0.00,1.10)$) {$\bullet$ Random band masking};
\foreach \i/\c in {0/blue!60,1/cyan!40,2/orange!70,3/red!30,4/yellow!70,5/blue!30,6/cyan!50,7/orange!70}{
  \fill[draw=black!30, fill=\c] ($(mask.west)+(.38+\i*.25,.55)$) rectangle ++(.2,.28);
}
\node[anchor=west] at ($(mask.west)+(0.00,0.20)$) {$\bullet$ Random pixel masking};
\foreach \i in {0,...,4}{
 \foreach \j in {0,...,4}{
  \pgfmathtruncatemacro{\m}{mod(\i+\j,3)}
  \ifnum\m=0 \fill[black!70] ($(mask.west)+(.25+\i*.14,-.65+\j*.14)$) rectangle ++(.12,.12);
  \else \fill[gray!20] ($(mask.west)+(.25+\i*.14,-.65+\j*.14)$) rectangle ++(.12,.12);\fi
  \draw[black!45] ($(mask.west)+(.25+\i*.14,-.65+\j*.14)$) rectangle ++(.12,.12);
 }}
\draw[arrow] ($(mask.west)+(1.00,-.30)$)--($(mask.west)+(1.70,-.30)$);
\foreach \i in {0,...,4}{
 \foreach \j in {0,...,4}{
  \pgfmathtruncatemacro{\m}{mod(2*\i+\j,4)}
  \ifnum\m=0 \fill[white] ($(mask.west)+(1.75+\i*.14,-.65+\j*.14)$) rectangle ++(.12,.12);
  \else \fill[gray!40] ($(mask.west)+(1.75+\i*.14,-.65+\j*.14)$) rectangle ++(.12,.12);\fi
  \draw[black!45] ($(mask.west)+(1.75+\i*.14,-.65+\j*.14)$) rectangle ++(.12,.12);
 }}
\node[anchor=west] at ($(mask.west)+(.00,-0.90)$) {$\bullet$ Mirror padding};
\foreach \i in {0,...,4}{
 \foreach \j in {0,...,4}{
  \draw[black!35] ($(mask.west)+(2.05+\i*.14,-1.50+\j*.14)$) rectangle ++(.14,.14);
 }}
\draw[green!60!black] ($(mask.west)+(2.20,-1.36)$) rectangle ++(.42,.42);
\node[panel, minimum width=10.0cm, minimum height=9.0cm, anchor=north west] (B) at (3.45,0) {};
\node[titleB] at ($(B.north)+(0,-.25)$) {(B) Dual Tokenization \& Modality Encoding};
\node[subblue, minimum width=6.7cm, minimum height=3.50cm, anchor=north west] (spec) at ($(B.north west)+(.10,-.65)$) {};
\node[titleB, anchor=west] at ($(spec.north west)+(.20,-.28)$) {1) Spectral Tokenizer {\normalfont(per band token)}};
\node[font=\bfseries] at ($(spec.north)+(-1.6,-.65)$) {$\mathbf{S}^0\in\mathbb{R}^{B \times C\times D}$};
\node[align=center] at ($(spec.west)+(0.60,.40)$) {Each band\\as a token};
\foreach \i/\c in {0/blue!40,1/cyan!40,2/green!40,3/orange!55,4/red!40}{
  \draw[fill=\c, draw=black!25] ($(spec.west)+(.05+\i*.28,-.70+\i*.02)$) -- ++(.12,.75) -- ++(.22,-.15) -- ++(-.12,-.75) -- cycle;
}
\node at ($(spec.west)+(0.70,-1.10)$) {Band 1};
\draw[arrow] ($(spec.west)+(1.45,-.35)$)--($(spec.west)+(1.80,-.35)$);
\node[align=center] at ($(spec.west)+(2.10,.50)$) {Patch\\flattening};
\foreach \i in {0,...,6}{
  \draw[fill=blue!20, draw=blue!40!black] ($(spec.west)+(1.80+\i*.04,-0.90+\i*.04)$) rectangle ++(.45,.85);
}
\node at ($(spec.west)+(2.10,-1.10)$) {Band C};
\draw[arrow] ($(spec.west)+(2.50,-.35)$)--($(spec.west)+(2.90,-.35)$);
\foreach \j in {0,...,4}{
  \fill[draw=black!30, fill=red!50] ($(spec.west)+(2.90,-.90+\j*.23)$) rectangle ++(.2,.18);
}
\draw[arrow] ($(spec.west)+(3.10,-.35)$)--($(spec.west)+(3.50,-.35)$);
\node[subblue, minimum width=3.20cm, minimum height=2.45cm, anchor=east] (strans) at ($(spec.east)+(-.05,-0.22)$) {};
\node[font=\scriptsize, align=center]at ($(strans.north)+(0,-.20)$){Spectral Former (depth $L_{\lambda}$)};
\node[ln]   (ln1)  at ($(strans.west)+(.35,0.40)$) {LN};
\node[mhsa] (mhsa1) at ($(strans.west)+(1.5,0.40)$) {MHSA\\{\tiny band-wise}};
\node (add1) at ($(strans.west)+(2.50,0.40)$) {\Large $\oplus$};
\node[ln]   (ln2) at ($(strans.west)+(2.50,-0.60)$) {LN};
\node[mlp]  (mlp1) at ($(strans.west)+(1.40,-0.60)$) {MLP};
\node (add2) at ($(strans.west)+(0.50,-0.60)$) {\Large $\oplus$};
\node at ($(strans.south)+(1.3,.25)$) {$\times L_{\lambda}$}; 
\draw[->, thick] (ln1.east) -- (mhsa1.west);
\draw[arrow,thick] ($(mhsa1.east)$)--($(add1.west)+(0.20,-.00)$);
\draw[->, thick] (ln1.north) -- ++(0,0.20) -| ([yshift=-0.20cm]add1.north);
\draw[->, thick] ([yshift=0.15cm]add1.south) -- (ln2.north);
\draw[->, thick] (ln2.west) -- (mlp1.east);
\draw[->, thick] (mlp1.west) -- ([xshift=-0.20cm]add2.east);
\draw[->, thick] (ln2.south) |- ++(0,-0.20) -| ([yshift=0.18cm]add2.south);
\node[subgreen, minimum width=6.7cm, minimum height=3.50cm, anchor=north west] (spat) at ($(B.north west)+(0.08,-5.05)$) {};
\node[titleB, color=green!45!black, anchor=west] at ($(spat.north west)+(.25,-.28)$) {2) Spatial Tokenizer {\normalfont(per pixel token)}};
\node[font=\bfseries] at ($(spat.north)+(-1.6,-.65)$) {$\mathbf{A}^0\in\mathbb{R}^{B \times P^2 \times D}$};
\node[align=center] at ($(spat.west)+(0.60,0.10)$) {Each pixel\\as a token};
\foreach \i in {0,...,4}{
 \foreach \j in {0,...,4}{
   \fill[draw=black!35, fill=gray!10] ($(spat.west)+(.15+\i*.18,-1.15+\j*.18)$) rectangle ++(.18,.18);
 }}
\fill[red!65] ($(spat.west)+(0.50,-.80)$) rectangle ++(.18,.18);
\draw[arrow] ($(spat.west)+(1.20,-0.70)$)--($(spat.west)+(1.65,-0.70)$);
\foreach \i in {0,...,7}{
  \draw[fill=blue!12, draw=blue!40!black] ($(spat.west)+(1.70+\i*.04,-1.10+\i*.04)$) rectangle ++(.55,.95);
}
\node[align=center] at ($(spat.west)+(2.00,-1.35)$) {Full spectrum at each pixel};
\draw[arrow] ($(spat.west)+(2.60,-0.70)$)--($(spat.west)+(3.10,-0.70)$);
\node[subgreen, minimum width=3.20cm, minimum height=2.45cm, anchor=east] (atrans) at ($(spat.east)+(-.05,-0.22)$) {};
\node[font=\scriptsize, align=center] at ($(atrans.north)+(0,-.22)$) {Spatial Former (depth $L_s$)};
\node[ln] (ln1) at ($(atrans.west)+(.35,0.40)$) {LN};
\node[mhsa] (mhsa1) at ($(atrans.west)+(1.5,0.40)$) {MHSA\\{\tiny spatial}};
\node (add1) at ($(atrans.west)+(2.50,0.40)$) {\Large $\oplus$};
\node[ln]   (ln2) at ($(atrans.west)+(2.50,-0.60)$) {LN};
\node[mlp]  (mlp1) at ($(atrans.west)+(1.40,-0.60)$) {MLP};
\node (add2) at ($(atrans.west)+(0.50,-0.60)$) {\Large $\oplus$};
\node at ($(atrans.south)+(1.3,.25)$) {$\times L_s$}; 
\draw[->, thick] (ln1.east) -- (mhsa1.west);
\draw[arrow,thick] ($(mhsa1.east)$)--($(add1.west)+(0.20,-.00)$);
\draw[->, thick] (ln1.north) -- ++(0,0.20) -| ([yshift=-0.20cm]add1.north);
\draw[->, thick] ([yshift=0.15cm]add1.south) -- (ln2.north);
\draw[->, thick] (ln2.west) -- (mlp1.east);
\draw[->, thick] (mlp1.west) -- ([xshift=-0.20cm]add2.east);
\draw[->, thick] (ln2.south) |- ++(0,-0.20) -| ([yshift=0.18cm]add2.south);
\node[subpurple, minimum width=2.90cm, minimum height=5.00cm, anchor=north west] (fusion) at ($(B.north west)+(7.05,-2.30)$) {};
\node[font=\scriptsize\color{purple!70!black}, align=center] at ($(fusion.north)+(0,-.35)$) {3) Fusion Transformer\\(depth $L_f$)};
\node[smallbox, fill=purple!20, minimum width=1.2cm] at ($(fusion.west)+(.80,1.20)$) {CLS Token};
\node[smallbox, dashed, minimum width=1.2cm] at ($(fusion.west)+(0.80,0.60)$) {Embedding};
\node[smallbox, fill=blue!15, minimum width=1.2cm] at ($(fusion.west)+(.80,-0.30)$) {Spec Tokens};
\node[smallbox, dashed, minimum width=1.2cm] at ($(fusion.west)+(.80,-0.90)$) {Embedding};
\node[smallbox, fill=green!15, minimum width=1.2cm] at ($(fusion.west)+(.80,-1.70)$) {Spat Tokens};
\node[smallbox, dashed, minimum width=1.2cm] at ($(fusion.west)+(.80,-2.25)$) {Embedding};
\node[draw=black!50, rounded corners=5pt, minimum width=1.30cm, minimum height=4.00cm] (finner) at ($(fusion.east)+(-0.75,-0.40)$) {};
\node[ln] (fln1) at ($(finner.north)+(0,-.35)$) {LN};
\node[mhsa, minimum height=.75cm] (fmhsa) at ($(finner.north)+(0,-1.15)$) {MHSA\\{\tiny global}};
\node (fadd1) at ($(finner.north)+(0,-1.8)$) {\Large $\oplus$};
\node[ln] (fln2) at ($(finner.north)+(0,-2.35)$) {LN};
\node[mlp] (fmlp) at ($(finner.north)+(0,-3.10)$) {MLP};
\node (fadd2) at ($(finner.north)+(0,-3.65)$) {\Large $\oplus$};
\node at ($(fusion.south)+(1.00,.25)$) {$\times L_f$};
\draw[->, thick] (fln1.west) -- ++(-0.25,0) |- ([xshift=-0.25cm]fadd1.west) -- (fadd1.west)-- ++(0.20,0);
\draw[->, thick] (fadd1.east) -- ++(0.25,0) |- ([xshift=0.25cm]fadd2.east) -- ([xshift=-0.20cm]fadd2.east);
\draw[arrow] (strans.east)--(fusion.west|-strans.east);
\draw[arrow] (atrans.east)--(fusion.west|-atrans.east);
\node[font=\scriptsize] (sceneText) at ($(fusion.south)+(0,-1.50)$) {Scene (CLS) Representation};
\foreach \i in {0,...,6}{
  \node[cls] (cls\i) at ($(fusion.south)+(-.75+\i*.25,-1.00)$) {};
}
\draw[arrow] (fusion.south) -- ([yshift=0.5cm]sceneText.north);
\node[suborange, minimum width=2.50cm, minimum height=1.35cm, anchor=south] (base) at ($(fusion.north)+(0.0,.4)$) {};
\node[font=\bfseries] at ($(base.north)+(0,-.25)$) {Base Classifier};
\node[smallbox, minimum width=1.15cm] at ($(base.center)+(0,-.15)$) {MLP Head}; %
\node at ($(base.south)+(0,.2)$) {$\ell_{base}$}; %
\draw[arrow] ($(fusion.north)+(0.0,0)$)--(base.south);
\node[panel, minimum width=7.15cm, minimum height=9.0cm, anchor=north west] (C) at (13.70,0) {};
\node[titleC] at ($(C.north)+(0,-.25)$) {(C) World Model \& Rollout};
\node[suborange, minimum width=3.30cm, minimum height=5.1cm, anchor=north west] (env) at ($(C.north west)+(0.10,-1.1)$) {};
\node[font=\scriptsize \color{orange!90!black}, align=center] at ($(env.north)+(0,-.28)$) {4) Environment State Encoder};
\node[anchor=west] at ($(env.west)+(0.0,2.00)$) {Center State};
\foreach \i in {0,...,5}{
  \node[token, fill=orange!20, minimum height=.22cm] at ($(env.west)+(.35+\i*.16,1.75)$) {};
}
\coordinate (centerTokens) at ($(env.west)+(0.75,1.75)$);
\node[anchor=west, align=left] at ($(env.west)+(0.0,0.80)$) {Neighbor States\\{\tiny (1-ring)}};
\foreach \i in {0,...,7}{
  \node[token, fill=orange!20, minimum height=.22cm] at ($(env.west)+(.35+\i*.14,.35)$) {};
}
\coordinate (neighborTokens) at ($(env.west)+(0.85,.35)$);
\node[anchor=west, align=left] at ($(env.west)+(0.0,-.45)$) {Spectral Summary\\{\tiny (mean over Bands)}};
\foreach \i in {0,...,6}{
  \node[token, fill=blue!20, minimum height=.22cm] at ($(env.west)+(.35+\i*.14,-.85)$) {};
}
\coordinate (spectralTokens) at ($(env.west)+(0.78,-.85)$);
\node[anchor=west, align=left] at ($(env.west)+(0.0,-1.45)$) {\footnotesize Scene State\\(CLS)};
\foreach \i in {0,...,6}{
  \node[token, fill=purple!20, minimum height=.22cm] at ($(env.west)+(.35+\i*.14,-1.85)$) {};
}
\coordinate (sceneTokens) at ($(env.west)+(0.78,-1.70)$);
\node[suborange,minimum width=.95cm,minimum height=2.8cm,fill=orange!13,align=center] (mlpenc) at ($(env.east)+(-.65,0.15)$) {MLP\\Encoder};
\draw[arrow] ([xshift=0.50cm]centerTokens.east) -- ([yshift=0.70cm]mlpenc.west);
\draw[arrow] ([xshift=0.60cm]neighborTokens.east) -- ([xshift=-0.05cm]mlpenc.west);
\draw[arrow] ([xshift=0.70cm]spectralTokens.east) -- ([yshift=-0.65cm]mlpenc.west);
\draw[arrow] ([xshift=0.60cm]sceneTokens.east) -- ([yshift=-0.95cm]mlpenc.west);
\node[latent] (z0enc) at ($(env.south)+(1.03,.55)$) {$z_0$};
\node at ($(env.south)+(0,.45)$) {$z_0\in\mathbb{R}^{B \times D}$};
\draw[arrow] (mlpenc.south)--(z0enc.north);
\node[suborange, minimum width=3.60cm, minimum height=2.25cm, anchor=north east] (roll) at ($(C.north east)+(-.05,-1.15)$) {};
\node[font=\footnotesize\color{orange!90!black}] at ($(roll.north)+(0,-.28)$) {5) Stochastic Latent Rollout};
\node[smallbox, dashed, minimum width=1.55cm, minimum height=.65cm]
at ($(roll.north)+(0,-.85)$)
{Stochastic Transition\\$G_\theta,R_\theta$};
\foreach \i/\lab in {0/z_0,1/z_1,2/z_2,4/z_T}{
  \node[latent,minimum size=.30cm,inner sep=1.2pt,font=\tiny] (rz\i) at ($(roll.south)+(-1.50+\i*.75,.45)$) {$\lab$};}
\draw[arrow] (rz0)--(rz1);
\draw[arrow] (rz1)--(rz2);
\draw[dotted, thick] (rz2)--(rz4);
\node at ($(rz2)!.5!(rz4)+(0,.1)$) {$\cdots$};
\node[subgreen, minimum width=3.60cm, minimum height=1.25cm, anchor=north east] (plan) at ($(C.north east)+(-.05,-4.2)$) {};
\node[font=\footnotesize\color{green!40!black}] at ($(plan.north)+(0,-.28)$) {6) Rollout Planner};
\foreach \i in {0,...,4}{
  \draw[fill=green!50!black!40] ($(plan.west)+(.55+\i*.48,-.25)$) rectangle ++(.08,{.15+.12*mod(\i,3)});
  \node at ($(plan.west)+(.6+\i*.48,-0.5)$) {$w_\i$};
}
\node at ($(plan.east)+(-.35,-0.25)$) {$\cdots$};
\node[subgreen, minimum width=2.55cm, minimum height=.9cm, anchor=north] (pstate) at ($(plan.south)+(0,-0.7)$) {};
\node[align=center] at (pstate.center) {Planned State\\$z_{plan}=\sum_t w_t z_t$};
\node[suborange, minimum width=2.8cm, minimum height=1.0cm, anchor=north] (traj) at ($(pstate.south)+(0,-.72)$) {};
\node[font=\footnotesize, align=center] at (traj.center) {7) Trajectory Classifier\\{\normalfont$\ell_{traj}$}};
\draw[arrow] (roll.south)--(plan.north);
\draw[arrow] (plan.south)--(pstate.north);
\draw[arrow] (pstate.south)--(traj.north);
\draw[arrow,blue] (z0enc.east) -| ($(rz0.west)+(-0.30,0)$) -- (rz0.west);
\node[subred, minimum width=3.20cm, minimum height=1.15cm, anchor=south] (final) at ($(C.south)+(-1.80,0.16)$) {};
\node[font=\footnotesize\color{red!70!black}] at ($(final.north)+(0,-.25)$) {Final Prediction};
\node at ($(final.south)+(0,.40)$) {
    $\begin{aligned}
    \boldsymbol{\ell}
    &= \mathbf{G}\odot\boldsymbol{\ell}_{\mathrm{base}} + \\
    &\quad \left(\mathbf{1}_{B\times K}-\mathbf{G}\right)
    \odot \boldsymbol{\ell}_{\mathrm{traj}}
    \end{aligned}$
};
\draw[arrow] (traj.west)--(final.east);
\draw[arrow](z0enc.south)-- ++(0,-0.85) -| (final.north);
\node[draw=black!90,dashed,rounded corners=5pt,thick,minimum width=17.90cm,minimum height=4.00cm,anchor=north west] (D) at (0,-9.20) {};
\node[titleD] at ($(D.north)+(0,-0.25)$) {(D) Training Objectives (Auxiliary Supervision)};
\def\cardw{3.35cm}
\def\cardh{3.55cm}
\def\cardy{-11.40cm}
\def\xA{1.80}
\def\xB{5.40}
\def\xC{9.00}
\def\xD{12.65}
\def\xE{16.30}
\node[draw=blue!60!black,rounded corners=5pt,fill=blue!3,minimum width=\cardw,minimum height=\cardh,align=center] (recon) at (\xA,\cardy) {};
\node[font=\footnotesize\bfseries\color{blue!70!black}] at ($(recon.north)+(0,-.20)$) {Masked Reconstruction};
\node[align=center, font=\scriptsize] at ($(recon.north)+(0,-.62)$) {Reconstruct masked\\bands \& pixels};
\node[anchor=west, font=\scriptsize] at ($(recon.west)+(.00,0.60)$)
{$\bullet$ Spectral Recon};
\begin{scope}
  \coordinate (sr0) at ($(recon.west)+(.20,-0.30)$);
  \foreach \i/\c in {
    0/blue!55,1/cyan!45,2/orange!55,3/red!35,4/yellow!55,
    5/blue!35,6/cyan!30,7/orange!40,8/blue!20,9/cyan!20
  }{
    \fill[draw=black!25, fill=\c] ($(sr0)+(\i*.13,0.4)$) rectangle ++(.12,.14);
  }
  \draw[-{Latex[length=1.3mm]}, thin] ($(recon.west)+(1.70,0.20)$) -- ++(.40,0);
  \node[font=\scriptsize] at ($(recon.east)+(-.65,0.20)$) {$\mathcal{L}_{spec\text{-}rec}$};
\end{scope}
\node[anchor=west, font=\scriptsize] at ($(recon.west)+(.00,-0.50)$)
{$\bullet$ Spatial Recon};
\begin{scope}
  \coordinate (sg0) at ($(recon.west)+(.18,-1.50)$);
  \foreach \i in {0,...,6}{
    \foreach \j in {0,...,6}{
      \pgfmathtruncatemacro{\m}{mod(\i+\j,3)}
      \ifnum\m=0
        \fill[black!70] ($(sg0)+(\i*.10,\j*.10)$) rectangle ++(.09,.09);
      \else
        \fill[gray!20] ($(sg0)+(\i*.10,\j*.10)$) rectangle ++(.09,.09);
      \fi
      \draw[black!35] ($(sg0)+(\i*.10,\j*.10)$) rectangle ++(.09,.09);
    }
  }
  \draw[-{Latex[length=1.3mm]}, thin] ($(recon.west)+(1.70,-1.20)$) -- ++(.40,0);
  \node[font=\scriptsize] at ($(recon.east)+(-.65,-1.20)$) {$\mathcal{L}_{spat\text{-}rec}$};
\end{scope}
\node[draw=green!60!black,rounded corners=5pt,fill=green!3,minimum width=\cardw,minimum height=\cardh,align=center] (dyn) at (\xB,\cardy) {};
\node[font=\footnotesize\bfseries\color{green!50!black}] at ($(dyn.north)+(0,-.20)$) {Local Dynamics};
\node[align=center, font=\scriptsize] at ($(dyn.north)+(0,-.62)$) {Predict next band \&\\neighbor states};
\node[anchor=west, font=\scriptsize] at ($(dyn.west)+(0.00,0.60)$) {$\bullet$ Spectral Dynamics};
\begin{scope}
  \node[circle, draw=green!60!black, fill=green!25, minimum size=3mm] (d1) at ($(dyn.west)+(.25,-.02)$) {};
  \node[circle, draw=green!60!black, fill=green!25, minimum size=3mm] (d2) at ($(dyn.west)+(1.25,-.02)$) {};
  \node[circle, draw=black!50, fill=white, minimum size=3mm] (d3) at ($(dyn.west)+(2.00,-.02)$) {};
  \node[font=\scriptsize] at ($(dyn.west)+(.35,-.35)$) {$\hat{z}_{\lambda-1}$};
  \node[font=\scriptsize] at ($(dyn.west)+(1.25,-.35)$) {$z_{\lambda-1}$};
  \node[font=\scriptsize] at ($(dyn.west)+(2.15,-.35)$) {};
  \draw[-{Latex[length=1.3mm]}, thick] (d1.east) -- (d2.west);
  \draw[-{Latex[length=1.3mm]}, thick] (d2.east) -- (d3.west);
  \node[font=\scriptsize] at ($(dyn.east)+(-.55,-.02)$) {$\mathcal{L}_{spec\text{-}dyn}$};
  \node[font=\scriptsize] at ($(dyn.west)+(1.50,0.20)$) {vs};
\end{scope}
\node[anchor=west, font=\scriptsize] at ($(dyn.west)+(.00,-0.70)$)
{$\bullet$ Spatial Dynamics};
\begin{scope}
  \coordinate (dg0) at ($(dyn.west)+(.20,-1.02)$);
  \foreach \i in {0,...,5}{
    \foreach \j in {0,...,5}{
      \fill[gray!10] ($(dg0)+(\i*.10,\j*-.10)$) rectangle ++(.09,.09);
      \draw[black!35] ($(dg0)+(\i*.10,\j*-.10)$) rectangle ++(.09,.09);
    }
  }
  \fill[green!40] ($(dg0)+(.20,-.20)$) rectangle ++(.09,.09);
  \fill[red!55]   ($(dg0)+(.30,-.30)$) rectangle ++(.09,.09);
  \draw[-{Latex[length=1.3mm]}, thick] ($(dyn.west)+(0.90,-1.20)$) -- ++(.45,0);
  \coordinate (dg1) at ($(dyn.west)+(1.45,-1.50)$);
  \foreach \i in {0,...,5}{
    \foreach \j in {0,...,5}{
      \fill[green!10] ($(dg1)+(\i*.10,\j*.10)$) rectangle ++(.09,.09);
      \draw[black!35] ($(dg1)+(\i*.10,\j*.10)$) rectangle ++(.09,.09);
    }
  }
  \fill[green!40] ($(dg1)+(.20,.30)$) rectangle ++(.09,.09);
  \fill[red!55]   ($(dg1)+(.30,.20)$) rectangle ++(.09,.09);
  \node[font=\scriptsize] at ($(dyn.east)+(-.55,-1.20)$) {$\mathcal{L}_{spat\text{-}dyn}$};
  \node[font=\scriptsize] at ($(dyn.west)+(1.10,-1.05)$) {vs};
\end{scope}
\node[draw=orange!70!black,rounded corners=5pt,fill=orange!3,minimum width=\cardw,minimum height=\cardh,align=center] (latentloss) at (\xC,\cardy) {};
\node[font=\footnotesize\bfseries\color{orange!80!black}] at ($(latentloss.north)+(0,-.20)$) {Latent Rollout Loss};
\node[align=center, font=\scriptsize] at ($(latentloss.north)+(0,-.62)$) {Match predicted distribution\\to future targets};
\foreach \i/\lab in {0/z_0,1/z_1,2/z_2,4/z_T}{
  \node[circle,draw=orange!70!black,fill=orange!30,minimum size=2.0mm,inner sep=0.6pt,font=\tiny]
  (lz\i) at ($(latentloss.west)+(.25+\i*.68,0.50)$) {$\lab$};
}
\draw[-{Latex[length=0.9mm]}, semithick] ([xshift=0.4mm]lz0.east) -- ([xshift=-0.4mm]lz1.west);
\draw[-{Latex[length=0.9mm]}, semithick] ([xshift=0.4mm]lz1.east) -- ([xshift=-0.4mm]lz2.west);
\draw[densely dotted, semithick] ([xshift=0.4mm]lz2.east) -- ($(latentloss.east)+(-1.05,.50)$);
\draw[-{Latex[length=0.9mm]}, semithick] ($(latentloss.east)+(-1.05,.50)$) -- ([xshift=-0.4mm]lz4.west);
\foreach \i/\lab in {0/z,1/z,2/z,4/z}{
  \node[circle, draw=gray!60, dashed, fill=white, minimum size=3.0mm,inner sep=0.6pt, font=\tiny]
  (lb\i) at ($(latentloss.west)+(.25+\i*.68,-.75)$) {$\lab$};
}
\draw[{Latex[length=1.1mm]}-{Latex[length=1.1mm]}, thin] (lz0.south) -- (lb0.north);
\draw[{Latex[length=1.1mm]}-{Latex[length=1.1mm]}, thin] (lz1.south) -- (lb1.north);
\draw[{Latex[length=1.1mm]}-{Latex[length=1.1mm]}, thin] (lz2.south) -- (lb2.north);
\draw[{Latex[length=1.1mm]}-{Latex[length=1.1mm]}, thin] (lz4.south) -- (lb4.north);
\node[font=\scriptsize] at ($(latentloss.west)+(1.00,-1.18)$) {$\mathcal{L}_{env}$\,(\text{\tiny NLL/SMOOTH})};
\node[font=\large] at ($(latentloss.center)+(0.60,-1.18)$) {$+$};
\node[font=\scriptsize] at ($(latentloss.east)+(-0.40,-1.18)$) {$\mathcal{L}_{KL}$};
\node[draw=yellow!70!black,rounded corners=5pt,fill=yellow!3,minimum width=\cardw,minimum height=\cardh,align=center] (phys) at (\xD,\cardy) {};
\node[font=\footnotesize\bfseries\color{orange!90!black}] at ($(phys.north)+(0,-.20)$) {Physical Regularization};
\node[anchor=west, font=\scriptsize] at ($(phys.west)+(0.00,1.20)$) {$\bullet$ Spectral Angle Mapper};
\begin{scope}
  \coordinate (p0) at ($(phys.west)+(.20,0.00)$);
  \coordinate (pr)    at ($(p0)+(0,.62)$);          
  \coordinate (pf)    at ($(p0)+(35:0.78)$);        
  \coordinate (prbar) at ($(p0)+(0:0.82)$);         
  \draw[-{Latex[length=1.0mm]}, thin] (p0) -- (pr);
  \draw[-{Latex[length=1.0mm]}, thin] (p0) -- (pf);
  \draw[-{Latex[length=1.0mm]}, thin] (p0) -- (prbar);
  \node[font=\scriptsize, anchor=south] at ($(pr)+(0,.06)$) {$\hat{\mathbf{r}}$};
  \node[font=\scriptsize, anchor=south west] at ($(pf)+(.02,.00)$) {$\hat{\mathbf{f}}$};
  \node[font=\scriptsize, anchor=west] at ($(prbar)+(.02,.00)$) {$\mathbf{r}$};
  \draw[thin] ($(p0)+(90:0.18)$) arc[start angle=90, end angle=35, radius=0.18];
  \node[font=\tiny] at ($(p0)+(58:0.30)$) {$\theta$};
  \draw[thin] ($(p0)+(35:0.26)$) arc[start angle=35, end angle=0, radius=0.26];
  \node[font=\tiny] at ($(p0)+(18:0.40)$) {$\phi$};
  \draw[-{Latex[length=1.2mm]}, thin] ($(phys.west)+(1.68,.02)$) -- ++(.50,0);
  \node[font=\scriptsize] at ($(phys.east)+(-.50,.04)$) {$\mathcal{L}_{SAM}$};
\end{scope}
\node[anchor=west, font=\scriptsize] at ($(phys.west)+(.00,-.22)$) {$\bullet$ Spectral Smoothness};
\begin{scope}
      \draw[blue!70!black, thick]
      ($(phys.west)+(.20,-.78)$) --
      ($(phys.west)+(.36,-.72)$) --
      ($(phys.west)+(.50,-.57)$) --
      ($(phys.west)+(.66,-.52)$) --
      ($(phys.west)+(.82,-.60)$) --
      ($(phys.west)+(.96,-.54)$) --
      ($(phys.west)+(1.10,-.62)$);
    \draw[black!30] ($(phys.west)+(.10,-.86)$) -- ($(phys.west)+(.10,-.48)$);
    \draw[black!30] ($(phys.west)+(.10,-.86)$) -- ($(phys.west)+(1.10,-.86)$);
    \draw[-{Latex[length=1.2mm]}, thin] ($(phys.west)+(1.55,-.78)$) -- ++(.50,0);
    \node[font=\scriptsize] at ($(phys.east)+(-.50,-.78)$) {$\mathcal{L}_{smooth}$};
\end{scope}
\node[anchor=west, font=\scriptsize] at ($(phys.west)+(.00,-1.10)$) {$\bullet$ Spatial Consistency};
\begin{scope}
  \coordinate (ps0) at ($(phys.west)+(.50,-1.70)$);
  \coordinate (ps1) at ($(phys.west)+(1.50,-1.70)$);
  \foreach \i in {0,...,4}{
    \foreach \j in {0,...,4}{
      \pgfmathtruncatemacro{\m}{mod(\i+\j,3)}
      \ifnum\m=0
        \fill[blue!60] ($(ps0)+(\i*.09,\j*.09)$) rectangle ++(.085,.085);
      \else
        \fill[blue!15] ($(ps0)+(\i*.09,\j*.09)$) rectangle ++(.085,.085);
      \fi
      \draw[black!25] ($(ps0)+(\i*.09,\j*.09)$) rectangle ++(.085,.085);
    }
  }
  \foreach \i in {0,...,4}{
    \foreach \j in {0,...,4}{
      \pgfmathtruncatemacro{\m}{mod(2*\i+\j,4)}
      \ifnum\m=0
        \fill[blue!60] ($(ps1)+(\i*.09,\j*.09)$) rectangle ++(.085,.085);
      \else
        \fill[blue!15] ($(ps1)+(\i*.09,\j*.09)$) rectangle ++(.085,.085);
      \fi
      \draw[black!25] ($(ps1)+(\i*.09,\j*.09)$) rectangle ++(.085,.085);
    }
  }
  \draw[-{Latex[length=1.2mm]}, thick] ($(phys.west)+(1.05,-1.50)$) -- ++(.40,0);
  \node[font=\scriptsize] at ($(phys.east)+(-.50,-1.50)$) {$\mathcal{L}_{spatial}$};
\end{scope}
\node[draw=pink!65!black,rounded corners=5pt,fill=pink!3,minimum width=\cardw,minimum height=\cardh,align=center] (trajloss) at (\xE,\cardy) {};
\node[font=\footnotesize\bfseries\color{pink!80!black}] at ($(trajloss.north)+(0,-.20)$) {Trajectory Consistency};
\node[align=center, font=\scriptsize] at ($(trajloss.north)+(0,-.62)$) {Align base \& trajectory\\predictions};
\begin{scope}
  \coordinate (tbL) at ($(trajloss.west)+(.55,-.10)$);
  \coordinate (tbR) at ($(trajloss.west)+(2.15,-.10)$);
  \foreach \x/\h in {0/.45,1/.62,2/.40,3/.58}{
    \fill[purple!50] ($(tbL)+(.12+\x*.18,0)$) rectangle ++(.10,\h);
    \draw[black!30] ($(tbL)+(.12+\x*.18,0)$) rectangle ++(.10,\h);
  }
  \foreach \x/\h in {0/.35,1/.72,2/.44,3/.66}{
    \fill[blue!55] ($(tbR)+(.12+\x*.18,0)$) rectangle ++(.10,\h);
    \draw[black!30] ($(tbR)+(.12+\x*.18,0)$) rectangle ++(.10,\h);
  }
  \node[font=\scriptsize] at ($(trajloss.west)+(.99,-.25)$) {Base Prob.};
  \node[font=\scriptsize] at ($(trajloss.east)+(-.99,-.25)$) {Traj. Prob.};
  \draw[blue!70!black, thick, <->] ($(trajloss.center)+(-0.20,0.10)$) -- ++(.65,0);
  \node[font=\scriptsize] at ($(trajloss.center)+(0,-.50)$) {$\mathcal{L}_{consist}$};
\end{scope}
\node[draw=pink!65!black,rounded corners=5pt,fill=pink!5, minimum width=2.0cm, minimum height=.85cm,anchor=north,align=center] (trajcls)
at ($(trajloss.south)+(0,0.90)$) {Trajectory Classification\\[2pt] $\mathcal{L}_{\text{traj-cls}}$};
\node[draw=purple!50!black, dashed, rounded corners=5pt, thick, fill=purple!2, minimum width=2.75cm, minimum height=4.00cm, anchor=north west] (E) at (18.15,-9.20) {};
\node[titleE] at ($(E.north)+(0.10,-.25)$) {(E) Inference};
\foreach \i/\txt in {
1/{Input Patch},
2/{Dual Tokenization},
3/{Modality Encoding},
4/{Environment State $z_0$},
5/{Latent Rollout $z_0 \rightarrow z_T$},
6/{Rollout Planner $z_{plan}$},
7/{Final Prediction}
}{
    \node[circle,draw=purple!70!black,fill=purple!70!black,text=white,font=\scriptsize,minimum size=3.0mm,inner sep=0pt] at ($(E.north west)+(.28,-.48-.48*\i)$) {\i};
  \node[anchor=west,align=left,font=\tiny,text width=1.55cm] at ($(E.north west)+(.50,-.48-.48*\i)$) {\txt};
}
\node[legendbox, minimum width=20.70cm, minimum height=.65cm, anchor=north] (legend) at (10.5,-13.4) {};
\node[ln] at ($(legend.west)+(.6,0)$) {LN};
\node[anchor=west] at ($(legend.west)+(.9,0)$) {LayerNorm};
\node[mhsa, minimum width=.45cm, minimum height=.45cm] at ($(legend.west)+(2.85,0)$) {MHSA};
\node[anchor=west, align=left] at ($(legend.west)+(3.30,0)$) {Multi-Head Self-Attention};
\node[mlp] at ($(legend.west)+(6.70,0)$) {MLP};
\node[anchor=west, align=left] at ($(legend.west)+(7.10,0)$) {Feed-Forward Network};
\node at ($(legend.west)+(9.85,0)$) {\Large $\oplus$};
\node[anchor=west, align=left] at ($(legend.west)+(10.00,0)$) {Residual Connection};
\node[draw=black!90, dashed, minimum width=.45cm, minimum height=.35cm] at ($(legend.west)+(12.80,0)$) {};
\node[anchor=west, align=left] at ($(legend.west)+(13.00,0)$) {Learnable Embedding};
\node[latent] at ($(legend.west)+(15.80,0)$) {$z_t$};
\node[anchor=west] at ($(legend.west)+(16.10,0)$) {Latent State};
\foreach \i in {0,...,3}{\node[cls] at ($(legend.west)+(17.80+\i*.18,0)$) {};}
\node[anchor=west, align=left] at ($(legend.west)+(18.50,0)$) {CLS Token (Scene)};
\end{tikzpicture}
\end{adjustbox}
\caption{Overview of LWM framework. (A) Input hyperspectral patches are preprocessed with spectral band and spatial pixel masking. (B) Each patch is decomposed into spectral tokens (per-band) and spatial tokens (per-pixel), which are encoded by dedicated transformer branches and fused through a joint transformer to produce a unified representation and a global scene embedding (CLS token) for base classification. (C) An environment state is constructed by aggregating center, neighborhood, spectral, and global features, and is propagated through a stochastic latent transition model to generate multi-step latent rollouts via Gaussian parameterization and recurrent refinement. A rollout planner adaptively aggregates latent states into a planned representation for trajectory-based classification, which is combined with the base prediction. (D) The model is trained using masked reconstruction, spectral and spatial latent dynamics, stochastic rollout regularization with KL constraint, trajectory supervision, and physically motivated regularization. (E) During inference, masking is disabled and prediction is obtained through a forward pass with latent rollout and trajectory aggregation.}
\label{fig:full_architecture}
\end{figure*}

\section{Mathematical Modeling}
\label{sec:method}

The proposed LWM model formulates HSIC as a trajectory-aware spatial-spectral representation learning problem. Each input patch is decomposed into spectral and spatial token sequences and encoded through dedicated Transformer branches, whose outputs are fused into a joint representation. From this representation, a compact environment state is constructed and propagated through a stochastic LWM module to generate multi-step rollouts in representation space. Importantly, the term ``world modeling'' does not imply temporal simulation: the latent state represents the observed spatial-spectral context, while the learned stochastic transition operator maps this state to plausible future latent configurations under the structural dynamics captured by the model. Accordingly, a ``state transition'' denotes a learned transformation in representation space, and a ``rollout'' denotes repeated application of this transition to generate a sequence of latent hypotheses. These latent trajectories do not correspond to explicit temporal observations or imply that the hyperspectral scene evolves over time.

For static HSIC, the purpose of such latent transitions is to provide structurally grounded alternatives to unavailable temporal supervision. Specifically, spectral adjacency, spatial neighborhood relations, and their fused context define the relational structure of the latent transition space. The resulting rollout therefore explores plausible representations consistent with the observed patch structure, rather than simulating physical or temporal evolution. A rollout planner aggregates the latent trajectory into a planned state, which is combined with the current state to produce trajectory-aware class predictions. In this way, the trajectory serves both as a predictive regularizer and as a decision mechanism, allowing the classifier to aggregate multiple structurally plausible latent alternatives rather than relying exclusively on a single static embedding. The overall procedure is summarized in Algorithm~\ref{alg:wmsst} and Fig.~\ref{fig:full_architecture}.

\begin{algorithm}[t]
\caption{Stochastic Latent World Modeling}
\label{alg:wmsst}
\begin{algorithmic}[1]
\Require HSI mini-batch $\mathbf{x}\in\mathbb{R}^{B\times C\times P\times P}$, labels $\mathbf{y}$, coordinates $\mathbf{q}$, rollout steps $T$
\Ensure Final logits $\boldsymbol{\ell}$ and total loss $\mathcal{L}_{\mathrm{total}}$
\State Construct spectral tokens $\mathbf{S}^{0}=\phi_{\mathrm{spec}}(\mathbf{x})\in\mathbb{R}^{B\times C\times D}$
\State Construct spatial tokens $\mathbf{A}^{0}=\phi_{\mathrm{spat}}(\mathbf{x})\in\mathbb{R}^{B\times N\times D}$
\State Encode sequences $\mathbf{S}=\operatorname{SpecEnc}_{\Theta{\lambda}}(\mathbf{S}^{0})$, $\mathbf{A}=\operatorname{SpatEnc}_{\Theta{s}}(\mathbf{A}^{0})$
\State Form fused tokens: $\mathbf{F}^{0}$ (see Eq. \ref{eq7})
\State Fuse tokens $\mathbf{F}=\operatorname{Fuse}_{\Theta}(\mathbf{F}^{0})$
\State Extract components: $\mathbf{h}_{\mathrm{cls}},\mathbf{Z}_{\lambda},\mathbf{Z}_{s}$
\State Compute base logits: $\boldsymbol{\ell}_{\mathrm{base}}=g_{\mathrm{base}}(\mathbf{h}_{\mathrm{cls}})$
\State Construct environment state $\mathbf{z}_{0}$ (see Eq. \ref{eq12})
\State Initialize rollout $\mathbf{Z}_{\mathrm{roll}}\leftarrow[\mathbf{z}_{0}]$
\For{$t=0$ to $T-1$}
\State Predict transition: $[\boldsymbol{\mu}_{t+1},\log\boldsymbol{\sigma}^{2}_{t+1}] = G_{\theta}(\mathbf{z}_{t})$
\State Clip log-variance: $\log\boldsymbol{\sigma}^{2}_{t+1} \leftarrow \operatorname{clip} \left(\log\boldsymbol{\sigma}^{2}_{t+1},a,b\right)$
\State Latent state: $\widetilde{\mathbf{z}}_{t+1} = \boldsymbol{\mu}_{t+1} + \boldsymbol{\epsilon}_{t+1}\odot \exp\left(\tfrac{1}{2}\log\boldsymbol{\sigma}^{2}_{t+1}\right)$
\State Refine state: $\mathbf{z}_{t+1} = R_{\theta}(\widetilde{\mathbf{z}}_{t+1},\mathbf{z}_{t})$
\State Define projected target: $\widehat{\mathbf{z}}_{t+1} = \operatorname{sg}(Q_{\theta}(\mathbf{z}_{t}))$
\State Append $\mathbf{z}_{t+1}$ to rollout
\EndFor
\State Compute rollout weights: $w_t = \operatorname{softmax}(u_{\theta}(\mathbf{z}_{t}))$
\State Aggregate planned state: $\mathbf{z}_{\mathrm{plan}} = \sum_{t=0}^{T}w_t\mathbf{z}_{t}$
\State Compute trajectory logits: $\boldsymbol{\ell}_{\mathrm{traj}}$ (see Eq. \ref{eq20})
\State Final prediction: $\boldsymbol{\ell} = \mathbf{G}\odot\boldsymbol{\ell}_{\mathrm{base}} + \left(\mathbf{1}_{B\times K}-\mathbf{G}\right) \odot \boldsymbol{\ell}_{\mathrm{traj}}$
\State Compute $\mathcal{L}_{\mathrm{total}}$ (see Eq. \ref{Loss})
\State \Return $\boldsymbol{\ell},\mathcal{L}_{\mathrm{total}}$
\end{algorithmic}
\end{algorithm}

\subsubsection{\textbf{Problem Formulation and Notation}} HSIC assigns a semantic label to each spatial location using spectral signatures and local spatial context. Let the HSI be $\mathbf{X}\in\mathbb{R}^{H_0\times W_0\times C}$, where $H_0$, $W_0$, and $C$ denote spatial dimensions and spectral bands. For a labeled pixel at spatial coordinates $\mathbf{q}_i=(r_i,c_i)$, where $r_i$ and $c_i$ denote its row and column indices, respectively, a local odd patch of size $P\times P$ is extracted using mirror padding, yielding $\mathbf{x}_i\in\mathbb{R}^{C\times P\times P}$. For a mini-batch of $B$ samples, the input is $\mathbf{x}\in\mathbb{R}^{B\times C\times P\times P}$ with labels $\mathbf{y}\in\{0,\ldots,K-1\}^{B}$, where $K$ is the number of classes. 

\subsubsection{\textbf{Spectral and Spatial Tokenization}} The first step transforms each hyperspectral patch into complementary spectral and spatial token representations \cite{ViTICLR2021dosovitskiy}. 

\noindent\textbf{Spectral Tokenization.} The spectral tokenizer constructs tokens along the wavelength axis, where each spectral band is treated as an individual token retaining its local spatial context. For the $i$-th sample, the $c$-th spectral band $\mathbf{x}_{i,c,:,:}\in\mathbb{R}^{P\times P}$ is flattened into a vector $\mathbf{u}_{i,c}\in\mathbb{R}^{P^2}$ and projected into a $D$-dimensional embedding space. A learnable spectral positional encoding is added to preserve the ordering of wavelength indices. For a mini-batch, this process is expressed as:

\begin{equation}
    \mathbf{S}^{0} = \phi_{\mathrm{spec}}(\mathbf{x}) = \mathbf{x}_{\mathrm{spec}} \times_{3} \mathbf{W}_{\mathrm{spec}} + \widetilde{\mathbf{P}}_{\lambda} \in \mathbb{R}^{B\times C\times D},
\end{equation}
where $\mathbf{x}_{\mathrm{spec}} \in \mathbb{R}^{B\times C\times P^2}$ is the flattened spectral representation, $\mathbf{W}_{\mathrm{spec}} \in \mathbb{R}^{P^2\times D}$ is a learnable projection matrix, and $\mathbf{P}_{\lambda} \in \mathbb{R}^{1\times C\times D}$ denotes the learnable spectral positional embedding. The batch-expanded positional embedding is defined as $\widetilde{\mathbf{P}}_{\lambda} = \mathbf{1}_{B}\otimes\mathbf{P}_{\lambda} \in \mathbb{R}^{B\times C\times D}$, where $\mathbf{1}_{B}$ is a $B$-dimensional vector of ones and $\otimes$ denotes replication along the batch dimension.

\noindent\textbf{Spatial Tokenization.} The spatial tokenizer constructs tokens along the spatial domain, where each token corresponds to a spatial location and carries its full spectral signature. The input tensor $\mathbf{x}\in\mathbb{R}^{B\times C\times P\times P}$ is rearranged by flattening the two spatial dimensions into a single spatial-token dimension, such that $\mathbf{x}_{\mathrm{spat}}\in\mathbb{R}^{B\times N\times C}$ with $N=P^2$. Thus, for the $i$-th sample and the $p$-th spatial location, the corresponding spectral vector is $\mathbf{x}_{\mathrm{spat},i,p,:}\in\mathbb{R}^{C}$, which is projected into the same $D$-dimensional embedding space. A learnable spatial positional encoding is added to retain spatial layout information. The resulting spatial token sequence is given by

\begin{equation}
    \mathbf{A}^{0} = \phi_{\mathrm{spat}}(\mathbf{x}) = \mathbf{x}_{\mathrm{spat}}\times_{3}\mathbf{W}_{\mathrm{spat}}  + \widetilde{\mathbf{P}}{xy} \in\mathbb{R}^{B\times N\times D},
\end{equation}
where $\mathbf{x}_{\mathrm{spat}}\in\mathbb{R}^{B\times N\times C}$ is the spatially rearranged tensor, $\mathbf{W}_{\mathrm{spat}}\in\mathbb{R}^{C\times D}$ is a learnable projection matrix, and $\mathbf{P}_{xy}\in\mathbb{R}^{1\times N\times D}$ represents the spatial positional embeddings. The batch-expanded positional embedding is defined as $\widetilde{\mathbf{P}}_{xy} = \mathbf{1}_{B} \otimes \mathbf{P}_{xy} \in\mathbb{R}^{B\times N\times D}$.

\subsubsection{\textbf{Transformer Encoding}} Let $\operatorname{SpecEnc}_{\Theta{\lambda}}(\cdot)$ denote the spectral transformer encoder composed of $L_{\lambda}$ transformer layers, and let $\operatorname{SpatEnc}_{\Theta{s}}(\cdot)$ denote the spatial transformer encoder composed of $L_{s}$ transformer layers. The corresponding encoded representations are defined as

\begin{equation}
    \mathbf{S} = \operatorname{SpecEnc}_{\Theta{\lambda}}(\mathbf{S}^{0}) \in\mathbb{R}^{B\times C\times D},
\end{equation}
\begin{equation}
    \mathbf{A} = \operatorname{SpatEnc}_{\Theta{s}}(\mathbf{A}^{0}) \in\mathbb{R}^{B\times N\times D},
\end{equation}
where $C$ and $N=P^2$ denote the sequence lengths of the spectral and spatial branches, respectively. The parameters $\Theta_{\lambda}$ and $\Theta_{s}$ are learned independently. For a token sequence $\mathbf{T}^{\ell-1}\in\mathbb{R}^{B\times M\times D}$ at layer $\ell-1$, a transformer block updates the representation through pre-normalized self-attention and feed-forward sublayers:

\begin{equation}
    \widetilde{\mathbf{T}}^{\ell} = \mathbf{T}^{\ell-1} + \operatorname{DropPath}\left(\operatorname{MHSA}\left(\operatorname{LN}(\mathbf{T}^{\ell-1})\right)\right),
\end{equation}
\begin{equation}
    \mathbf{T}^{\ell} = \widetilde{\mathbf{T}}^{\ell} + \operatorname{DropPath}\left(\operatorname{MLP}\left(\operatorname{LN}(\widetilde{\mathbf{T}}^{\ell})\right)\right),
\end{equation}
where $\operatorname{MHSA}(\cdot)$ denotes multi-head self-attention \cite{vaswani2017attention}, $\operatorname{MLP}(\cdot)$ denotes a feed-forward network, $\operatorname{LN}(\cdot)$ denotes layer normalization, and $\operatorname{DropPath}(\cdot)$ introduces stochastic depth. The sequence length $M$ is set to $C$ for the spectral branch and $N$ for the spatial branch. Applying this transformer block for $L_{\lambda}$ and $L_s$ layers yields the encoded representations $\mathbf{S}$ and $\mathbf{A}$, respectively.

\noindent\textbf{Fusion.} To enable joint modeling, the spectral and spatial representations are fused using a transformer encoder (using $L_f$ transformer layers). A learnable classification token $\mathbf{c}_{0}\in\mathbb{R}^{1\times 1\times D}$ and type embeddings $\mathbf{e}_{\mathrm{cls}},\mathbf{e}_{\lambda},\mathbf{e}_{s}\in\mathbb{R}^{1\times 1\times D}$ are introduced to distinguish token types. Since these embeddings are shared across the mini-batch, their batch-expanded forms are defined as $\widetilde{\mathbf{c}}_{0} = \mathbf{1}_{B}\otimes\mathbf{c}_{0} \in\mathbb{R}^{B\times 1\times D}$, $\widetilde{\mathbf{e}}_{\mathrm{cls}} = \mathbf{1}_{B}\otimes\mathbf{e}_{\mathrm{cls}} \in\mathbb{R}^{B\times 1\times D}$, $\widetilde{\mathbf{e}}_{\lambda} = \mathbf{1}_{B}\otimes\mathbf{e}_{\lambda} \in\mathbb{R}^{B\times 1\times D}$, and $\widetilde{\mathbf{e}}_{s} = \mathbf{1}_{B}\otimes\mathbf{e}_{s} \in\mathbb{R}^{B\times 1\times D}$. Let $\operatorname{Broadcast}_{C}(\cdot)$ and $\operatorname{Broadcast}_{N}(\cdot)$ denote replication along the spectral and spatial sequence dimensions, respectively, and let $\operatorname{Concat}{\mathrm{seq}}(\cdot)$ denote concatenation along the sequence dimension. The fused input sequence is defined as

\begin{equation}
\begin{aligned}
    \mathbf{F}^{0} &= \operatorname{Concat}_{\mathrm{seq}} \Big(\widetilde{\mathbf{c}}_{0} + \widetilde{\mathbf{e}}_{\mathrm{cls}}, \mathbf{S} + \operatorname{Broadcast}_{C} \left(\widetilde{\mathbf{e}}_{\lambda}\right), \\
    & \mathbf{A} + \operatorname{Broadcast}_{N} \left(\widetilde{\mathbf{e}}_{s}\right)\Big) \in \mathbb{R}^{B\times(1+C+N)\times D}.
    \label{eq7}
\end{aligned}
\end{equation}

The fused sequence is processed by the fusion transformer encoder as $\mathbf{F} = \operatorname{Fuse}_{\Theta}(\mathbf{F}^{0}) \in \mathbb{R}^{B\times(1+C+N)\times D}$. The output is decomposed as $\mathbf{h}_{\mathrm{cls}} = \mathbf{F}_{:,0,:}\in\mathbb{R}^{B\times D}$, $\mathbf{Z}_{\lambda} = \mathbf{F}_{:,1:1+C,:}\in\mathbb{R}^{B\times C\times D}$, and $\mathbf{Z}_{s} = \mathbf{F}_{:,1+C:,:}\in\mathbb{R}^{B\times N\times D}$.
The classification token $\mathbf{h}_{\mathrm{cls}}$ provides a global representation for the base prediction. The base classification head $g_{\mathrm{base}}:\mathbb{R}^{D}\rightarrow\mathbb{R}^{K}$ consists of layer normalization, a linear layer, GELU activation, dropout, and a final linear layer mapping to $K$ classes. The base logits are given by
\begin{equation}
    \boldsymbol{\ell}_{\mathrm{base}} = g_{\mathrm{base}}(\mathbf{h}_{\mathrm{cls}}) \in\mathbb{R}^{B\times K},
\end{equation}
and serve as an initial prediction that is refined through latent rollout modeling in subsequent stages.

\subsection{Environment State Construction}

Following spatial-spectral encoding and fusion, a compact latent representation is constructed to capture the local environment of the target pixel. While the classification token $\mathbf{h}_{\mathrm{cls}}$ provides a global summary, it does not explicitly separate local spatial structure and spectral composition. To address this, the environment state is formed by aggregating complementary components derived from the fused tokens. Let $n_0=\frac{P^2-1}{2}$ denote the flattened index of the center location in the spatial token sequence, assuming $P$ is odd, and let $\mathcal{N}_8(n_0)$ denote the set of indices corresponding to its eight first-ring neighbors. The center state is defined as $\mathbf{z}_{c}=\mathbf{Z}_{s,:,n_0}\in\mathbb{R}^{B\times N \times D}$, which encodes the latent representation of the target pixel. The neighbor state is computed as:

\begin{equation}
    \mathbf{z}_{n} = \frac{1}{|\mathcal{N}_8(n_0)|} \sum_{j\in\mathcal{N}_8(n_0)} \mathbf{Z}_{s,:,j,:} \in\mathbb{R}^{B\times D},
\end{equation}
capturing local spatial context. The global spectral state is given by:

\begin{equation}
    \mathbf{z}_{\lambda} = \frac{1}{C} \sum_{c=1}^{C} \mathbf{Z}_{\lambda,:,c,:} \in\mathbb{R}^{B\times D},
\end{equation}
and the scene-level representation is $\mathbf{z}_{g}=\mathbf{h}_{\mathrm{cls}}\in\mathbb{R}^{B\times D}$. These components are concatenated along the column dimension using the operator $\operatorname{Concat}_{\mathrm{col}}(\cdot)$, yielding $\mathbf{z}_{\mathrm{cat}} = \operatorname{Concat}_{\mathrm{col}} \left(\mathbf{z}_{c}, \mathbf{z}_{n}, \mathbf{z}_{\lambda}, \mathbf{z}_{g}\right) \in\mathbb{R}^{B\times 4D}$. The concatenated representation is then mapped into a unified latent representation:

\begin{equation}
    \mathbf{z}_{0} = E_{\mathrm{env}}(\mathbf{z}_{\mathrm{cat}}) \in\mathbb{R}^{B\times D},
    \label{eq12}
\end{equation}
where $E_{\mathrm{env}}:\mathbb{R}^{B \times 4D}\rightarrow\mathbb{R}^{B \times D}$ is implemented using layer normalization, linear projection, Gaussian error linear unit (GELU) activation, dropout, and a final normalization layer. The resulting $\mathbf{z}_{0}$ serves as the initial latent environment state, integrating local spatial, spectral, and global contextual information for subsequent latent rollout modeling.

\subsection{Stochastic Latent World Modeling}

Given the environment state $\mathbf{z}_{0}$, the model simulates its evolution in latent space through a stochastic rollout mechanism \cite{105281zenodo1207631, hafner2019planet}. This process captures variations of the local spatial-spectral representation by modeling uncertainty in latent transitions \cite{29694422969572}. The stochastic LWM module defines a multi-step rollout starting from $\mathbf{z}_{0}$ \cite{chung2015recurrent}. For $t=\{0,\ldots,T-1\}$, a transition network $G_{\theta}$ predicts the parameters of a Gaussian latent transition:

\begin{equation}
    [\boldsymbol{\mu}_{t+1},\log\boldsymbol{\sigma}^{2}_{t+1}] = G_{\theta}(\mathbf{z}_{t}) \in\mathbb{R}^{B\times 2D},
\end{equation}
where $G_{\theta}:\mathbb{R}^{B\times D}\rightarrow\mathbb{R}^{B\times 2D}$ is implemented as a feed-forward network consisting of layer normalization, a linear layer mapping D to 2D, GELU activation, dropout, and a final linear layer mapping 2D to 2D. The output is split along the feature dimension into the predicted mean $\boldsymbol{\mu}_{t+1}\in\mathbb{R}^{B\times D}$ and log-variance $\log\boldsymbol{\sigma}^{2}_{t+1}\in\mathbb{R}^{B\times D}$. For numerical stability, the predicted log-variance is clipped element-wise to a fixed interval $[a,b]$, where $a<b$ are predefined lower and upper bounds, respectively:

\begin{equation}
    \log\boldsymbol{\sigma}^{2}_{t+1} \leftarrow \operatorname{clip}\left(\log\boldsymbol{\sigma}^{2}_{t+1},a,b\right),
\end{equation}
where, for each element $v$ of $\log\boldsymbol{\sigma}^{2}_{t+1}$ is $\operatorname{clip}(v,a,b)=\min\{\max\{v,a\},b\}$. Thus, all predicted log-variance values are constrained to the interval $[a,b]$. During training, the next latent state is sampled using the reparameterization trick \cite{34957243497510, kingma2014autoencoding}:

\begin{equation}
    \widetilde{\mathbf{z}}_{t+1} = \boldsymbol{\mu}_{t+1} + \boldsymbol{\epsilon}_{t+1} \odot \exp\left(\frac{1}{2}\log\boldsymbol{\sigma}^{2}_{t+1}\right),
\end{equation}
where $\boldsymbol{\epsilon}_{t+1} \sim \mathcal{N}_{B\times D} \left(\mathbf{0}_{B\times D},\mathbf{I}_{B}\otimes\mathbf{I}_{D}\right)$, so that $\boldsymbol{\epsilon}_{t+1}\in\mathbb{R}^{B\times D}$ and $\odot$ denotes element-wise Hadamard multiplication. During inference, stochastic sampling is disabled and $\widetilde{\mathbf{z}}_{t+1}=\boldsymbol{\mu}_{t+1}$. The state is then refined via the recurrent update

\begin{equation}
    \mathbf{z}_{t+1} = R_{\theta} \left(\widetilde{\mathbf{z}}_{t+1},\mathbf{z}_{t}\right) \in\mathbb{R}^{B\times D},
    \label{Eq16}
\end{equation}
where $R_{\theta}$ is implemented as a gated recurrent unit (GRU) cell \cite{chung2014empirical}. This yields a latent trajectory. Since each latent state, predicted mean, and log-variance is represented as a matrix in $\mathbb{R}^{B\times D}$, we define $\operatorname{Concat}_{3}(\cdot)$ as concatenation along a new third dimension. For $T$ matrices $\{\mathbf{X}_{1},\ldots,\mathbf{X}_{T}\}\in\mathbb{R}^{B\times D}$ as $\operatorname{Concat}_{3} \left(\mathbf{X}_{1},\ldots,\mathbf{X}_{T} \right) \in\mathbb{R}^{B\times D\times T}$, and $\operatorname{reshape}_{B\times T\times D}(\cdot)$ denotes the corresponding reshaping of this tensor into the batch-trajectory-feature ordering. The latent trajectory is therefore defined as $\mathbf{Z}_{\mathrm{roll}} = \operatorname{reshape}_{B\times(T+1)\times D} \left(\operatorname{Concat}_{3} \left(\mathbf{z}_{0},\mathbf{z}_{1},\ldots,\mathbf{z}_{T}\right)\right)$, where $\mathbf{Z}_{\mathrm{roll}}\in\mathbb{R}^{B\times(T+1)\times D}$. 

Similarly, the predicted means and log-variances are collected as $\boldsymbol{\mu} = \operatorname{reshape}_{B\times T\times D} \left(\operatorname{Concat}_{3}\left(\boldsymbol{\mu}_{1},\ldots,\boldsymbol{\mu}_{T}\right)\right) \in\mathbb{R}^{B\times T\times D}$, and $\mathbf{V} = \operatorname{reshape}_{B\times T\times D} \left(\operatorname{Concat}_{3} \left(\log\boldsymbol{\sigma}^{2}_{1},\ldots,\log\boldsymbol{\sigma}^{2}_{T}\right)\right) \in\mathbb{R}^{B\times T\times D}$. 

To construct an auxiliary latent target, let $Q_{\theta}:\mathbb{R}^{B\times D}\rightarrow\mathbb{R}^{B\times D}$ denote a learnable latent projection network that maps the current latent state to a target representation. The stop-gradient operator $\operatorname{sg}(\cdot)$ returns its input while preventing gradients from propagating through that input during backpropagation, i.e., $\operatorname{sg}(\mathbf{u})=\mathbf{u},~ \frac{\partial \operatorname{sg}(\mathbf{u})}{\partial\mathbf{u}}=\mathbf{0}$. The auxiliary latent target is then defined as:

\begin{equation}
    \widehat{\mathbf{z}}_{t+1} = \operatorname{sg}\left(Q_{\theta}(\mathbf{z}_{t})\right) \in\mathbb{R}^{B\times D},
\end{equation}
and the corresponding sequence is constructed using the same third-dimension concatenation and reshaping operation $\widehat{\mathbf{Z}} = \operatorname{reshape}_{B\times T\times D} \left(\operatorname{Concat}_{3} \left(\widehat{\mathbf{z}}_{1},\ldots,\widehat{\mathbf{z}}_{T}\right)\right) \in\mathbb{R}^{B\times T\times D}$. This target acts as an internal projection used to regularize the stochastic transitions. The rollout operates entirely in representation space and does not correspond to explicit temporal observations. The resulting latent trajectory provides a set of states that are subsequently used for trajectory-aware aggregation and classification.

\subsection{Trajectory Planning and Classification}

Given the latent trajectory ${\mathbf{z}_{t}}_{\{t=0\}}^{T}$, the model performs trajectory-aware aggregation to obtain a decision-relevant representation. For each rollout state $\mathbf{z}_{t}\in\mathbb{R}^{B\times D}$, a learnable scoring network $u_{\theta}:\mathbb{R}^{D}\rightarrow\mathbb{R}$ assigns a scalar score independently to each sample. Specifically, $u_{\theta}$ is implemented as a two-layer MLP with layer normalization, a linear projection from $D$ to $D/2$, GELU activation, and a linear projection to a scalar:

\begin{equation}
    u_{\theta}(\mathbf{z}) = \mathbf{w}_{u}^{\top} \operatorname{GELU} \left(\mathbf{W}_{u} \operatorname{LN}(\mathbf{z}) + \mathbf{b}_{u}\right) + b_{u},
\end{equation}
where $\mathbf{W}_{u}\in\mathbb{R}^{\frac{D}{2}\times D}$, $\mathbf{b}_{u}\in\mathbb{R}^{D/2}$, $\mathbf{w}_{u}\in\mathbb{R}^{D/2}$, and $b_{u}\in\mathbb{R}$ are learnable parameters. Thus, for sample $i$ and rollout step $t$, the corresponding score is $\alpha_{i,t} = u_{\theta}(\mathbf{z}_{i,t}) \in\mathbb{R},~i=\{1,\ldots,B\},~t=\{0,\ldots,T\}$, where $\mathbf{z}_{i,t}\in\mathbb{R}^{D}$ denotes the $i$-th row of $\mathbf{z}_{t}$. The scores are normalized independently for each sample along the trajectory dimension using the softmax operator:

\begin{equation}
    w_{i,t} = \operatorname{softmax}(\boldsymbol{\alpha}_{i})_{t} = \frac{\exp(\alpha_{i,t})} {\sum_{\tau=0}^{T}\exp(\alpha_{i,\tau})} ,
\end{equation}
where $\sum_{t=0}^{T}w_{i,t}=1,~ w_{i,t}\geq 0$ and $\boldsymbol{\alpha}_{i} = [\alpha_{i,0},\ldots,\alpha_{i,T}] \in\mathbb{R}^{T+1}$. Equivalently, $\operatorname{softmax}(\boldsymbol{\alpha}_{i})$ converts the scalar scores associated with the $T+1$ rollout states of sample $i$ into normalized trajectory weights. The planned representation is then obtained by a weighted aggregation of the latent states $\mathbf{z}_{\mathrm{plan},i} = \sum_{t=0}^{T} w_{i,t}\mathbf{z}_{i,t} \in\mathbb{R}^{D}$, for the complete mini-batch,

\begin{equation}
    \mathbf{z}_{\mathrm{plan}} = \sum_{t=0}^{T} \mathbf{w}_{t}\odot\mathbf{z}_{t} \in\mathbb{R}^{B\times D},
\end{equation}
where $\mathbf{w}_{t}=[w_{1,t},\ldots,w_{B,t}]^{\top}\in\mathbb{R}^{B}$ and the weights are broadcast across the feature dimension. The trajectory representation is constructed by concatenating the initial latent state and the planned state along the feature dimension:

\begin{equation}
    \mathbf{r}_{\mathrm{traj}} = \operatorname{Concat}_{\mathrm{col}} \left(\mathbf{z}_{0}, \mathbf{z}_{\mathrm{plan}}\right) \in\mathbb{R}^{B\times 2D}.
\end{equation}

The trajectory-based logits are produced by a trajectory classifier $g_{\mathrm{traj}}:\mathbb{R}^{2D}\rightarrow\mathbb{R}^{K}$ which consists of layer normalization, a linear projection from $2D$ to $D$, GELU activation, dropout, and a final linear projection from $D$ to $K$ as $g_{\mathrm{traj}}(\mathbf{r}) = \mathbf{W}_{\mathrm{traj},2} \operatorname{GELU} \left(\mathbf{W}_{\mathrm{traj},1} \operatorname{LN}(\mathbf{r}) + \mathbf{b}_{\mathrm{traj},1}\right) + \mathbf{b}_{\mathrm{traj},2}$, where $\mathbf{W}_{\mathrm{traj},1}\in\mathbb{R}^{D\times 2D}$, $\mathbf{b}_{\mathrm{traj},1}\in\mathbb{R}^{D}$, $\mathbf{W}_{\mathrm{traj},2}\in\mathbb{R}^{K\times D}$, and $\mathbf{b}_{\mathrm{traj},2}\in\mathbb{R}^{K}$ are learnable parameters, and dropout is applied between the GELU activation and the linear projection. Thus,

\begin{equation}
    \boldsymbol{\ell}_{\mathrm{traj}} = g_{\mathrm{traj}}\left(\mathbf{r}_{\mathrm{traj}}\right) \in\mathbb{R}^{B\times K}.
    \label{eq20}
\end{equation}

The final prediction combines the base and trajectory logits using a sample-dependent branch gate. The gate is computed from the concatenation of the fused CLS representation and the planned latent state as:

\begin{align*}
    g_i = \sigma\Big(\mathbf{w}_g^\top \operatorname{GELU}\Big(\mathbf{W}_g \operatorname{LN}\Big(\operatorname{Concat}_{\mathrm{col}} \big(\mathbf{h}_{\mathrm{cls},i}, \operatorname{sg}(\mathbf{z}_{\mathrm{plan},i})\big) \Big) \\ + \mathbf{b}_g \Big) + b_g\Big),
\end{align*}
where $\mathbf{h}_{\mathrm{cls},i}\in\mathbb{R}^{D}$ and $\mathbf{z}_{\mathrm{plan},i}\in\mathbb{R}^{D}$ denote the $i$-th samples of the CLS and planned representations, respectively, $\mathbf{W}_{g}\in\mathbb{R}^{D\times 2D}$, $\mathbf{b}_{g}\in\mathbb{R}^{D}$, $\mathbf{w}_{g}\in\mathbb{R}^{D}$, and $b_{g}\in\mathbb{R}$ are learnable parameters, and $\sigma(\cdot)$ denotes the sigmoid function. The stop-gradient operation $\operatorname{sg}(\cdot)$ prevents gradients from the branch-gating network from propagating through the planned state. Collecting the sample-wise gates gives $\mathbf{g} = [g_{1},\ldots,g_{B}]^{\top} \in(0,1)^{B\times 1}$. To apply the scalar gate to all $K$ class logits of each sample, define $\mathbf{G} = \mathbf{g}~\mathbf{1}_{K}^{\top} \in(0,1)^{B\times K}$, where $\mathbf{1}_{K}\in\mathbb{R}^{K}$ is a vector of ones. The final logits are then obtained by element-wise interpolation:

\begin{equation}
    \boldsymbol{\ell} = \mathbf{G}\odot\boldsymbol{\ell}_{\mathrm{base}} + \left(\mathbf{1}_{B\times K}-\mathbf{G}\right) \odot \boldsymbol{\ell}_{\mathrm{traj}} \in\mathbb{R}^{B\times K},
\end{equation}
where $\odot$ denotes element-wise multiplication and $\mathbf{1}_{B\times K}$ is a $B\times K$ matrix of ones. Thus, each sample uses a single learned gate to control the contribution of the base and trajectory branches across all classes. For the predicted label, let $\boldsymbol{\ell}_{i}\in\mathbb{R}^{K}$ denote the $i$-th row of the final logit matrix $\boldsymbol{\ell}\in\mathbb{R}^{B\times K}$. The predicted class is given by

\begin{equation}
    \widehat{y}_{i} = \argmax_{k\in\{0,\ldots,K-1\}} \operatorname{softmax} \left(\boldsymbol{\ell}_{i} \right)_{k}.
\end{equation}

This trajectory planning mechanism enables the model to aggregate information across the latent rollout while retaining the initial environment state for trajectory-based classification. By learning sample-dependent importance weights over the rollout states, the model incorporates structured information from the latent trajectory into the final prediction.

\subsection{Reconstruction and Dynamics Constraints}

The model incorporates reconstruction and latent dynamics constraints to regularize the spectral and spatial token representations. These constraints operate on the fused spectral and spatial representations $\mathbf{Z}_{\lambda}$ and $\mathbf{Z}_{s}$ obtained from the fusion transformer. In particular, the reconstruction heads recover the corresponding input-domain representations, while the dynamics modules impose predictive structure along the spectral sequence and across the local spatial neighborhood.

\noindent\textbf{Spectral Reconstruction.} The spectral reconstruction head $R_{\lambda}$ maps each $D$-dimensional spectral token back to the flattened spatial patch associated with the corresponding spectral band. Given $\mathbf{Z}_{\lambda}\in\mathbb{R}^{B\times C\times D}$, the same reconstruction mapping is applied independently to all $C$ spectral tokens. Specifically, $R_{\lambda}:\mathbb{R}^{D}\rightarrow\mathbb{R}^{N}$ is implemented using layer normalization, a linear projection from $D$ to $D$, GELU activation, and a final linear projection from $D$ to $N$:

\begin{equation}
    R_{\lambda}(\mathbf{z}) = \mathbf{W}_{\lambda,2}^{\mathrm{rec}} \operatorname{GELU} \left(\mathbf{W}_{\lambda,1}^{\mathrm{rec}} \operatorname{LN}(\mathbf{z}) + \mathbf{b}_{\lambda,1}^{\mathrm{rec}}\right) + \mathbf{b}_{\lambda,2}^{\mathrm{rec}},
\end{equation}
where $\mathbf{W}_{\lambda,1}^{\mathrm{rec}}\in\mathbb{R}^{D\times D}$, $\mathbf{b}_{\lambda,1}^{\mathrm{rec}}\in\mathbb{R}^{D}$, $\mathbf{W}_{\lambda,2}^{\mathrm{rec}}\in\mathbb{R}^{N\times D}$, and $\mathbf{b}_{\lambda,2}^{\mathrm{rec}}\in\mathbb{R}^{N}$ are learnable parameters. Applying $R_{\lambda}$ to every spectral token gives $\widehat{\mathbf{x}}_{\lambda} = R_{\lambda}(\mathbf{Z}_{\lambda}) \in\mathbb{R}^{B\times C\times N}$. The corresponding target is obtained by flattening the two spatial dimensions of the input patch $\mathbf{x}_{\lambda} = \operatorname{reshape}(\mathbf{x}) \in\mathbb{R}^{B\times C\times N}$. Thus, the $c$-th output token of $R_{\lambda}$ reconstructs the $P\times P$ spatial values belonging to the $c$-th spectral band.

\noindent\textbf{Spatial Reconstruction.} The spatial reconstruction head $R_{s}$ maps each spatial token back to its corresponding spectral vector. Given $\mathbf{Z}_{s}\in\mathbb{R}^{B\times N\times D}$, the same reconstruction mapping is applied independently to all $N$ spatial tokens. The mapping $R_{s}:\mathbb{R}^{D}\rightarrow\mathbb{R}^{C}$ is implemented using layer normalization, a linear projection, GELU activation, and a linear projection from $D$ to $C$:

\begin{equation}
    R_{s}(\mathbf{z}) = \mathbf{W}_{s,2}^{\mathrm{rec}} \operatorname{GELU} \left(\mathbf{W}_{s,1}^{\mathrm{rec}} \operatorname{LN}(\mathbf{z}) + \mathbf{b}_{s,1}^{\mathrm{rec}}\right) + \mathbf{b}_{s,2}^{\mathrm{rec}},
\end{equation}
where $\mathbf{W}_{s,1}^{\mathrm{rec}}\in\mathbb{R}^{D\times D}$, $\mathbf{b}_{s,1}^{\mathrm{rec}}\in\mathbb{R}^{D}$, $\mathbf{W}_{s,2}^{\mathrm{rec}}\in\mathbb{R}^{C\times D}$, and $\mathbf{b}_{s,2}^{\mathrm{rec}}\in\mathbb{R}^{C}$ are learnable parameters. Applying $R_{s}$ to all spatial tokens yields $\widehat{\mathbf{x}}_{s} = R_{s}(\mathbf{Z}_{s}) \in\mathbb{R}^{B\times N\times C}$. The corresponding target is obtained by first permuting the input tensor so that the spatial dimensions precede the spectral dimension and then flattening the spatial dimensions $\mathbf{x}_{s} = \operatorname{reshape} \left(\operatorname{permute}(\mathbf{x})\right) \in\mathbb{R}^{B\times N\times C}$. Hence, each spatial token is trained to reconstruct the complete $C$-dimensional spectral signature of its corresponding spatial location. When masking is applied, the reconstruction losses are evaluated only on the masked spectral bands or spatial locations; otherwise, all entries are used. Let $\rho_{\beta}(\cdot,\cdot)$ denote the element-wise Smooth-$\ell_{1}$ penalty with transition parameter $\beta>0$:

\begin{equation}
\rho_{\beta}(a,b) =
    \begin{cases}
        \dfrac{1}{2}(a-b)^{2}/\beta, & |a-b|<\beta, \\
        |a-b|-\dfrac{1}{2}\beta, & |a-b|\geq\beta.
    \end{cases}
\end{equation}

The spectral and spatial reconstruction losses are then defined as $\mathcal{L}_{\mathrm{spec\text{-}rec}} = \operatorname{mean} \left[\rho_{\beta_{\mathrm{rec}}} \left(\widehat{\mathbf{x}}_{\lambda},\mathbf{x}_{\lambda}\right)\right]$ and $\mathcal{L}_{\mathrm{spat\text{-}rec}} = \operatorname{mean} \left[\rho_{\beta_{\mathrm{rec}}} \left(\widehat{\mathbf{x}}_{s}, \mathbf{x}_{s}\right)\right]$, where the mean is taken over the selected masked entries when masking is enabled and over all entries otherwise.

\noindent\textbf{Spectral Dynamics.} The spectral dynamics module models the transition from one spectral token to the next. Given $\mathbf{Z}_{\lambda}=[\mathbf{z}_{\lambda,1},\ldots,\mathbf{z}_{\lambda,C}]$ with $\mathbf{z}_{\lambda,c}\in\mathbb{R}^{B\times D}$, the initial recurrent state is the first spectral token $\mathbf{h}_{\lambda,1} = \mathbf{Z}_{\lambda,:,1,:} \in\mathbb{R}^{B\times D}$. For each subsequent spectral position $c=\{2,\ldots,C\}$, the recurrent state is updated using a GRU cell:

\begin{equation}
    \mathbf{h}_{\lambda,c} = \operatorname{GRUCell}_{\lambda} \left(\mathbf{Z}_{\lambda,:,c-1,:}, \mathbf{h}_{\lambda,c-1}\right) \in\mathbb{R}^{B\times D}.
\end{equation}

The updated state is mapped to the predicted representation of the next spectral token using a prediction head consisting of layer normalization, a linear projection, GELU activation, dropout, and a linear projection from $D$ to $D$:

\begin{equation}
    \widehat{\mathbf{z}}_{\lambda,c} = H_{\lambda} \left(\mathbf{h}_{\lambda,c}\right) \in\mathbb{R}^{B\times D},
\end{equation}
where $H_{\lambda}(\mathbf{h}) = \mathbf{W}_{\lambda,2}^{\mathrm{dyn}} \operatorname{Dropout} [\operatorname{GELU} (\mathbf{W}_{\lambda,1}^{\mathrm{dyn}} \operatorname{LN}(\mathbf{h}) + \mathbf{b}_{\lambda,1}^{\mathrm{dyn}})] + \mathbf{b}_{\lambda,2}^{\mathrm{dyn}}$. The resulting predictions are collected as $\widehat{\mathbf{Z}}_{\lambda}^{+} = \operatorname{stack} \left(\widehat{\mathbf{z}}_{\lambda,2}, \ldots, \widehat{\mathbf{z}}_{\lambda,C} \right) \in\mathbb{R}^{B\times(C-1)\times D}$. The corresponding target is the sequence of subsequent spectral tokens with gradients detached:

\begin{equation}
    \mathbf{Z}_{\lambda}^{+} = \operatorname{sg} \left(\mathbf{Z}_{\lambda,:,2,:}\right) \in\mathbb{R}^{B\times(C-1)\times D},
\end{equation}

Thus, $\widehat{\mathbf{Z}}_{\lambda}^{+}$ is not a slice of the latent rollout tensor $\widehat{\mathbf{Z}}$; it is the sequence of next-band predictions generated directly from the spectral tokens by the spectral dynamics module. The spectral dynamics loss is defined as the mean-squared error between the predicted next-band representations and their detached targets $\mathcal{L}_{\mathrm{spec\text{-}dyn}} = \operatorname{mean} [(\widehat{\mathbf{Z}}_{\lambda}^{+} - \mathbf{Z}_{\lambda}^{+})^{2}]$.

\noindent\textbf{Spatial Dynamics.} The spatial dynamics module models the relationship between the center spatial token and its eight first-ring neighbors. Let $\mathcal{N}_{8}(n_{0})$ denote the set of the eight spatial indices obtained from the offsets $(\Delta_r,\Delta_c) \in \{-1,0,1\}^{2}\setminus{(0,0)}$, around the center location. The corresponding neighboring spatial tokens are $\mathbf{Z}_{s,\mathcal{N}} = \mathbf{Z}_{s,:,,\mathcal{N}_{8}(n_{0}),:} \in\mathbb{R}^{B\times 8\times D}$, while the center token is $\mathbf{z}_{s,c} = \mathbf{Z}_{s,:,n_{0},:} \in\mathbb{R}^{B\times D}$. The center token is passed through a prediction head consisting of layer normalization, a linear projection, GELU activation, dropout, and a linear projection from $D$ to $D$:

\begin{equation}
    \widehat{\mathbf{z}}_{s,n} = H_{s} \left(\mathbf{z}_{s,c}\right) \in\mathbb{R}^{B\times D},
\end{equation}
where $H_{s}(\mathbf{z}) = \mathbf{W}_{s,2}^{\mathrm{dyn}} \operatorname{Dropout} [\operatorname{GELU} (\mathbf{W}_{s,1}^{\mathrm{dyn}} \operatorname{LN}(\mathbf{z}) + \mathbf{b}_{s,1}^{\mathrm{dyn}})] + \mathbf{b}_{s,2}^{\mathrm{dyn}}$. The same predicted representation is compared with each of the eight first-ring neighbors $\widehat{\mathbf{Z}}_{s,\mathcal{N}} = \operatorname{Broadcast}_{8} \left(\widehat{\mathbf{z}}_{s,n} \right) \in\mathbb{R}^{B\times 8\times D}$, where $\operatorname{Broadcast}_{8}(\cdot)$ replicates the predicted center representation across the eight neighbor positions. The corresponding target is detached from the encoder $\mathbf{Z}_{s,\mathcal{N}}^{\mathrm{stop}} = \operatorname{sg}\left(\mathbf{Z}_{s,:,,\mathcal{N}_{8}(n_{0}),:} \right) \in\mathbb{R}^{B\times 8\times D}$. The spatial dynamics loss is therefore $\mathcal{L}_{\mathrm{spat\text{-}dyn}} = \operatorname{mean} [(\widehat{\mathbf{Z}}_{s,\mathcal{N}} - \mathbf{Z}_{s,\mathcal{N}}^{\mathrm{stop}})^{2}]$.

\noindent\textbf{Composite Training Objective.} The complete training objective combines the final classification loss, base classification loss, reconstruction constraints, spectral and spatial dynamics losses, stochastic latent rollout regularization, trajectory supervision, trajectory consistency, and the additional spectral and spatial structural constraints. To avoid fixed weighting constants, all relative contributions are controlled by nonnegative learnable training hyperparameters:

\begin{equation}
\begin{aligned}
    \mathcal{L}_{\mathrm{total}} ={}&  \lambda_{\mathrm{final}}\mathcal{L}_{\mathrm{final}} + \lambda_{\mathrm{base}}\mathcal{L}_{\mathrm{base}} + \lambda_{\mathrm{sr}}\mathcal{L}_{\mathrm{spec\text{-}rec}} + \\ &
    \lambda_{\mathrm{pr}}\mathcal{L}_{\mathrm{spat\text{-}rec}} + \lambda_{\mathrm{sd}}\mathcal{L}_{\mathrm{spec\text{-}dyn}} + \lambda_{\mathrm{pd}}\mathcal{L}_{\mathrm{spat\text{-}dyn}} + \\ &
    \lambda_{\mathrm{env}}\mathcal{L}_{\mathrm{env}} + \lambda_{\mathrm{KL}}\mathcal{L}_{\mathrm{KL}} + \lambda_{\mathrm{traj}}\mathcal{L}_{\mathrm{traj\text{-}cls}} + \lambda_{\mathrm{tc}}\mathcal{L}_{\mathrm{traj\text{-}cons}} + \\ &
    \lambda_{\mathrm{SAM}}\mathcal{L}_{\mathrm{SAM}} + \lambda_{\mathrm{sm}}\mathcal{L}_{\mathrm{smooth}} + \lambda_{\mathrm{sc}}\mathcal{L}_{\mathrm{spat\text{-}cons}},
\end{aligned}
\label{Loss}
\end{equation}
where all coefficients $\lambda_{\mathrm{final}}, \lambda_{\mathrm{base}}, \lambda_{\mathrm{sr}}, \lambda_{\mathrm{pr}}, \lambda_{\mathrm{sd}}, \lambda_{\mathrm{pd}}, \lambda_{\mathrm{env}},$ $\lambda_{\mathrm{KL}}, \lambda_{\mathrm{traj}}, \lambda_{\mathrm{tc}}, \lambda_{\mathrm{SAM}}, \lambda_{\mathrm{sm}}, \lambda_{\mathrm{sc}} \geq 0$ are hyperparameters. The final and base classification losses use label-smoothed cross-entropy. Let $\boldsymbol{\ell}_{i}\in\mathbb{R}^{K}$ denote the final logits for sample $i$, $\boldsymbol{\ell}_{\mathrm{base},i}\in\mathbb{R}^{K}$ the corresponding base logits, and $y_i\in \{0,\ldots,K-1\}$ the ground-truth (GT) class. With label-smoothing parameter $\varepsilon_{\mathrm{ls}}\in[0,1)$, define $q_{i,k} = (1-\varepsilon_{\mathrm{ls}}) \mathbb{I}[k=y_i] + \frac{\varepsilon_{\mathrm{ls}}}{K}$ and $\mathcal{L}_{\mathrm{final}} = -\frac{1}{B} \sum_{i=1}^{B} \sum_{k=0}^{K-1} q_{i,k} \log \operatorname{softmax} (\boldsymbol{\ell}_{i}){k}$ and $\mathcal{L}_{\mathrm{base}} = -\frac{1}{B} \sum_{i=1}^{B} \sum_{k=0}^{K-1} q_{i,k} \log \operatorname{softmax} (\boldsymbol{\ell}_{\mathrm{base},i}){k}$.

The trajectory-classification loss $\mathcal{L}_{\mathrm{traj\text{-}cls}}$ is defined analogously using the trajectory logits $\boldsymbol{\ell}_{\mathrm{traj},i}$ as $\mathcal{L}_{\mathrm{traj\text{-}cls}} = -\frac{1}{B} \sum_{i=1}^{B} \sum_{k=0}^{K-1} q_{i,k} \log \operatorname{softmax} (\boldsymbol{\ell}_{\mathrm{traj},i})_{k}$. 

The trajectory-consistency loss encourages the trajectory branch to remain consistent with the detached base prediction. Let $\mathbf{p}_{\mathrm{base}} = \operatorname{softmax} \left(\operatorname{sg}(\boldsymbol{\ell}_{\mathrm{base}})\right)$ and $\log \mathbf{p}_{\mathrm{traj}} = \log\operatorname{softmax} (\boldsymbol{\ell}_{\mathrm{traj}})$. Then $\mathcal{L}_{\mathrm{traj\text{-}cons}} = \frac{1}{B} \sum_{i=1}^{B} \sum_{k=0}^{K-1} p_{\mathrm{base},i,k} \log \frac{p_{\mathrm{base},i,k}}{p_{\mathrm{traj},i,k}}$.

For the stochastic latent rollout, let $\widehat{\mathbf{Z}}\in\mathbb{R}^{B\times T\times D}$ denote the detached rollout targets, let $\boldsymbol{\mu}\in\mathbb{R}^{B\times T\times D}$ denote the predicted Gaussian means, and let $\mathbf{V}\in\mathbb{R}^{B\times T\times D}$ denote the predicted log-variances. The Gaussian negative log-likelihood is

\begin{equation}
    \mathcal{L}_{\mathrm{NLL}} = \frac{1}{2} \operatorname{mean} \left[\mathbf{V} + \frac{\left(\widehat{\mathbf{Z}} - \boldsymbol{\mu}\right)^{2}}{\exp(\mathbf{V})} \right].
\end{equation}

The rollout smoothness term penalizes abrupt changes between consecutive latent states $\mathcal{L}_{\mathrm{smooth\text{-}roll}} = \operatorname{mean} \left[\left| \mathbf{Z}_{\mathrm{roll}}[:,1:,:] - \mathbf{Z}_{\mathrm{roll}}[:,:-1,:] \right| \right]$, where $\mathbf{Z}_{\mathrm{roll}}\in\mathbb{R}^{B\times(T+1)\times D}$ denotes the latent rollout trajectory. The complete environment rollout loss is $\mathcal{L}_{\mathrm{env}} = \mathcal{L}_{\mathrm{NLL}} + \lambda_{\mathrm{smooth\text{-}roll}} \mathcal{L}_{\mathrm{smooth\text{-}roll}}$, where $\lambda_{\mathrm{smooth\text{-}roll}}\geq0$ is a hyperparameter rather than a fixed constant. The latent Gaussian regularization term is the KL divergence between the predicted Gaussian transition distribution and a standard normal prior $\mathcal{L}_{\mathrm{KL}} = -\frac{1}{2} \operatorname{mean} \left[ 1 + \mathbf{V} - \boldsymbol{\mu}^{2} - \exp(\mathbf{V})\right]$.

Finally, the spectral-angle, spectral-smoothness, and spatial-consistency terms are defined from the corresponding model outputs. Let $\widehat{\mathbf{x}}_{\lambda}$ denote the spectral reconstruction. The spectral-angle loss is computed on the center spatial position of each reconstructed patch:
\begin{equation}
    \mathcal{L}_{\mathrm{SAM}} = \frac{1}{B} \sum_{i=1}^{B} \arccos \left(\frac{\widehat{\mathbf{v}}_{i}^{\top}\mathbf{v}_{i}}{ |\widehat{\mathbf{v}}_{i}|^{2} |\mathbf{v}_{i}|^{2} + \epsilon_{\mathrm{SAM}}}\right),
\end{equation}
where $\widehat{\mathbf{v}}_{i},\mathbf{v}_{i}\in\mathbb{R}^{C}$ are respectively the reconstructed and true center-pixel spectral vectors and $\epsilon_{\mathrm{SAM}}>0$ is a numerical-stability parameter. The spectral smoothness loss penalizes changes between adjacent reconstructed spectral bands $\mathcal{L}_{\mathrm{smooth}} = \operatorname{mean} \left[\left| \widehat{\mathbf{x}}_{\lambda}[:,2,:] - \widehat{\mathbf{x}}_{\lambda}[:,1,:] \right| \right]$.

For a batch containing spatially neighboring labeled pixels, let $\mathbf{p}_{i} = \operatorname{softmax}(\boldsymbol{\ell}_{i})$ and let $\mathcal{P}$ denote the set of sample pairs whose spatial coordinates are within the prescribed local neighborhood. The spatial-consistency loss is $\mathcal{L}_{\mathrm{spat\text{-}cons}} = \frac{1}{|\mathcal{P}|} \sum_{(i,j)\in\mathcal{P}} \left| \mathbf{p}_{i}-\mathbf{p}_{j} \right|^{2}$. When no valid neighboring pair is present in the mini-batch, $\mathcal{L}_{\mathrm{spat\text{-}cons}}$ is set to zero.

\section{Experimental Setup}
\label{sec:exp_setup}

The proposed LWM framework evaluates trajectory-aware spatial-spectral representation learning for HSIC in a within-scene, label-scarce setting, where limited labeled samples from a fixed hyperspectral scene are used to learn structured latent representations for classifying held-out samples from the same scene.

\noindent\textbf{Datasets.} We evaluate the proposed LWM framework on five publicly available HSI benchmarks: WHU-HongHu (HH) \cite{zhong2020whu}, Tangdaowan (TD) \cite{FU2023115}, DFC18 (training set only) \cite{8727489}, a subset of HyRANK e.g., Dioni, and Loukia \cite{karantzalos_2018_1222202}. These datasets encompass a broad range of scene types, spatial resolutions, spectral characteristics, and classification complexities, providing a comprehensive evaluation of the proposed framework across agricultural, urban, coastal, and mixed land-cover environments.

\noindent\textbf{Implementation Details.} The proposed LWM model is optimized using AdamW with an initial learning rate of $2\times10^{-4}$ and weight decay of $7\times10^{-4}$. A cosine annealing learning-rate scheduler is employed over 100 training epochs, while early stopping with a patience of 15 epochs is used to prevent overfitting. Gradient clipping with a maximum norm of 1.0 is applied for training stability, and mixed-precision training is utilized to improve computational efficiency. For all datasets, hyperspectral cubes are normalized using per-band standardization. Local spatial neighborhoods are extracted using mirror padding with a fixed patch size of $25\times25$. Unless otherwise specified, the labeled samples are divided into training, validation, and testing subsets using stratified sampling, with ratios of 2\%, 2\%, and 96\% for the Dioni, HH, and TD datasets, and 10\%, 10\%, and 80\% for the DFC18 and Loukia datasets, respectively. Each selected pixel serves as the center of its corresponding spatial-spectral patch; thus, the sampling ratios refer to labeled center pixels. Unless otherwise stated, a batch size of 64 is used, and label smoothing is set to 0.1.

The transformer backbone employs an embedding dimension of $D=128$, with spectral, spatial, and fusion transformer depths set to 2. Each transformer layer uses 8 attention heads, an MLP expansion ratio of 4.0, dropout of 0.1, and stochastic depth with a maximum drop-path rate of 0.05. During training, random spectral-band masking and spatial-pixel masking are applied with masking ratios of 25\% and 15\%, respectively. For LWM module, the rollout horizon is fixed to $T=3$. The stochastic transition model predicts Gaussian latent distributions with log-variance constrained to the interval $[a,b]$, where the lower and upper bounds are fixed as $a=-6$ and $b=2$. Rollout states are refined through a GRU-based latent transition module and aggregated using the proposed rollout planner. The latent state dimension is set equal to the transformer embedding dimension ($D=128$). The overall objective combines classification, reconstruction, latent dynamics, rollout regularization, trajectory supervision, and physically motivated constraints. Unless otherwise specified, the corresponding weighting coefficients are set to $\lambda_{\mathrm{spec\text{-}rec}}=0.25$, $\lambda_{\mathrm{spat\text{-}rec}}=0.25$, $\lambda_{\mathrm{spec\text{-}dyn}}=0.20$, $\lambda_{\mathrm{spat\text{-}dyn}}=0.20$, $\lambda_{\mathrm{roll}}=0.25$, $\lambda_{\mathrm{KL}}=0.001$, $\lambda_{\mathrm{traj\text{-}cons}}=0.10$, $\lambda_{\mathrm{traj\text{-}cls}}=0.25$, $\lambda_{\mathrm{SAM}}=0.08$, $\lambda_{\mathrm{smooth}}=0.05$, and $\lambda_{\mathrm{spat\text{-}cons}}=0.05$. These hyperparameters are optimized using Bayesian optimization and kept fixed for each dataset unless otherwise specified.

\noindent\textbf{Evaluation Metrics.} The classification performance is evaluated using Overall Accuracy (OA), Average Accuracy (AA), and Cohen's Kappa coefficient ($\kappa$), which are the standard metrics for HSIC. To provide a more comprehensive assessment, we additionally report the Macro-F1 score and Matthews Correlation Coefficient (MCC). Macro-F1 evaluates the harmonic mean of precision and recall averaged across all classes, whereas MCC provides a balanced measure of classification quality even in highly imbalanced scenarios.

Model calibration is assessed using the Expected Calibration Error (ECE) and multiclass Brier Score \cite{NEURIPS20223915a87d, 9879475}. ECE measures the discrepancy between prediction confidence and empirical accuracy, while the Brier Score quantifies the mean squared error between predicted probabilities and GT labels. Lower values indicate better-calibrated probabilistic predictions. To evaluate the contribution of latent rollout reasoning, branch-wise analysis is performed by separately measuring the performance of the base classifier, trajectory classifier, and their fused prediction. In addition to OA, AA, and Kappa, we report the correction rate (CR), harm rate (HR), and net gain (NG). The CR measures the proportion of samples incorrectly classified by the base branch but correctly classified by the trajectory branch, whereas the HR measures the opposite case. Their difference, referred to as NG, quantifies the overall benefit introduced by latent rollout reasoning.
Log Loss measures the quality of the predicted class probabilities by computing the average negative log-probability assigned to the GT class. Top3 accuracy measures the proportion of test samples for which the GT class appears among the three classes with the highest predicted probabilities.


\section{Latent Rollout Contribution}
\label{sec:branch}

To investigate the contribution of latent rollout reasoning, we analyze the performance of the individual prediction branches and their fusion. Specifically, the proposed framework produces three predictions: (i) the \emph{Base} branch, which relies solely on the fused transformer representation; (ii) the \emph{Trajectory} branch, which performs classification using the planned latent state obtained from stochastic rollout reasoning; and (iii) the \emph{Final} prediction, which combines both branches through logit fusion. This analysis provides insight into whether latent rollout modeling contributes complementary information beyond static spatial-spectral encoding. Table~\ref{tab:Branch} reports the branch-wise classification results. The performance difference between an individual branch and the final prediction is quantified as $\Delta = \mathrm{OA}_{\mathrm{Final}} - \mathrm{OA}_{\mathrm{Branch}}$, where $\mathrm{OA}_{\mathrm{Branch}}$ denotes the OA of either the Base or Trajectory branch. A positive $\Delta$ indicates that fusion improves upon the corresponding individual predictor.

\begin{table}[!hbt]
\centering
\caption{Branch-wise performance of the Base, Trajectory, and Final prediction branches. $\Delta$ denotes the OA improvement of the final fused prediction over the corresponding branch. Best results are in boldface.}
    \begin{tabular}{lccccc} \toprule
        \multicolumn{2}{c}{\textbf{Branch}} & OA & AA & $\kappa$ & $\Delta$ \\ \midrule

        \multirow{3}{*}{\rotatebox{90}{\textbf{DFC18}}}
        & Base & 0.9100 & 0.9198 & 0.8858 & +0.0149 \\
        & Trajectory & 0.9113 & 0.9273 & 0.8976 & +0.0136 \\
        & Final & \textbf{0.9249} & \textbf{0.9352} & \textbf{0.9021} & -- \\ \midrule

        \multirow{3}{*}{\rotatebox{90}{\textbf{Dioni}}}
        & Base & 0.8648 & 0.8835 & 0.8343 & +0.0523 \\
        & Trajectory & 0.9037 & 0.9208 & 0.8837 & +0.0134 \\
        & Final & \textbf{0.9171} & \textbf{0.9346} & \textbf{0.8979} & -- \\ \midrule

        \multirow{3}{*}{\rotatebox{90}{\textbf{Loukia}}}
        & Base & 0.8470 & 0.8596 & 0.8207 & +0.0437 \\
        & Trajectory & 0.8811 & 0.8678 & 0.8613 & +0.0096 \\
        & Final & \textbf{0.8907} & \textbf{0.8775} & \textbf{0.8709} & -- \\ \midrule

        \multirow{3}{*}{\rotatebox{90}{\textbf{HH}}}
        & Base & 0.9703 & 0.9643 & 0.9677 & +0.0211 \\
        & Trajectory & 0.9810 & 0.9763 & 0.9787 & +0.0104 \\
        & Final & \textbf{0.9914} & \textbf{0.9864} & \textbf{0.9891} & -- \\ \midrule
        
        \multirow{3}{*}{\rotatebox{90}{\textbf{TD}}}
        & Base & 0.9852 & 0.9859 & 0.9846 & +0.0100 \\
        & Trajectory & 0.9946 & 0.9953 & 0.9939 & +0.0006 \\
        & Final & \textbf{0.9952} & \textbf{0.9958} & \textbf{0.9945} & -- \\ \bottomrule

\end{tabular}
\label{tab:Branch}
\end{table}

Table~\ref{tab:Branch} shows that the Trajectory branch consistently outperforms the Base branch across all evaluated datasets. This observation indicates that latent rollout reasoning captures predictive structure that is not fully represented by static transformer features alone. The improvement is particularly pronounced on Dioni and Loukia, where the trajectory branch increases OA from 86.48\% to 90.37\% and from 84.70\% to 88.11\%, respectively. These gains suggest that stochastic latent evolution is especially beneficial for scenes exhibiting large intra-class variability and complex spatial organization. The fused prediction consistently achieves the highest performance. The largest fusion gain is observed on Dioni, where the final model improves OA by 5.23 percentage points relative to the Base branch, highlighting the importance of latent rollout reasoning in challenging remote sensing scenarios. Although the gains are smaller on HH due to its already saturated performance, the fused prediction still improves upon both individual branches.

Beyond branch-level accuracy, we further evaluate the complementarity between the Base and Trajectory branches. Let $y_i$ denote the GT label, $\hat{y}^{(b)}_i$ the prediction of the Base branch, and $\hat{y}^{(t)}_i$ the prediction of the Trajectory branch. The CR measures the proportion of samples that are misclassified by the Base branch but correctly classified by the Trajectory branch $\mathrm{CR} = \frac{1}{N_{\mathrm{test}}} \sum_{i=1}^{N_{\mathrm{test}}} \mathbb{I}\left(\hat{y}^{(b)}_i \neq y_i \,\land\, \hat{y}^{(t)}_i = y_i\right)$, where $\mathbb{I}(\cdot)$ is the indicator function and $N_{\mathrm{test}}$ is the number of test samples. Similarly, the HR measures the proportion of samples correctly classified by the Base branch but misclassified by the Trajectory branch $\mathrm{HR} = \frac{1}{N_{\mathrm{test}}} \sum_{i=1}^{N_{\mathrm{test}}} \mathbb{I}\left(\hat{y}^{(b)}_i = y_i \,\land\, \hat{y}^{(t)}_i \neq y_i\right)$.

The overall benefit of the trajectory branch is quantified through the NG as $\mathrm{NG} = \mathrm{CR} - \mathrm{HR}$. To estimate the maximum achievable benefit from combining both branches, we define the Oracle Accuracy $\mathrm{OA}_{\mathrm{oracle}} = \frac{1}{N_{\mathrm{test}}} \sum_{i=1}^{N_{\mathrm{test}}} \mathbb{I}\left(\hat{y}^{(b)}_i = y_i \,\lor\, \hat{y}^{(t)}_i = y_i\right)$.

The Agreement Rate (AGR) measures the fraction of samples on which both branches produce identical predictions $\mathrm{AGR} = \frac{1}{N_{\mathrm{test}}} \sum_{i=1}^{N_{\mathrm{test}}} \mathbb{I}\left(\hat{y}^{(b)}_i = \hat{y}^{(t)}_i\right)$.

Finally, the diversity between the predictive distributions of the two branches is measured using the Jensen-Shannon divergence (JSD) as $\mathrm{JSD} = \frac{1}{2} D_{\mathrm{KL}}\!\left(\mathbf{p}^{(b)} \,\|\, \mathbf{m}\right) + \frac{1}{2} D_{\mathrm{KL}}\!\left(\mathbf{p}^{(t)} \,\|\, \mathbf{m}\right)$, $\mathbf{m} = \frac{1}{2} \left(\mathbf{p}^{(b)} + \mathbf{p}^{(t)}\right)$ and the Kullback-Leibler divergence is defined as $D_{\mathrm{KL}}(\mathbf{p}\,\|\,\mathbf{q}) = \sum_{k=0}^{K-1} p_k \log \left(\frac{p_k}{q_k}\right)$, where, $\mathbf{p}^{(b)}$ and $\mathbf{p}^{(t)}$ denote the softmax probability vectors from the Base and Trajectory branches, respectively. A larger JSD indicates greater diversity between the Base and Trajectory predictions.

The correction analysis in Table~\ref{tab:traj_correction} provides a direct assessment of the complementary behavior between the Base and Trajectory branches. In most datasets, the CR exceeds the HR, resulting in positive NG values. The largest gains are observed on Dioni and Loukia, where the trajectory branch corrects 6.41\% and 6.37\% of the errors made by the Base branch while introducing only 1.51\% and 1.96\% additional errors, respectively. This yields NGs of 4.89\% and 4.41\%, demonstrating substantial complementary information from the rollout branch on these datasets. In contrast, for TD, the CR is 0.11\% while the HR is 0.18\%, resulting in a negative NG of $-0.06\%$. This indicates that the trajectory branch introduces slightly more additional errors than the number of Base-branch errors it corrects on this dataset.

\begin{table}[!hbt]
\centering
\caption{Trajectory branch correction analysis. Higher CR, NG, and OA$_{\mathrm{oracle}}$ indicate stronger complementary information provided by the latent rollout branch.}
\renewcommand{\arraystretch}{1.10}
    \resizebox{\columnwidth}{!}{\begin{tabular}{lcccccc}\toprule
    Dataset & CR (\%) & HR (\%) & NG (\%) &
    OA$_{\mathrm{oracle}}$ (\%) &
    AGR (\%) & JSD \\
    \midrule
    DFC18  & 1.76 & 1.25 & 0.51 & 93.67 & 96.31 & 0.0196 \\
    Dioni  & 6.41 & 1.51 & 4.89 & 92.89 & 90.86 & 0.0817 \\
    Loukia & 6.37 & 1.96 & 4.41 & 91.08 & 90.74 & 0.0433 \\
    HH     & 0.25 & 0.17 & 0.08 & 99.28 & 99.54 & 0.0033 \\
    TD     & 0.11 & 0.18 & -0.06 & 99.64 & 99.70 & 0.0017 \\
\bottomrule
\end{tabular}}
\label{tab:traj_correction}
\end{table}

The OA$_{\mathrm{oracle}}$ further reveals the potential upper bound obtainable from combining both branches, reaching 92.89\% on Dioni and 93.67\% on DFC18. Moreover, the JSD correlates strongly with the observed gains. Dioni exhibits the largest JSD (0.0817) and the largest NG, indicating that greater predictive diversity between branches translates into larger performance improvements after fusion. In contrast, HH exhibits very high AGR (99.54\%) and very small JSD (0.0033), indicating that both branches converge to nearly identical predictions on this near-saturated dataset.


\section{Robustness Analysis}
\label{sec:robustness}

A practical HSIC system should remain reliable under incomplete spectral observations and sensor noise. To evaluate the robustness of the proposed LWM framework, we consider two challenging perturbation scenarios: (i) random spectral band removal and (ii) additive Gaussian noise. For spectral corruption, a fraction $\rho$ of spectral bands is randomly removed and set to zero. Let $\mathbf{m}\in \{0,1\}^{C}$ be a random binary spectral mask satisfying $\Pr(m_c=0)=\rho$. The mask is expanded across the batch and spatial dimensions as $\widetilde{\mathbf{m}} = \mathbf{1}_{B} \otimes \mathbf{m} \otimes \mathbf{1}_{P} \otimes \mathbf{1}_{P} \in \{0,1\}^{B\times C\times P\times P}$, where $\mathbf{1}_{B}$ and $\mathbf{1}_{P}$ denote vectors of ones, and $\otimes$ denotes replication along the corresponding dimensions. The corrupted input is then defined as $\widetilde{\mathbf{x}} = \mathbf{x} \odot \widetilde{\mathbf{m}}.$. 

For noise robustness, zero-mean Gaussian noise is added to the input $\widetilde{\mathbf{x}} = \mathbf{x} + \boldsymbol{\eta},~\boldsymbol{\eta} \sim \mathcal{N} \left(\mathbf{0}, \sigma^2\mathbf{I} \right)$, where $\boldsymbol{\eta}\in\mathbb{R}^{B\times C\times P\times P}$ has the same dimensionality as $\mathbf{x}$. The experiments are performed without retraining the model. Performance is evaluated using OA, AA, and $\kappa$ under increasing corruption levels. The results are reported in Table~\ref{tab:robustness}.

\begin{table}[!hbt]
\centering
\caption{Robustness evaluation under spectral band removal and additive Gaussian noise.}
\renewcommand{\arraystretch}{1.12}
    \begin{tabular}{llcccc} \toprule
    \textbf{Dataset} & \textbf{Perturbation} & \textbf{Level} & \textbf{OA} & \textbf{AA} & $\boldsymbol{\kappa}$ \\ \midrule

    \multirow{6}{*}{DFC18}
    & Missing Bands & 10\% & 0.9132 & 0.9341 & 0.8872 \\
    & Missing Bands & 20\% & 0.8752 & 0.9130 & 0.8389 \\
    & Missing Bands & 30\% & 0.7925 & 0.8514 & 0.7355 \\
    & Gaussian Noise & $\sigma=0.01$ & 0.9249 & 0.9352 & 0.9020 \\
    & Gaussian Noise & $\sigma=0.03$ & 0.9247 & 0.9351 & 0.9018 \\
    & Gaussian Noise & $\sigma=0.05$ & 0.9242 & 0.9344 & 0.9012 \\ \midrule

    \multirow{6}{*}{Dioni}
    & Missing Bands & 10\% & 0.9167 & 0.9092 & 0.8967 \\
    & Missing Bands & 20\% & 0.8928 & 0.8662 & 0.8661 \\
    & Missing Bands & 30\% & 0.8372 & 0.7869 & 0.7937 \\
    & Gaussian Noise & $\sigma=0.01$ & 0.9170 & 0.9346 & 0.8978 \\
    & Gaussian Noise & $\sigma=0.03$ & 0.9168 & 0.9345 & 0.8975 \\
    & Gaussian Noise & $\sigma=0.05$ & 0.9165 & 0.9343 & 0.8972 \\ \midrule

    \multirow{6}{*}{Loukia}
    & Missing Bands & 10\% & 0.8876 & 0.8666 & 0.8669 \\
    & Missing Bands & 20\% & 0.8717 & 0.8435 & 0.8476 \\
    & Missing Bands & 30\% & 0.8442 & 0.8012 & 0.8143 \\
    & Gaussian Noise & $\sigma=0.01$ & 0.8903 & 0.8770 & 0.8704 \\
    & Gaussian Noise & $\sigma=0.03$ & 0.8911 & 0.8782 & 0.8714 \\
    & Gaussian Noise & $\sigma=0.05$ & 0.8905 & 0.8779 & 0.8707 \\ \midrule

    \multirow{6}{*}{HH}
    & Missing Bands & 10\% & 0.9917 & 0.9860 & 0.9895 \\
    & Missing Bands & 20\% & 0.9887 & 0.9792 & 0.9857 \\
    & Missing Bands & 30\% & 0.9686 & 0.9476 & 0.9603 \\
    & Gaussian Noise & $\sigma=0.01$ & 0.9914 & 0.9865 & 0.9892 \\
    & Gaussian Noise & $\sigma=0.03$ & 0.9914 & 0.9865 & 0.9892 \\
    & Gaussian Noise & $\sigma=0.05$ & 0.9914 & 0.9864 & 0.9892 \\ \midrule
    
    \multirow{6}{*}{TD}
    & Missing Bands & 10\% & 0.9925 & 0.9940 & 0.9915 \\
    & Missing Bands & 20\% & 0.9802 & 0.9838 & 0.9774 \\
    & Missing Bands & 30\% & 0.9441 & 0.9499 & 0.9364 \\
    & Gaussian Noise & $\sigma=0.01$ & 0.9952 & 0.9958 & 0.9946 \\
    & Gaussian Noise & $\sigma=0.03$ & 0.9952 & 0.9958 & 0.9945 \\
    & Gaussian Noise & $\sigma=0.05$ & 0.9952 & 0.9958 & 0.9945 \\\bottomrule
\end{tabular}
\label{tab:robustness}
\end{table}

The robustness results reveal several important characteristics of the proposed LWM framework. First, LWM exhibits strong resilience to additive Gaussian noise across all datasets. Even at $\sigma=0.05$, the degradation in OA remains negligible. For example, on DFC18, the OA decreases from 92.49\% to 92.42\%, while on HH, the performance remains virtually unchanged. This behavior indicates that the learned latent representations are stable under moderate perturbations and are not overly dependent on individual spectral measurements.

Second, performance degrades more gradually under spectral band removal than would typically be expected for models relying heavily on individual wavelengths. Although accuracy decreases as the fraction of removed bands increases, the model retains strong predictive capability even when 30\% of the spectral information is discarded. On Dioni and Loukia, the OA remains above 83\%, while HH retains an OA of 96.86\%. These results suggest that the latent rollout mechanism learns distributed spatial-spectral representations that remain informative even when substantial portions of the spectrum are unavailable.

Third, the robustness trends are consistent with the design of LWM. The masked reconstruction objectives encourage recovery of missing information from surrounding spectral and spatial context, while the spectral and spatial dynamics constraints enforce structured relationships between latent representations. In addition, stochastic latent rollout reasoning promotes representation consistency across multiple plausible latent states rather than relying on a single deterministic embedding. Consequently, the model becomes less sensitive to localized corruption and measurement noise.

\section{Attention-based Interpretability Analysis}
\label{sec:attention}

\begin{figure}[!hbt]
    \centering
    \begin{subfigure}{0.23\textwidth}
	   \includegraphics[width=0.99\textwidth]{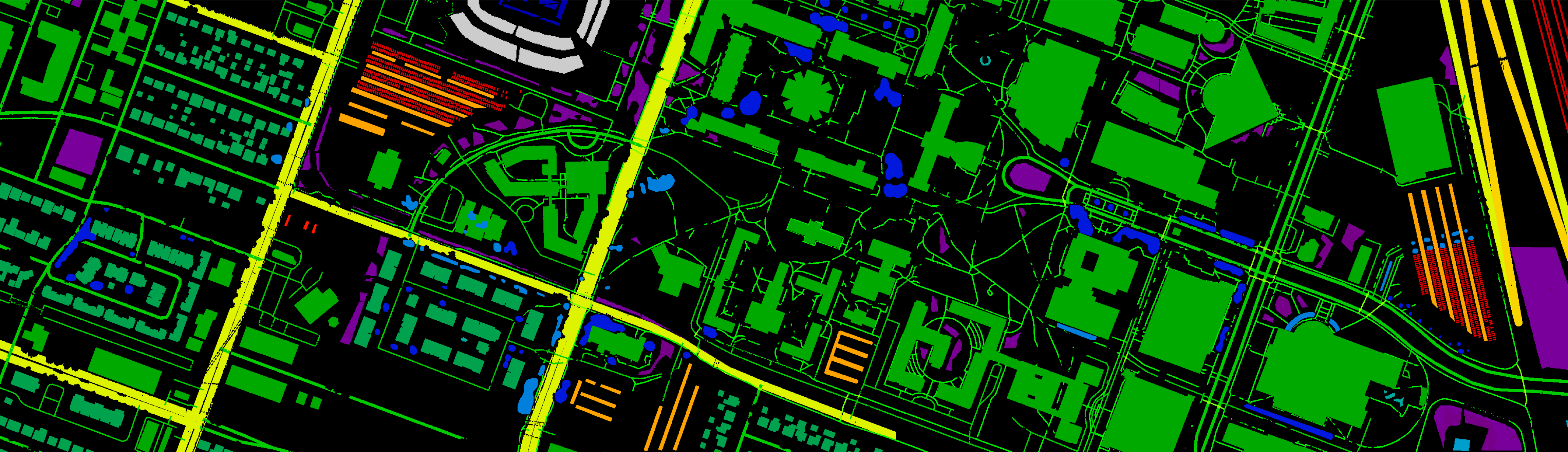}
	\caption{GT}
    \end{subfigure}
    \begin{subfigure}{0.23\textwidth}
	   \includegraphics[width=0.99\textwidth]{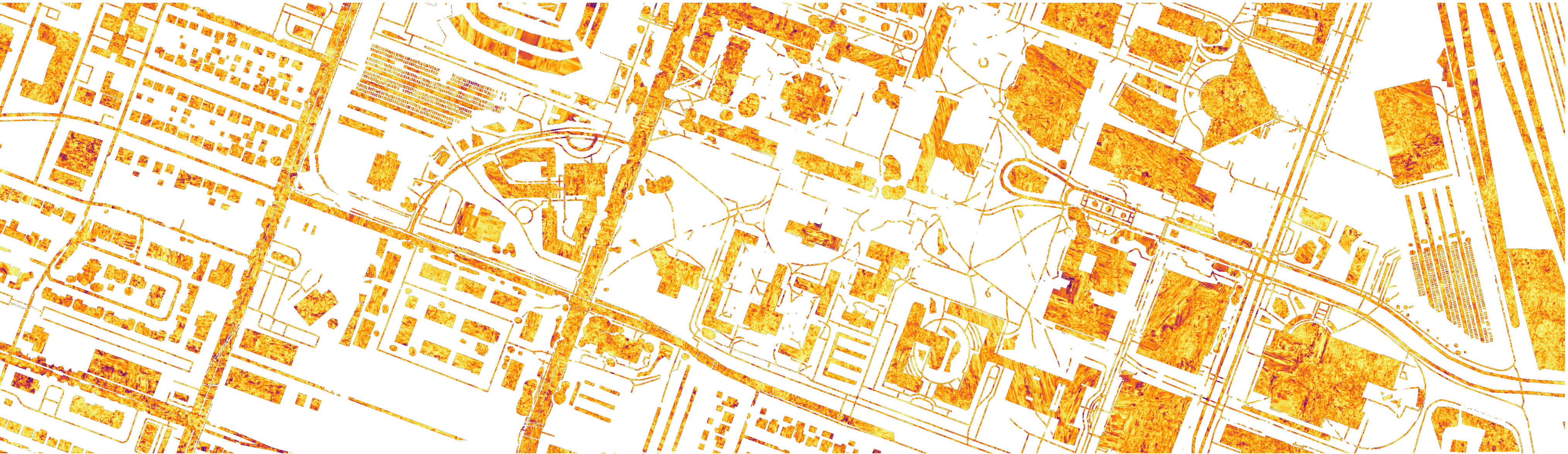}
	\caption{Attention Heatmap} 
    \end{subfigure}
    \begin{subfigure}{0.23\textwidth}
	   \includegraphics[width=0.99\textwidth]{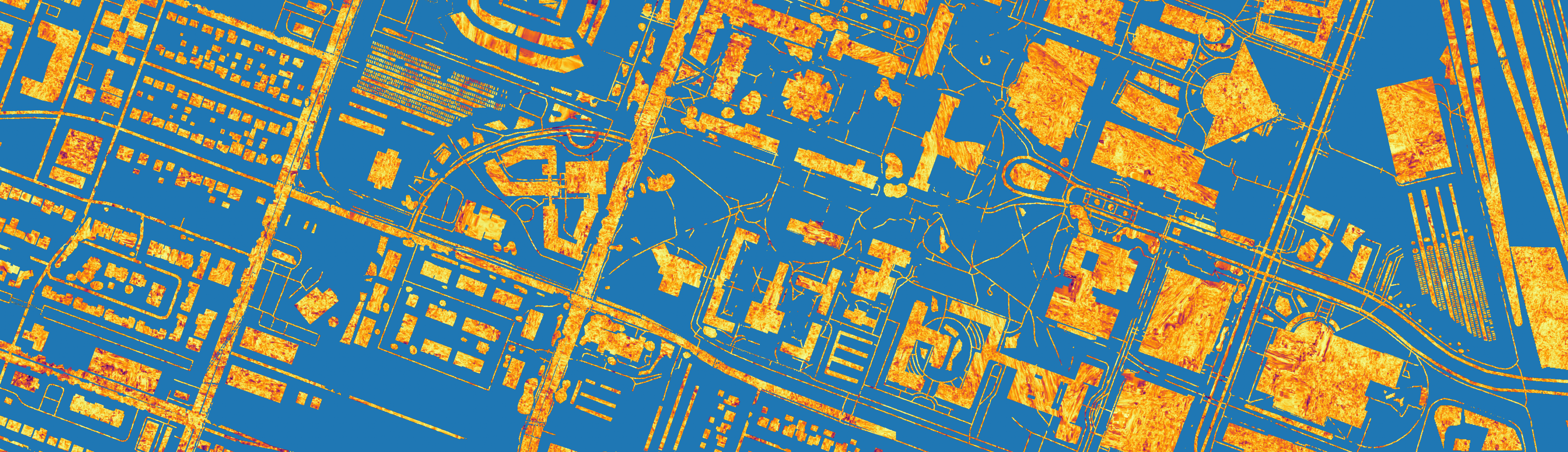}
	\caption{Attention Overlay} 
    \end{subfigure}
    \begin{subfigure}{0.23\textwidth}
	   \includegraphics[width=0.99\textwidth]{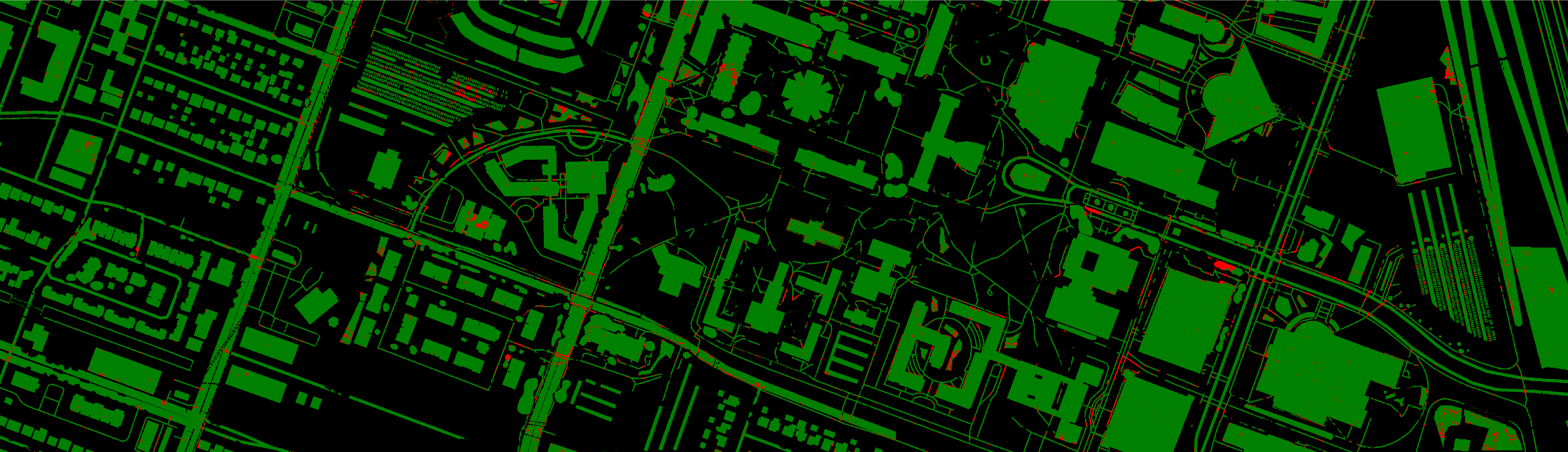}
	\caption{Error Map} 
    \end{subfigure}
    \caption{Attention-based interpretability analysis on the DFC18 dataset. (a) GT. (b) Spatial distribution of the learned attention responses. (c) Attention responses overlaid on GT, illustrating the regions that contribute most strongly to the classification decision. (d) Error map, where green pixels indicate correctly classified samples and red pixels indicate misclassified samples.}
    \label{fig:DFC18Attention}
\end{figure}

\begin{figure}[!hbt]
    \centering
    \begin{subfigure}{0.23\textwidth}
	   \includegraphics[width=0.99\textwidth]{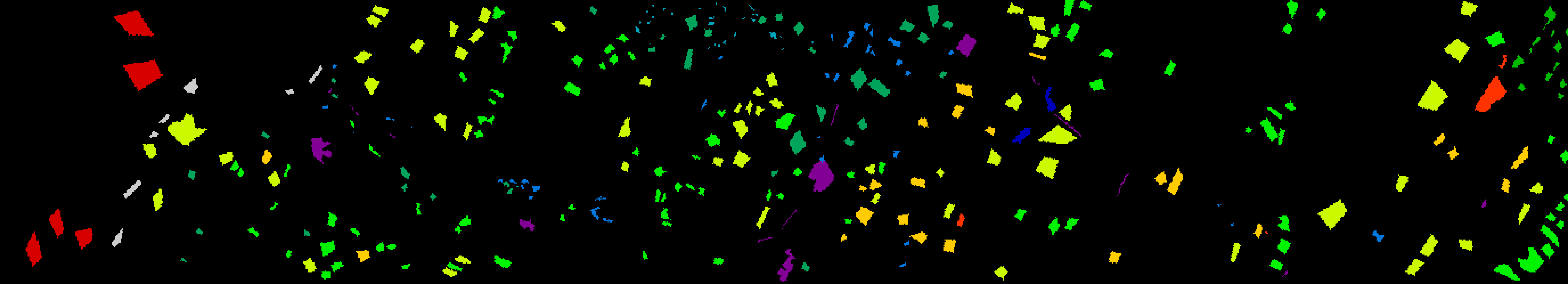}
	\caption{GT}
    \end{subfigure}
    \begin{subfigure}{0.23\textwidth}
	   \includegraphics[width=0.99\textwidth]{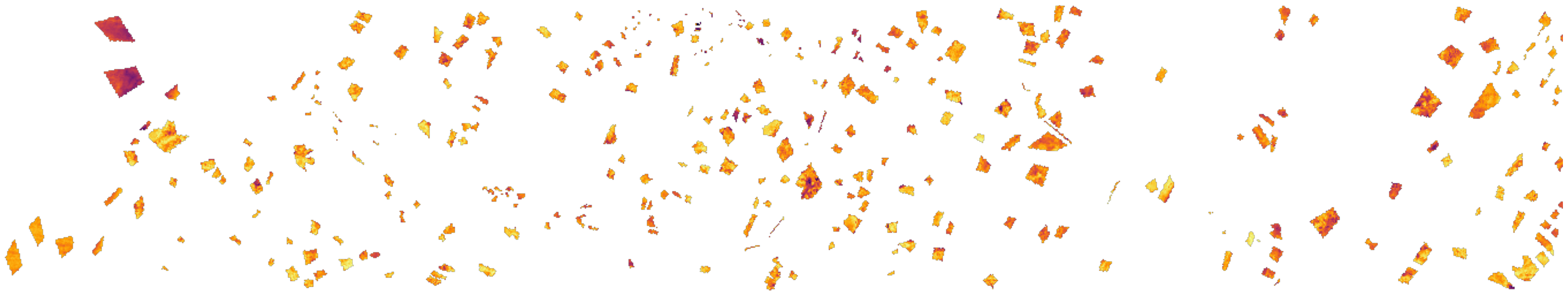}
	\caption{Attention Heatmap} 
    \end{subfigure}
    \begin{subfigure}{0.23\textwidth}
	   \includegraphics[width=0.99\textwidth]{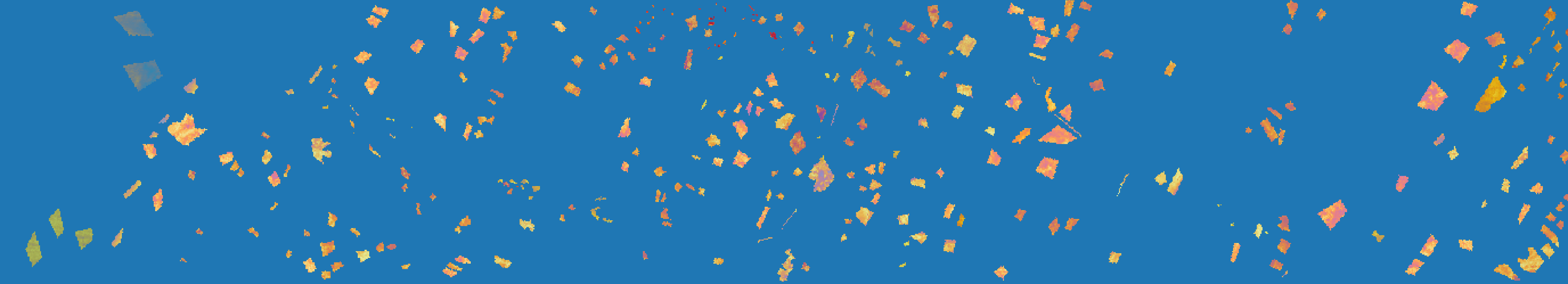}
	\caption{Attention Overlay} 
    \end{subfigure}
    \begin{subfigure}{0.23\textwidth}
	   \includegraphics[width=0.99\textwidth]{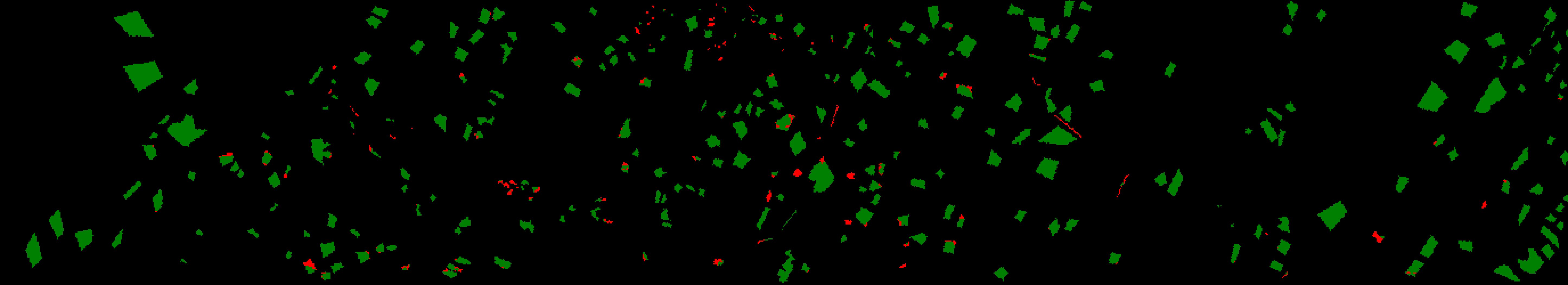}
	\caption{Error Map} 
    \end{subfigure}
    \caption{Attention-based analysis on the Dioni dataset.}
    \label{fig:DioniAttention}
\end{figure}

\begin{figure}[!hbt]
    \centering
    \begin{subfigure}{0.23\textwidth}
	   \includegraphics[width=0.99\textwidth]{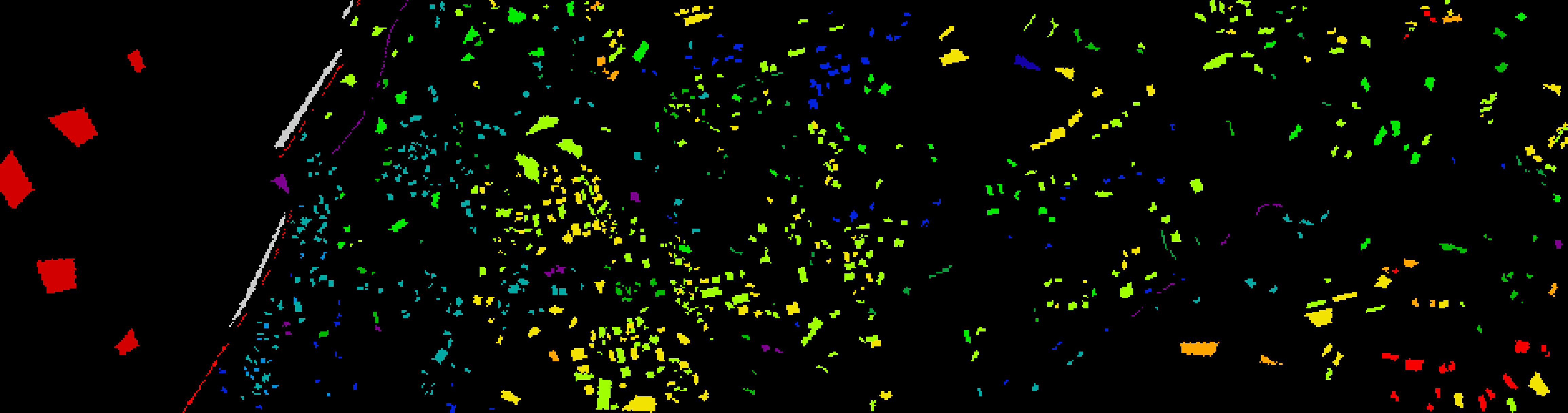}
	\caption{GT}
    \end{subfigure}
    \begin{subfigure}{0.23\textwidth}
	   \includegraphics[width=0.99\textwidth]{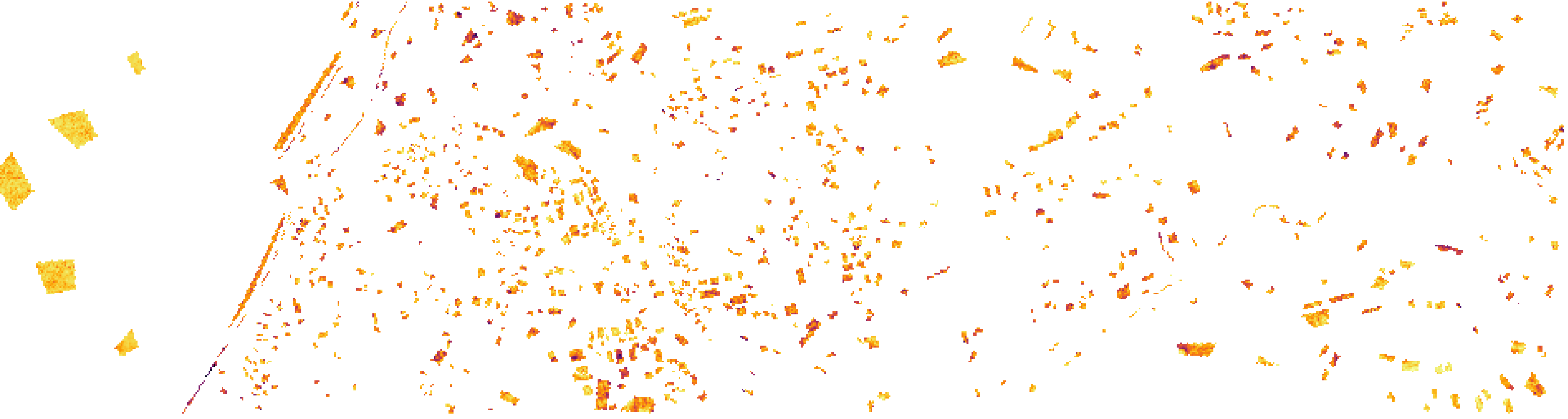}
	\caption{Attention Heatmap} 
    \end{subfigure}
    \begin{subfigure}{0.23\textwidth}
	   \includegraphics[width=0.99\textwidth]{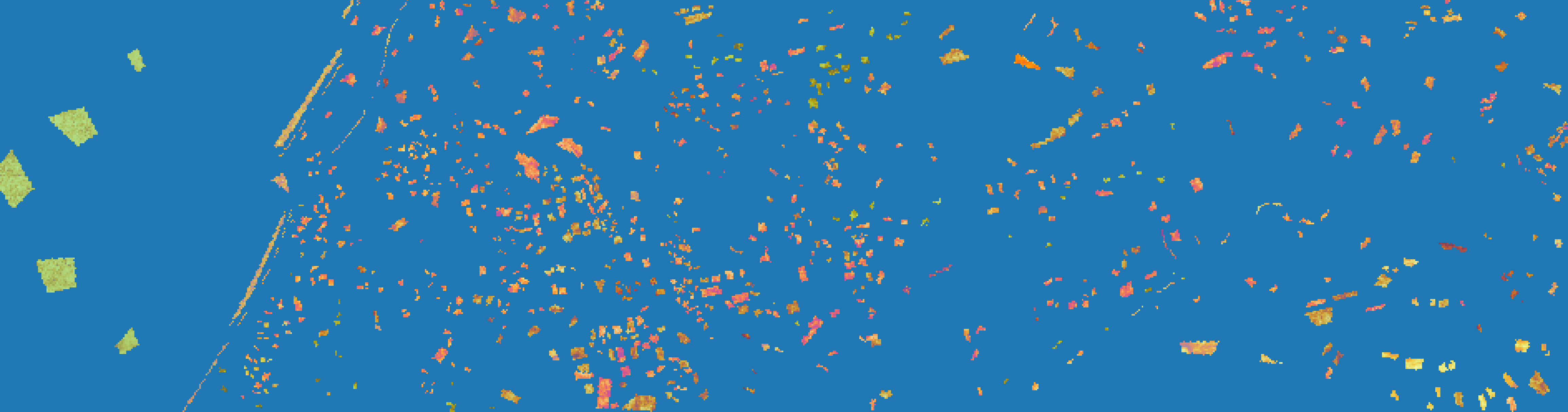}
	\caption{Attention Overlay} 
    \end{subfigure}
    \begin{subfigure}{0.23\textwidth}
	   \includegraphics[width=0.99\textwidth]{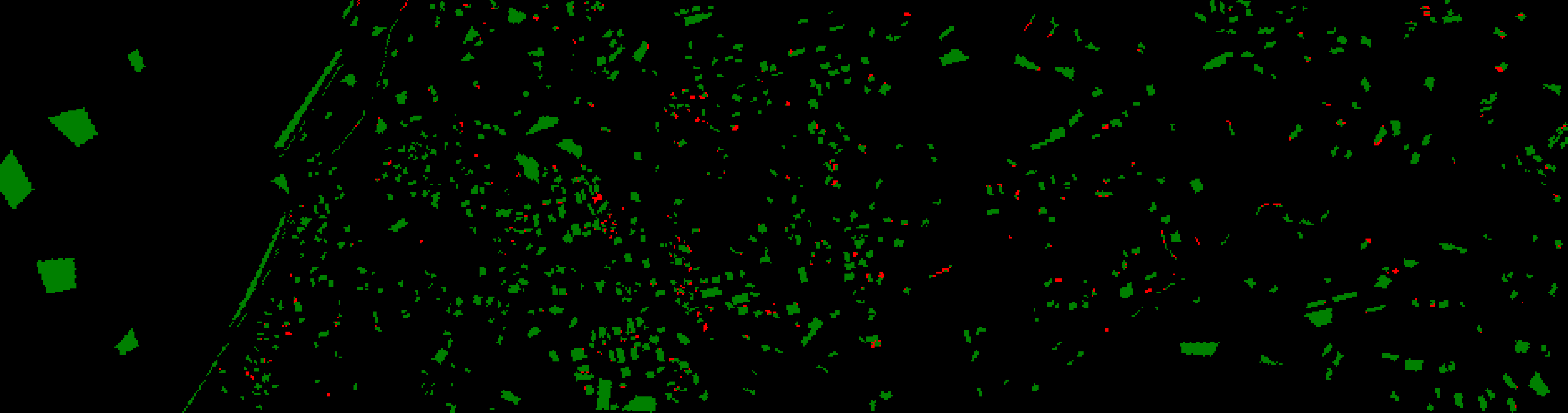}
	\caption{Error Map} 
    \end{subfigure}
    \caption{Attention-based analysis on the Loukia dataset.}
    \label{fig:LoukiaAttention}
\end{figure}

\begin{figure}[!hbt]
    \centering
    \begin{subfigure}{0.11\textwidth}
	   \includegraphics[width=0.99\textwidth]{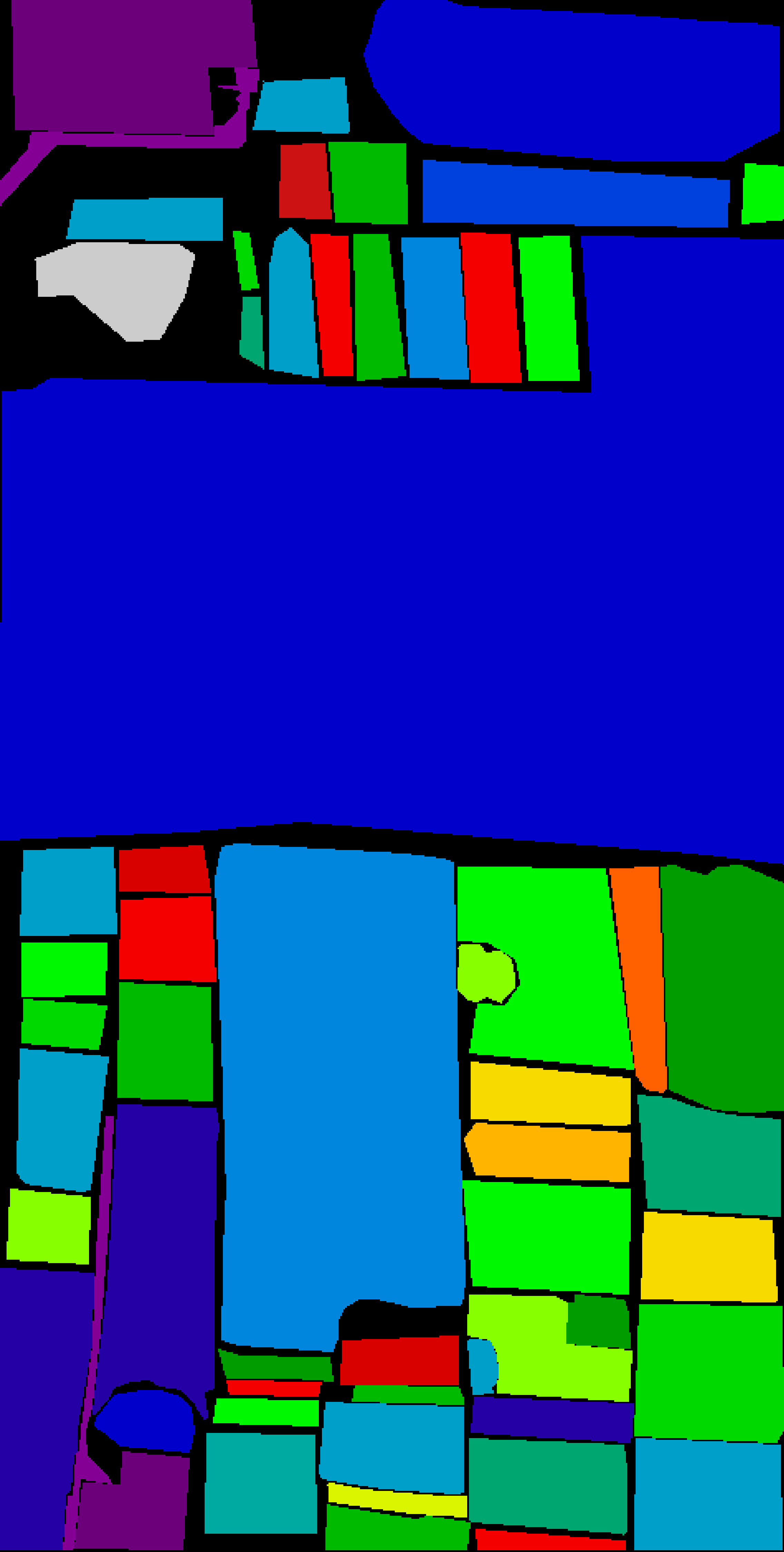}
	\caption{GT}
    \end{subfigure}
    \begin{subfigure}{0.14\textwidth}
	   \includegraphics[width=0.99\textwidth]{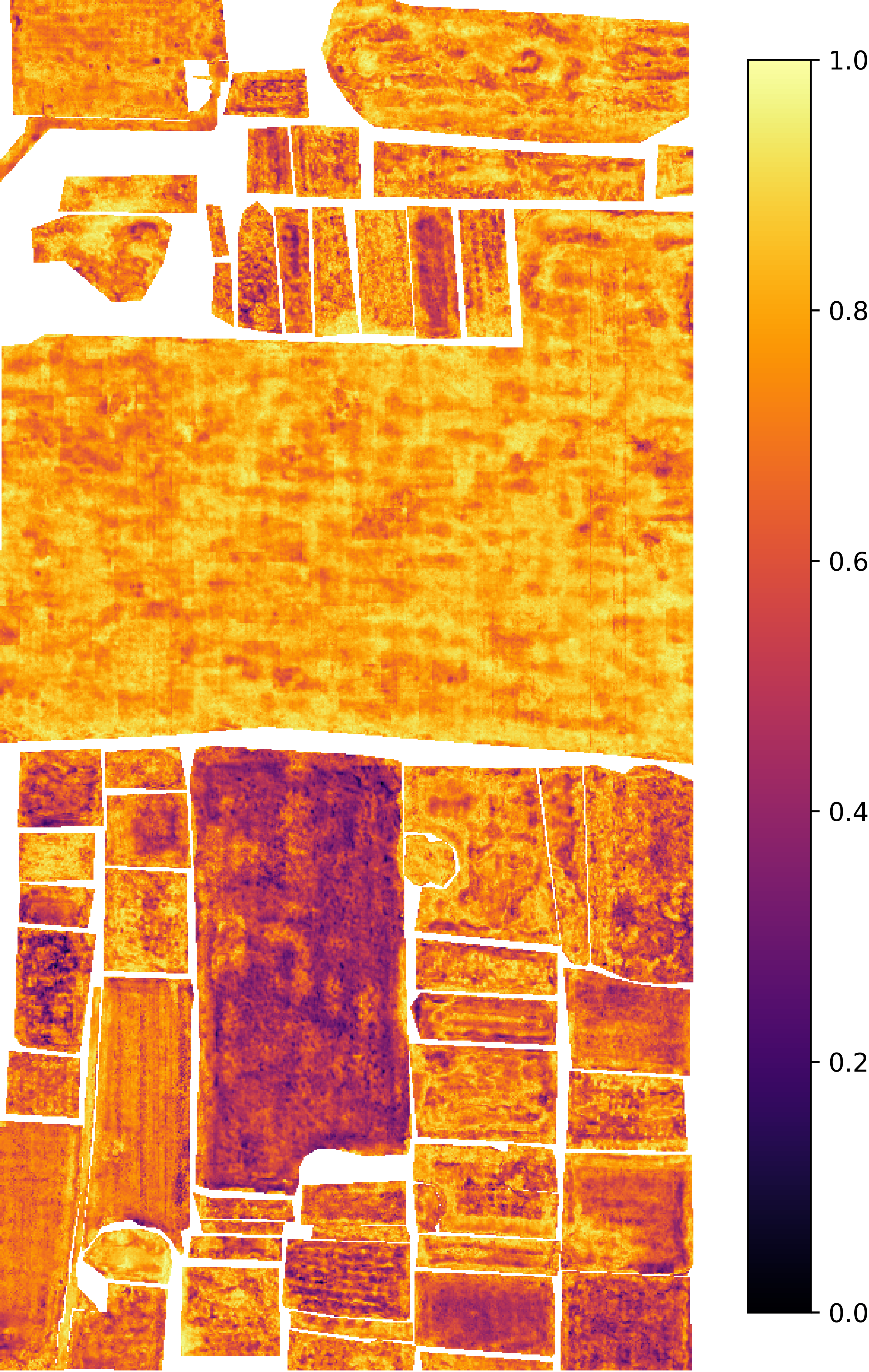}
	\caption{Heatmap} 
    \end{subfigure}
    \begin{subfigure}{0.11\textwidth}
	   \includegraphics[width=0.99\textwidth]{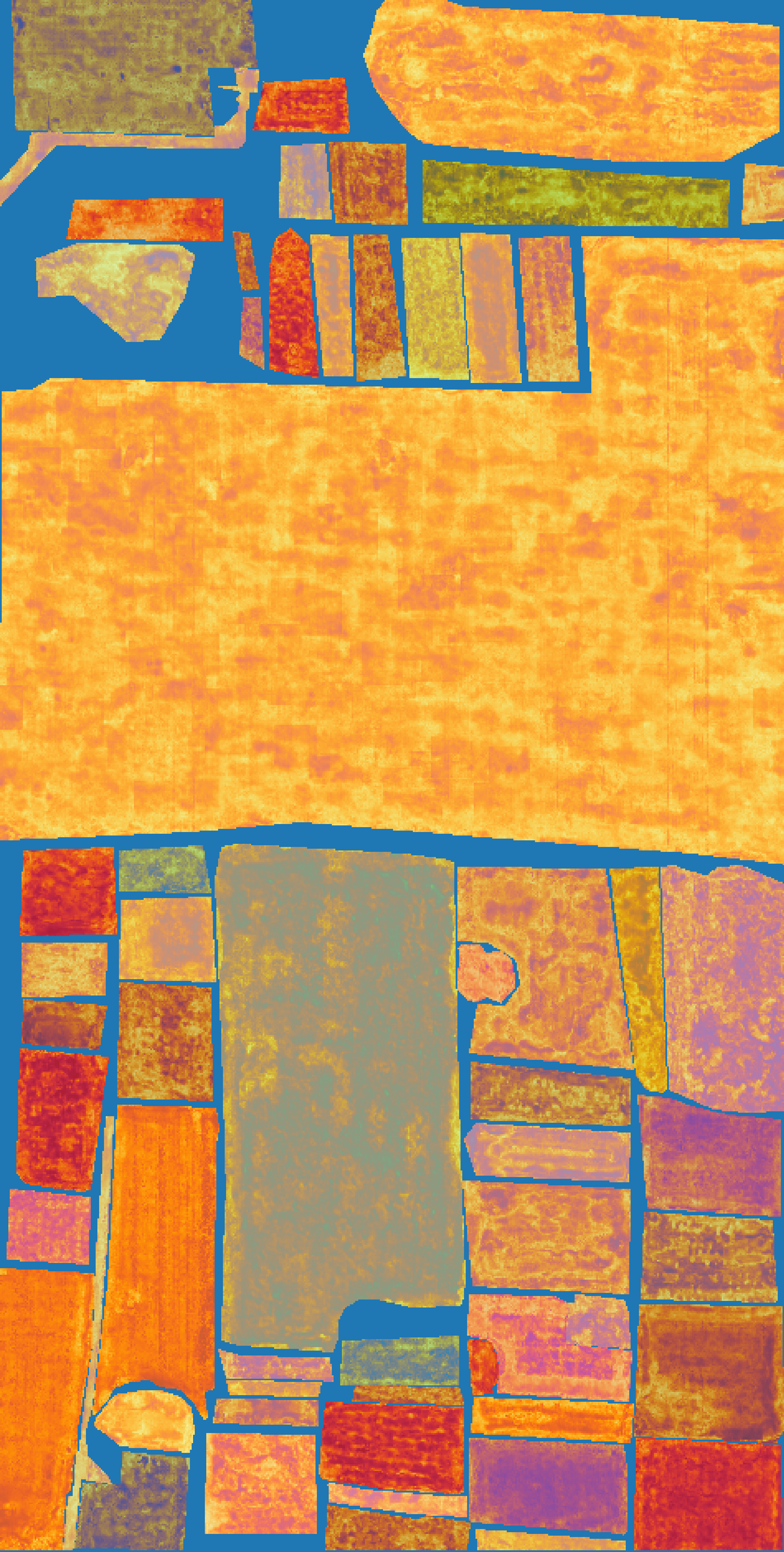}
	\caption{Overlay} 
    \end{subfigure}
    \begin{subfigure}{0.11\textwidth}
	   \includegraphics[width=0.99\textwidth]{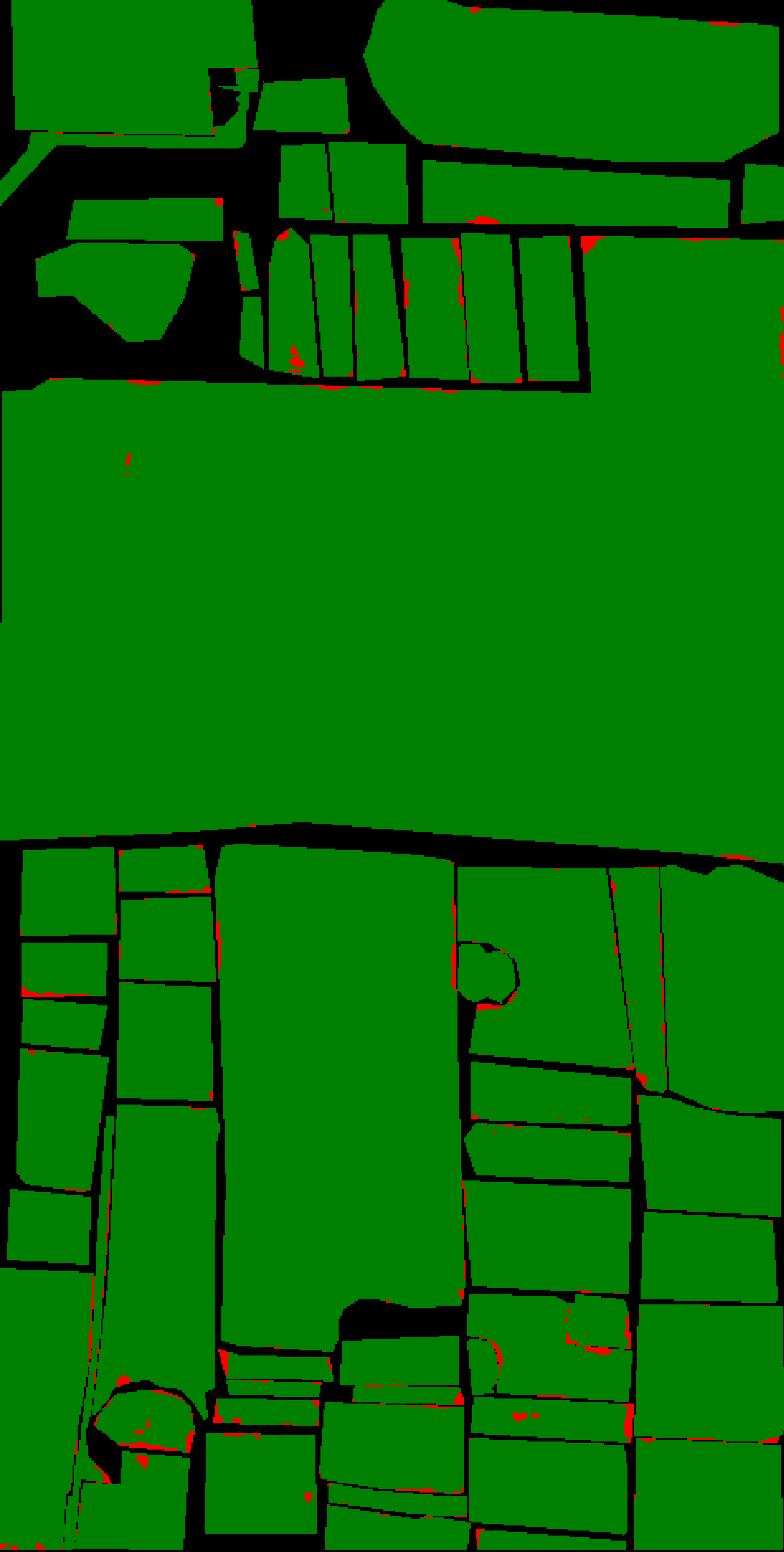}
	\caption{Error Map} 
    \end{subfigure}
    \caption{Attention-based analysis on the HH dataset.}
    \label{fig:HHAttention}
\end{figure}

\begin{figure}[!hbt]
    \centering
    \begin{subfigure}{0.11\textwidth}
	   \includegraphics[width=0.99\textwidth]{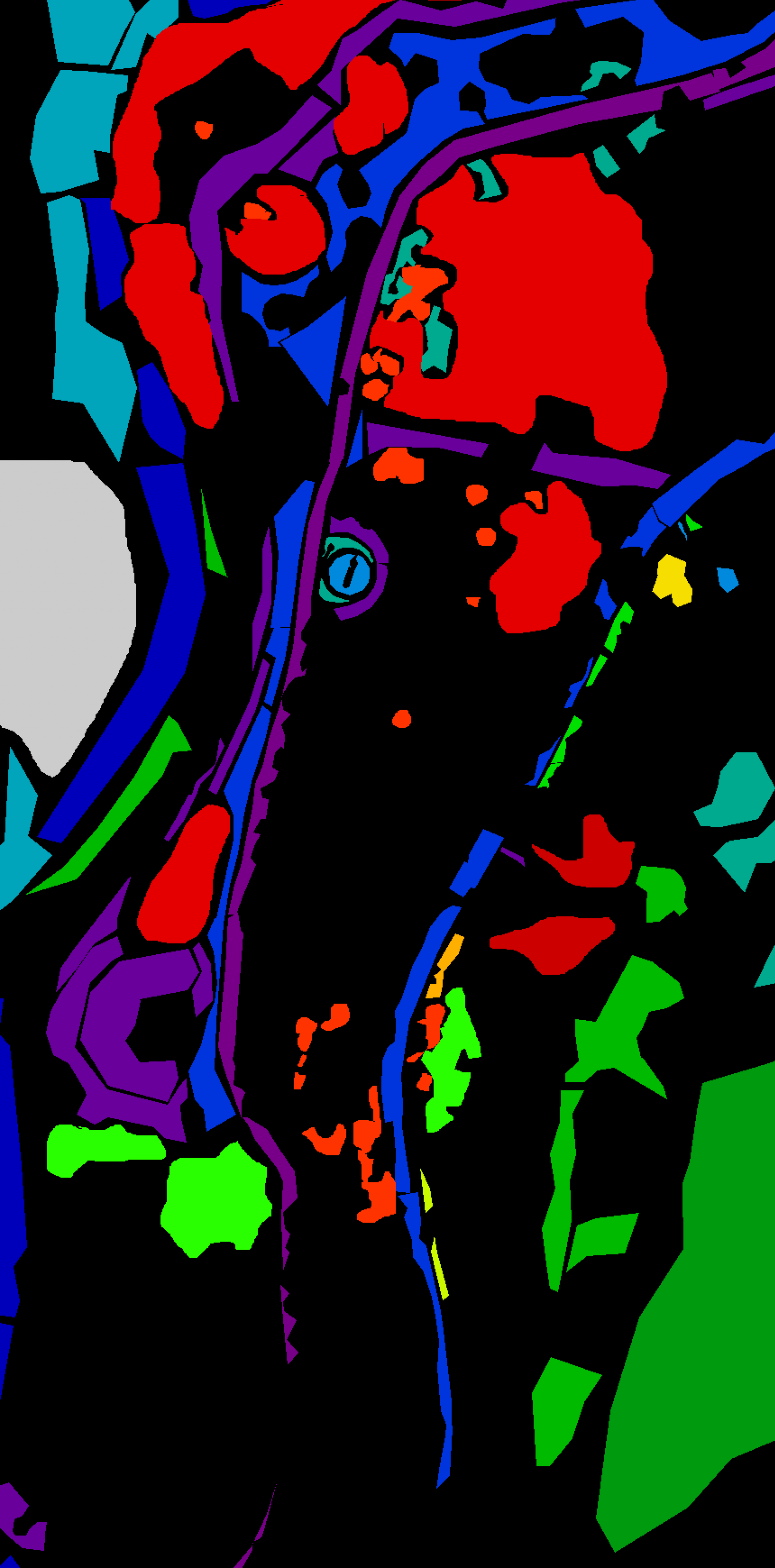}
	\caption{GT}
    \end{subfigure}
    \begin{subfigure}{0.14\textwidth}
	   \includegraphics[width=0.99\textwidth]{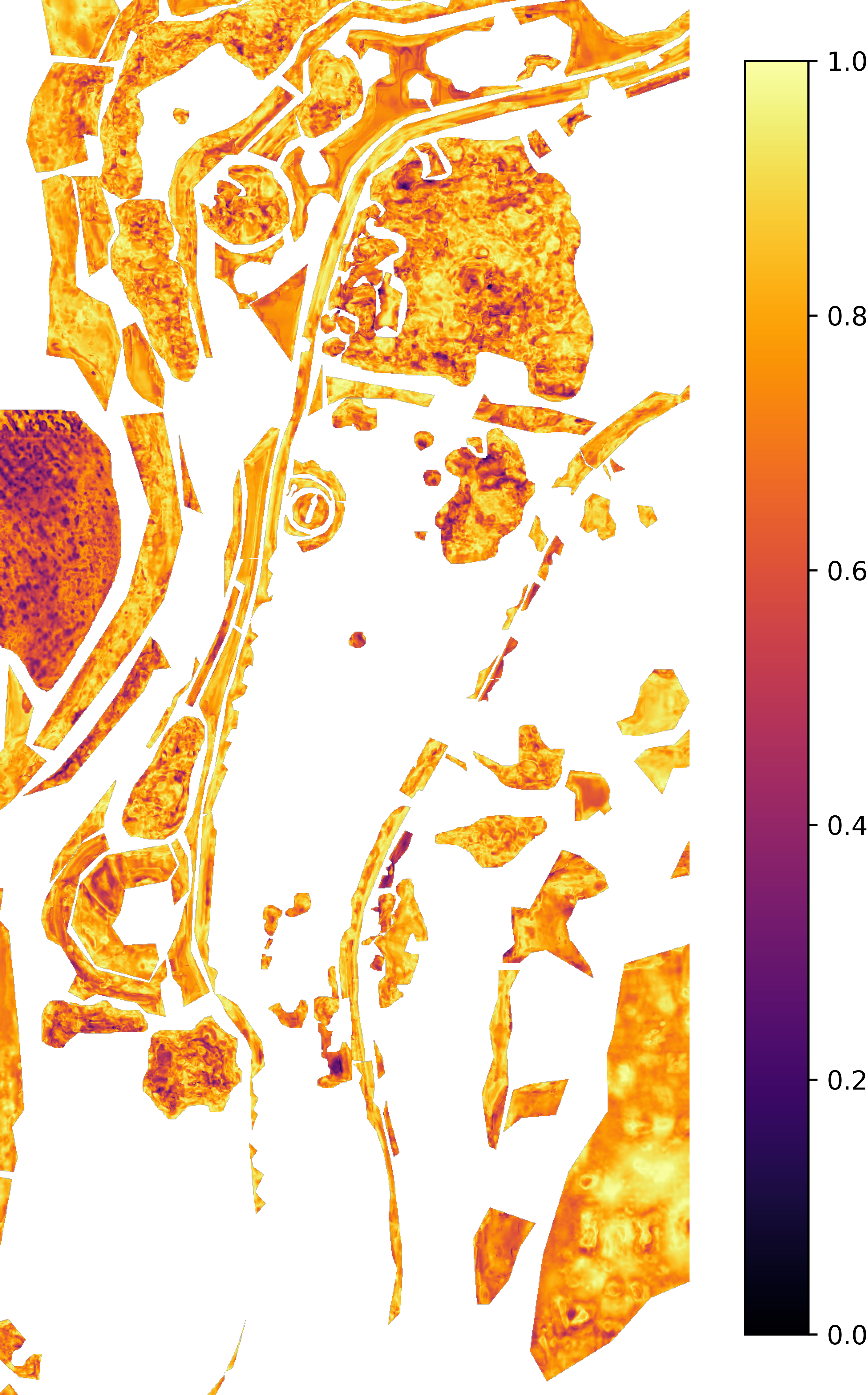}
	\caption{Heatmap} 
    \end{subfigure}
    \begin{subfigure}{0.11\textwidth}
	   \includegraphics[width=0.99\textwidth]{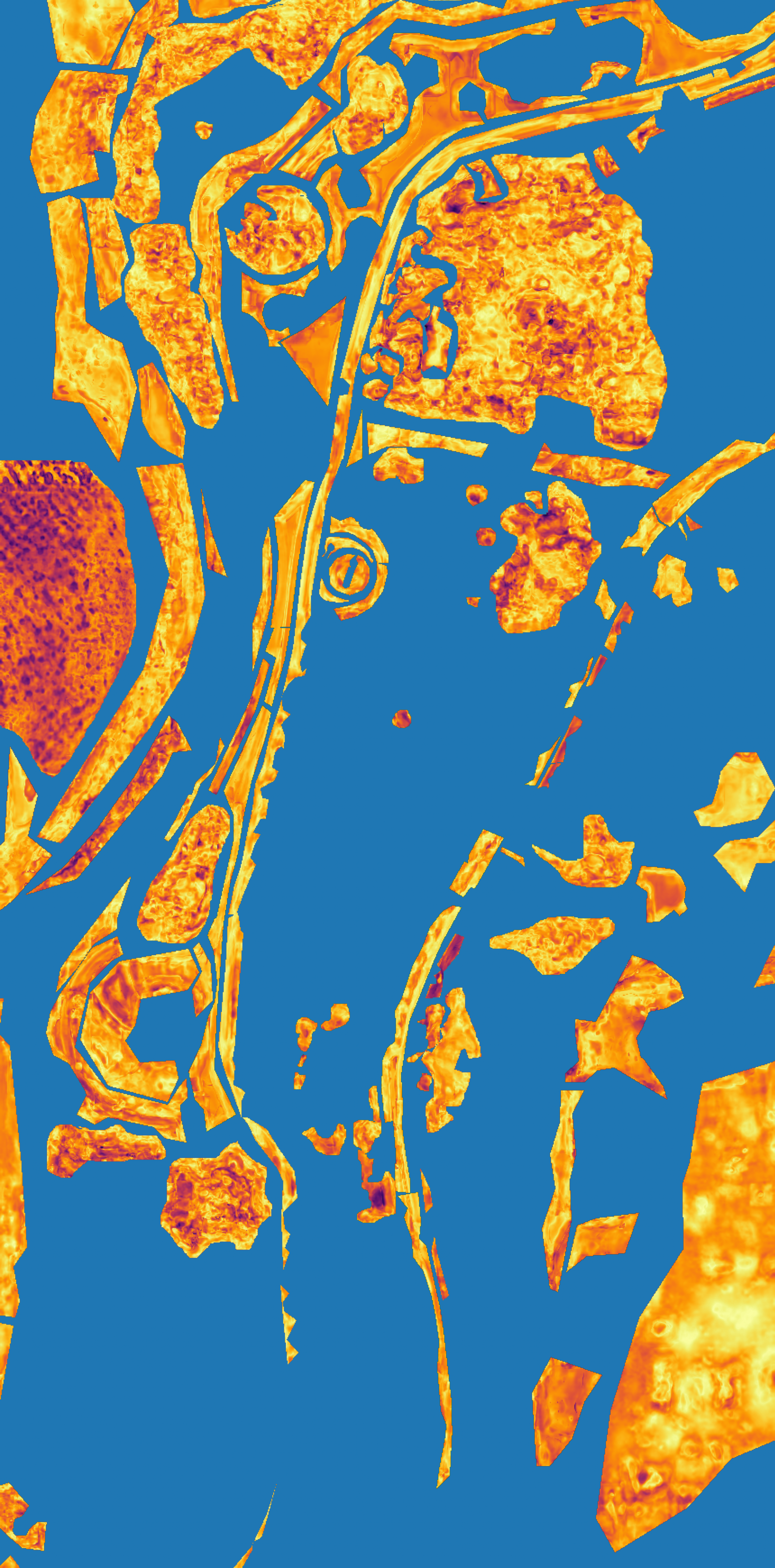}
	\caption{Overlay} 
    \end{subfigure}
    \begin{subfigure}{0.11\textwidth}
	   \includegraphics[width=0.99\textwidth]{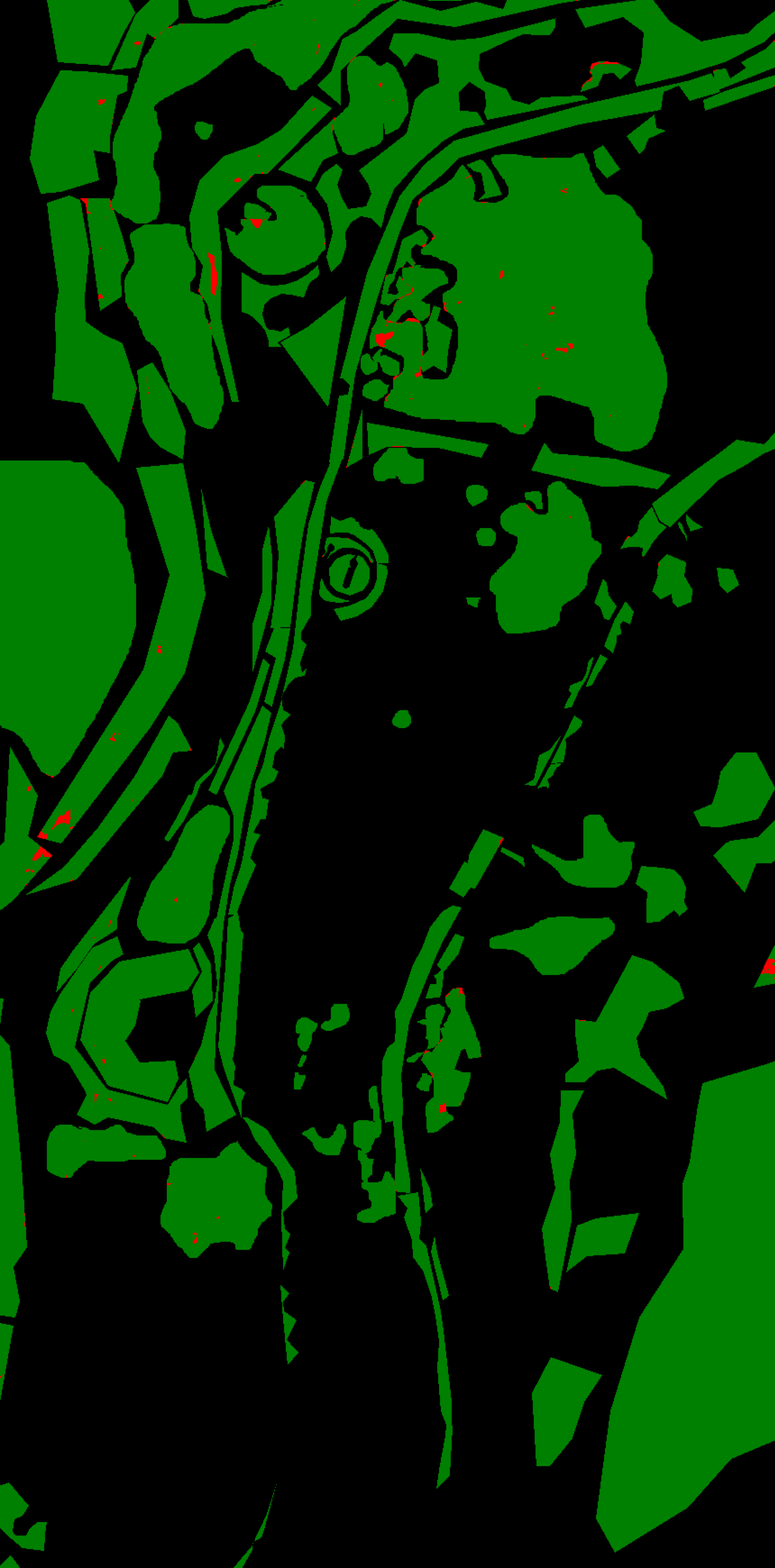}
	\caption{Error Map} 
    \end{subfigure}
    \caption{Attention-based analysis on the TD dataset.}
    \label{fig:TDAttention}
\end{figure}

To better understand the decision-making behavior, we analyze the spatial distribution of the learned attention responses. While quantitative metrics evaluate classification performance, they provide limited insight into how the model utilizes spatial-spectral information during inference. To address this limitation, the learned attention responses are projected back to the image plane and visualized as attention heatmaps and attention overlays on the corresponding GT scenes. These visualizations (Figs~\ref{fig:DFC18Attention}-\ref{fig:TDAttention}) are further compared with classification error maps to examine the relationship between attention localization and prediction quality. Such analysis provides qualitative evidence regarding whether the latent rollout mechanism guides the model toward discriminative and spatially coherent scene representations.

The attention visualizations reveal several consistent patterns. First, high-attention responses are concentrated within semantically homogeneous regions rather than being uniformly distributed across the scene. In DFC18 (Fig.~\ref{fig:DFC18Attention}), the learned attention focuses on dominant urban structures, including transportation corridors, vegetation regions, and built-up areas. For the Dioni and Loukia datasets (Figs.~\ref{fig:DioniAttention} and \ref{fig:LoukiaAttention}), attention responses remain spatially coherent over large land-cover regions, indicating that the model successfully captures long-range contextual relationships despite substantial intra-class variability. Similarly, for the agricultural HH dataset (Fig.~\ref{fig:HHAttention}), the attention maps exhibit highly compact and class-consistent patterns that align closely with crop distributions.

Second, the attention overlays demonstrate strong correspondence between the learned responses and semantically meaningful structures in the GT scenes. Regions receiving close attention generally coincide with dominant land-cover patterns and object boundaries that are critical for discrimination. This behavior suggests that the latent rollout mechanism encourages the model to focus on informative spatial-spectral structures while suppressing irrelevant background responses. Third, the error maps show that the majority of misclassifications occur near class boundaries and transition regions where neighboring categories exhibit similar spectral characteristics. In contrast, regions associated with strong attention responses are predominantly classified correctly. 

\begin{table*}[!hbt]
\centering
\caption{Ablation study of the proposed LWM framework. Relative Gain (\%) is computed with respect to the SST-only baseline. For OA, AA, $\kappa$, MCC, Top3, and Macro-F1, higher values are better, whereas for Loss, Log Loss, Brier Score, and ECE, lower values are preferred. Best values are highlighted in bold.}
\renewcommand{\arraystretch}{1.12}
    \resizebox{\textwidth}{!}{\begin{tabular}{llcccccccccc} \toprule
        \textbf{Dataset} & \textbf{Setting} & \textbf{Loss}$\downarrow$ & \textbf{OA}$\uparrow$ & \textbf{AA}$\uparrow$ & $\boldsymbol{\kappa}$$\uparrow$ & \textbf{MCC}$\uparrow$ & \textbf{Log Loss}$\downarrow$ & \textbf{Top3}$\uparrow$ & \textbf{Macro-F1}$\uparrow$ & \textbf{Brier}$\downarrow$ & \textbf{ECE}$\downarrow$ \\ \midrule

        \multirow{6}{*}{\textbf{DFC18}}
        & SST only & 0.5794 & 0.8992 & 0.9083 & 0.8683 & 0.8689 & 0.3925 & 0.9830 & 0.8919 & 0.1447 & 0.0385 \\
        & SST+Mask Recon & 0.6267 & 0.9076 & 0.9190 & 0.8796 & 0.8802 & 0.3390 & 0.9841 & 0.8994 & 0.1630 & 0.0493 \\
        & SST+World & 0.5196 & 0.9175 & 0.9324 & 0.8925 & 0.8931 & 0.2832 & 0.9878 & 0.9030 & 0.1260 & \textbf{0.0280} \\
        & LWM w/o Physical & 0.4601 & 0.9225 & 0.9325 & 0.8991 & 0.8996 & 0.2942 & 0.9877 & 0.9082 & 0.1234 & 0.0296 \\
        & \textbf{Full LWM} & \textbf{0.4100} & \textbf{0.9249} & \textbf{0.9352} & \textbf{0.9021} & \textbf{0.9024} & \textbf{0.2783} & \textbf{0.9899} & \textbf{0.9226} & \textbf{0.1192} & 0.0283 \\
        & Relative Gain (\%) & -29.23 & +2.86 & +2.96 & +3.89 & +3.86 & -29.10 & +0.70 & +3.44 & -17.62 & -26.49 \\ \midrule

        \multirow{6}{*}{\textbf{Dioni}}
        & SST only & 0.7405 & 0.8388 & 0.8696 & 0.8035 & 0.8069 & 0.4981 & 0.9829 & 0.8061 & 0.2341 & 0.0436 \\
        & SST+Mask Recon & 0.7712 & 0.8688 & 0.8888 & 0.8395 & 0.8408 & 0.4318 & 0.9847 & 0.8436 & 0.2017 & 0.0577 \\
        & SST+World & 0.6116 & 0.8967 & 0.9166 & 0.8731 & 0.8748 & 0.3362 & 0.9922 & 0.8625 & 0.1691 & 0.0356 \\
        & LWM w/o Physical & 0.5774 & 0.8920 & 0.8915 & 0.8677 & 0.8689 & 0.3702 & 0.9850 & 0.9014 & 0.1626 & 0.0183 \\
        & \textbf{Full LWM} & \textbf{0.5519} & \textbf{0.9171} & \textbf{0.9346} & \textbf{0.8979} & \textbf{0.8986} & \textbf{0.2894} & \textbf{0.9943} & \textbf{0.9146} & \textbf{0.1332} & \textbf{0.0080} \\
        & Relative Gain (\%) & -25.47 & +9.34 & +7.48 & +11.75 & +11.36 & -41.90 & +1.16 & +13.46 & -43.10 & -81.65 \\ \midrule

        \multirow{6}{*}{\textbf{Loukia}}
        & SST only & 0.7106 & 0.8541 & \textbf{0.8858} & 0.8293 & 0.8308 & 0.4360 & 0.9899 & 0.8426 & 0.2229 & 0.0415 \\
        & SST+Mask Recon & 0.7201 & 0.8557 & 0.8725 & 0.8307 & 0.8315 & 0.4579 & 0.9856 & 0.8400 & 0.2038 & 0.0411 \\
        & SST+World & 0.7296 & 0.8695 & 0.8626 & 0.8465 & 0.8470 & 0.4438 & 0.9820 & 0.8430 & 0.2021 & 0.0516 \\
        & LWM w/o Physical & 0.6806 & 0.8697 & 0.8718 & 0.8471 & 0.8479 & 0.4201 & 0.9891 & 0.8446 & 0.2208 & \textbf{0.0374} \\
        & \textbf{Full LWM} & \textbf{0.6112} & \textbf{0.8907} & 0.8775 & \textbf{0.8709} & \textbf{0.8713} & \textbf{0.3699} & \textbf{0.9902} & \textbf{0.8737} & \textbf{0.1723} & 0.0492 \\
        & Relative Gain (\%) & -14.00 & +4.29 & -0.94 & +5.02 & +4.88 & -15.16 & +0.03 & +3.69 & -22.70 & +18.55 \\ \midrule

        \multirow{6}{*}{\textbf{HH}}
        & SST only & 0.2452 & 0.9874 & 0.9805 & 0.9841 & 0.9841 & 0.0553 & \textbf{0.9992} & 0.9649 & 0.0252 & \textbf{0.0114} \\
        & SST+Mask Recon & 0.2694 & 0.9906 & 0.9841 & 0.9882 & 0.9882 & \textbf{0.0484} & \textbf{0.9992} & 0.9725 & 0.0197 & 0.0119 \\
        & SST+World & 0.1608 & 0.9840 & 0.9836 & 0.9799 & 0.9799 & 0.0650 & 0.9989 & 0.9739 & 0.0198 & 0.0118 \\
        & LWM w/o Physical & 0.1759 & 0.9878 & 0.9812 & 0.9846 & 0.9846 & 0.0556 & \textbf{0.9992} & 0.9774 & 0.0152 & 0.0130 \\
        & \textbf{Full LWM} & \textbf{0.1496} & \textbf{0.9914} & \textbf{0.9864} & \textbf{0.9891} & \textbf{0.9891} & 0.0486 & 0.9991 & \textbf{0.9826} & \textbf{0.0144} & 0.0131 \\
        & Relative Gain (\%) & -38.99 & +0.41 & +0.60 & +0.51 & +0.51 & -12.12 & -0.01 & +1.83 & -42.86 & +14.71 \\ \midrule
        
        \multirow{6}{*}{\textbf{TD}}
        & SST only & 0.1263 & 0.9893 & 0.9930 & 0.9878 & 0.9879 & 0.0652 & 0.9995 & 0.9848 & 0.0096 & 0.0148 \\
        & SST+Mask Recon & 0.1564 & 0.9904 & 0.9926 & 0.9891 & 0.9891 & 0.0533 & 0.9996 & 0.9927 & 0.0128 & 0.0147 \\
        & SST+World & 0.2661 & 0.9918 & 0.9946 & 0.9907 & 0.9907 & 0.0410 & \textbf{0.9999} & 0.9911 & 0.0191 & \textbf{0.0112} \\
        & LWM w/o Physical & 0.2091 & 0.9942 & \textbf{0.9959} & 0.9934 & 0.9934 & 0.0364 & 0.9995 & 0.9913 & 0.0162 & 0.0141 \\
        & \textbf{Full LWM} & \textbf{0.1166} & \textbf{0.9952} & 0.9958 & \textbf{0.9945} & \textbf{0.9945} & \textbf{0.0324} & \textbf{0.9999} & \textbf{0.9947} & \textbf{0.0079} & 0.0153 \\
        & Relative Gain (\%) & -7.68 & +0.60 & +0.28 & +0.68 & +0.67 & -50.31 & +0.04 & +1.01 & -17.71 & +3.38 \\ \bottomrule
\end{tabular}}
\label{tab:Ablation}
\end{table*}

\section{Ablation Study}
\label{sec:ablation}

To quantify the contribution of each component, we progressively construct the proposed LWM framework from a baseline Spatial-Spectral Transformer (SST). Specifically, five configurations are evaluated: (i) SST only, which contains only the spatial-spectral transformer backbone; (ii) SST+Mask Recon, which additionally introduces masked spectral and spatial reconstruction objectives; (iii) SST+World, which incorporates the stochastic LWM module and rollout reasoning; (iv) LWM w/o Physical Priors, which further includes latent rollout planning and trajectory supervision but excludes SAM, spectral smoothness, and spatial consistency constraints; and (v) the full LWM model. The relative improvement of the full LWM model over the SST baseline is computed as $\mathrm{Relative~Gain}(\%) = \frac{\mathcal{M}_{\mathrm{LWM}}-\mathcal{M}_{\mathrm{SST}}}{\mathcal{M}_{\mathrm{SST}}} \times 100$, where $\mathcal{M}$ denotes the evaluation metric of interest. For metrics where lower values are better, such as Log Loss, negative values indicate performance improvement.

From Table~\ref{tab:Ablation}, several important observations can be made. First, masked spatial-spectral reconstruction consistently improves performance over the SST baseline. Although the gains in OA are relatively modest, reconstruction supervision produces noticeable improvements in AA, $\kappa$, MCC, and calibration metrics. This indicates that recovering masked spectral bands and spatial pixels encourages the encoder to preserve richer spatial-spectral information, resulting in more robust latent representations under limited supervision.

Second, introducing the stochastic LWM yields substantially larger improvements than reconstruction alone. For example, on Dioni, OA increases from 83.88\% to 89.67\%, while $\kappa$ improves from 0.8035 to 0.8731. Similar trends are observed on DFC18 and Loukia. These results demonstrate that latent rollout reasoning provides information beyond conventional transformer-based feature aggregation. 

Third, trajectory planning and trajectory-aware supervision further improve performance. Comparing SST+World with LWM w/o physical priors shows consistent gains in OA, $\kappa$, MCC, and Macro-F1 on most datasets. This observation validates the effectiveness of aggregating multiple rollout states through learned importance weighting rather than relying solely on the initial latent representation. The results suggest that trajectory-aware aggregation enables the classifier to exploit complementary information distributed across the latent rollout trajectory.

Finally, the full LWM model achieves the best overall performance across almost all datasets. The inclusion of physically motivated priors, including spectral angle alignment, spectral smoothness, and spatial consistency regularization, consistently improves classification accuracy while simultaneously reducing probabilistic uncertainty. In particular, Dioni exhibits the largest gains, where the full model achieves relative improvements of 9.34\%, 7.48\%, 11.75\%, and 11.36\% in OA, AA, $\kappa$, and MCC, respectively, compared with the SST baseline. Furthermore, the Brier score is reduced by 43.10\% and ECE by 81.65\%, indicating substantially improved prediction calibration. Similar trends are observed on DFC18, where calibration improves by 17.62\% and 26.49\% in terms of Brier score and ECE, respectively. The HH dataset exhibits comparatively smaller improvements because the baseline transformer already achieves near-saturated performance.

\begin{table*}[!t]
\centering
\caption{Comparison with recent state-of-the-art HSIC methods. Results are reported as mean $\pm$ standard deviation over five runs. The best and second-best results are highlighted in bold and underlined italics, respectively. The training sample ratio used by each method is reported in the row ``Training Samples (\%)''.}
\renewcommand{\arraystretch}{1.15}
    \resizebox{\textwidth}{!}{\begin{tabular}{llcccccccc} \toprule
    \multirow{2}{*}{\textbf{Dataset}} & \multirow{2}{*}{\textbf{Metric}} & \multicolumn{8}{c}{\textbf{Methods}} \\ \cmidrule(lr){3-10}
    & & \textbf{LWM} & \textbf{WF} & \textbf{DF} & \textbf{MCMTN} & \textbf{SMFAT} & \textbf{MSST} & \textbf{S2CIFT} & \textbf{S2CAT} \\ \midrule

    \multicolumn{2}{c}{Training Samples (\%)} & 10 & 50 & 50 & 50 & 50 & 50 & 50 & 50 \\ \cmidrule(lr){2-10}
    \multirow{3}{*}{\rotatebox{90}{\textbf{DFC18}}} & OA & \textbf{92.4997$\pm$0.1096} & 91.4101$\pm$0.5559 & 89.6109$\pm$0.0707 & 90.4070$\pm$0.1673 & \underline{\textit{91.9506$\pm$0.1864}} & 87.0541$\pm$0.3509 & 91.6866$\pm$0.0159 & 91.5293$\pm$0.2206 \\
    & AA & \textbf{93.5275$\pm$0.0126} & 88.9789$\pm$0.1071 & 81.3057$\pm$0.3993 & 88.0639$\pm$0.1983 & 85.6948$\pm$0.1386 & 78.4143$\pm$0.7389 & \underline{\textit{90.9763$\pm$0.3494}} & 88.9707$\pm$0.9180 \\
    & $\kappa$ & \textbf{90.2191$\pm$0.0231} & \underline{\textit{89.9807$\pm$0.7378}} & 86.2516$\pm$0.1730 & 87.2409$\pm$0.1456 & 89.0046$\pm$0.1064 & 82.7725$\pm$0.4699 & 89.6388$\pm$0.0193 & 89.1599$\pm$0.3376 \\ \midrule

    \multirow{3}{*}{\rotatebox{90}{\textbf{Loukia}}} & OA & \textbf{89.2107$\pm$0.1094} & \underline{\textit{88.3205$\pm$0.2095}} & 88.0258$\pm$0.9530 & 80.4684$\pm$0.2983 & 82.7826$\pm$0.1956 & 76.2885$\pm$1.2358 & 87.3655$\pm$0.1885 & 87.2983$\pm$0.3037 \\
    & AA & \textbf{87.9875$\pm$0.1453} & \underline{\textit{87.2253$\pm$0.3477}} & 83.2571$\pm$2.2707 & 80.0375$\pm$0.2091 & 82.5046$\pm$0.2193 & 67.3756$\pm$0.2072 & 86.2443$\pm$0.1001 & 86.1633$\pm$0.5295 \\
    & $\kappa$ & \textbf{87.9916$\pm$0.1312} & \underline{\textit{87.0363$\pm$0.2558}} & 85.7381$\pm$1.1069 & 76.5459$\pm$0.1659 & 79.4113$\pm$0.2510 & 71.8070$\pm$1.4110 & 86.7121$\pm$0.2254 & 87.0125$\pm$0.3645 \\ \midrule

    \multicolumn{2}{c}{Training Samples (\%)} & 2 & 2 & 2 & 2 & 2 & 2 & 2 & 2 \\ \cmidrule(lr){2-10}
    \multirow{3}{*}{\rotatebox{90}{\textbf{Dioni}}} & OA & \textbf{91.7132$\pm$0.1453} & 90.8979$\pm$0.4669 & 90.7228$\pm$0.0196 & 88.9929$\pm$0.2319 & 83.2195$\pm$0.1639 & 89.0911$\pm$1.4596 & \underline{\textit{90.9907$\pm$0.8240}} & 90.9895$\pm$0.1569 \\
    & AA & \textbf{93.4698$\pm$0.1652} & \underline{\textit{89.7330$\pm$2.0580}} & 85.1594$\pm$0.7765 & 76.1774$\pm$0.1982 & 70.9206$\pm$0.2197 & 75.2173$\pm$6.3917 & 86.8359$\pm$1.9078 & 89.5406$\pm$3.6479 \\
    & $\kappa$ & \textbf{89.7989$\pm$0.1452} & 88.6610$\pm$0.5898 & 88.9784$\pm$0.0236 & 86.2635$\pm$0.1596 & 78.9040$\pm$0.1932 & 86.3184$\pm$1.9485 & \underline{\textit{89.5262$\pm$1.0279}} & 88.7806$\pm$0.2026 \\ \midrule

    \multirow{3}{*}{\rotatebox{90}{\textbf{HH}}} & OA & \textbf{99.1497$\pm$0.0914} & 96.9677$\pm$0.1979 & 96.8488$\pm$0.0625 & 98.1871$\pm$0.1934 & \underline{\textit{98.8740$\pm$0.1864}} & 97.1105$\pm$0.2391 & 97.8370$\pm$0.3560 & 97.0178$\pm$0.0101 \\
    & AA & \textbf{98.6489$\pm$0.0912} & 92.5037$\pm$1.0111 & 91.8830$\pm$0.0323 & 95.3153$\pm$0.1684 & \underline{\textit{97.5921$\pm$0.1274}} & 90.6244$\pm$2.2489 & 94.3517$\pm$1.8307 & 93.0144$\pm$0.4234 \\
    & $\kappa$ & \textbf{98.9198$\pm$0.0731} & 96.1681$\pm$0.2495 & 96.0204$\pm$0.0793 & 97.7084$\pm$0.1337 & \underline{\textit{98.5767$\pm$0.1425}} & 96.3459$\pm$0.3048 & 97.2674$\pm$0.4510 & 96.2311$\pm$0.0129 \\ \midrule

    \multirow{3}{*}{\rotatebox{90}{\textbf{TD}}} & OA & \textbf{99.5243$\pm$0.0176} & 96.8847$\pm$0.1554 & 96.9727$\pm$1.2515 & 98.1632$\pm$0.2984 & \underline{\textit{98.7562$\pm$0.1973}} & 97.3469$\pm$0.0576 & 97.8779$\pm$0.6498 & 96.7219$\pm$0.0587 \\
    & AA & \textbf{99.5853$\pm$0.0912} & 93.8958$\pm$0.8300 & 93.5081$\pm$4.6320 & 98.0945$\pm$0.2532 & \underline{\textit{98.6965$\pm$0.1673}} & 93.8685$\pm$0.7378 & 95.2817$\pm$1.0258 & 93.5659$\pm$1.1773 \\
    & $\kappa$ & \textbf{99.4543$\pm$0.0154} & 96.4494$\pm$0.1780 & 96.5505$\pm$1.4312 & 97.9096$\pm$0.1903 & \underline{\textit{98.5843$\pm$0.1437}} & 96.9757$\pm$0.0664 & 97.5839$\pm$0.7441 & 96.2641$\pm$0.0634 \\ \bottomrule
\end{tabular}}
\vspace{2ex}
  \begin{minipage}{\linewidth}
    \footnotesize
    \textsuperscript{}It is important to note that the compared methods were originally designed and evaluated under different training protocols and dataset complexities. Preliminary experiments showed that several baseline methods experienced severe performance degradation and unstable convergence when trained using the same extremely limited number of labeled samples employed by LWM on the more challenging benchmark datasets. Despite using substantially fewer labeled samples, LWM consistently achieves competitive performance across all datasets.
  \end{minipage}
\label{tab:Comparitive}
\end{table*}
\begin{figure*}[!htb]
    \centering
        \begin{subfigure}{0.32\textwidth}
            \includegraphics[width=0.99\textwidth]{DFC18/DFC18_GT.png}
            \caption{GT}
        \end{subfigure}
        \begin{subfigure}{0.32\textwidth}
            \includegraphics[width=0.99\textwidth]{DFC18/DFC18_Error.png}
            \caption{LWM}
        \end{subfigure}
        \begin{subfigure}{0.32\textwidth}
            \includegraphics[width=0.99\textwidth]{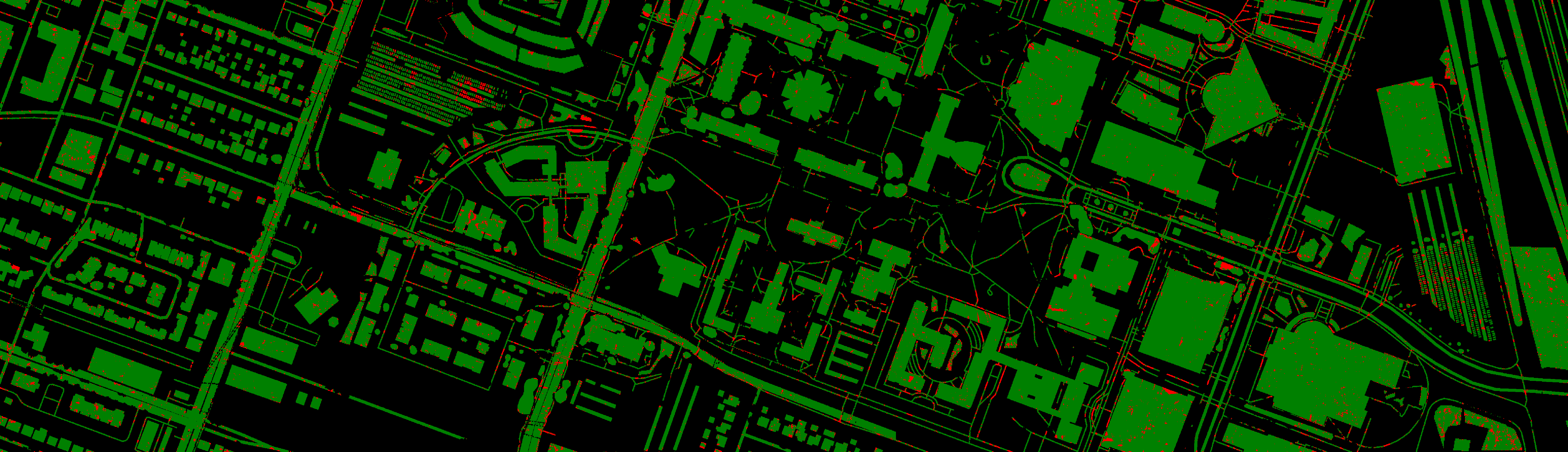}
            \caption{WF}
        \end{subfigure}
        \begin{subfigure}{0.32\textwidth}
            \includegraphics[width=0.99\textwidth]{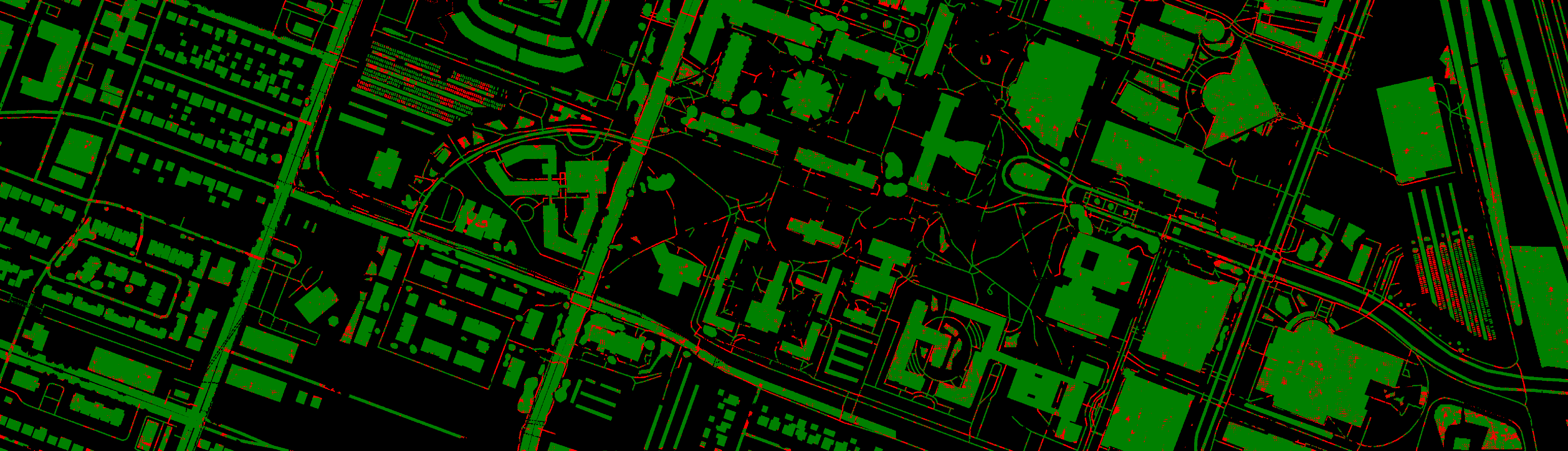}
            \caption{DF}
        \end{subfigure}
        \begin{subfigure}{0.32\textwidth}
            \includegraphics[width=0.99\textwidth]{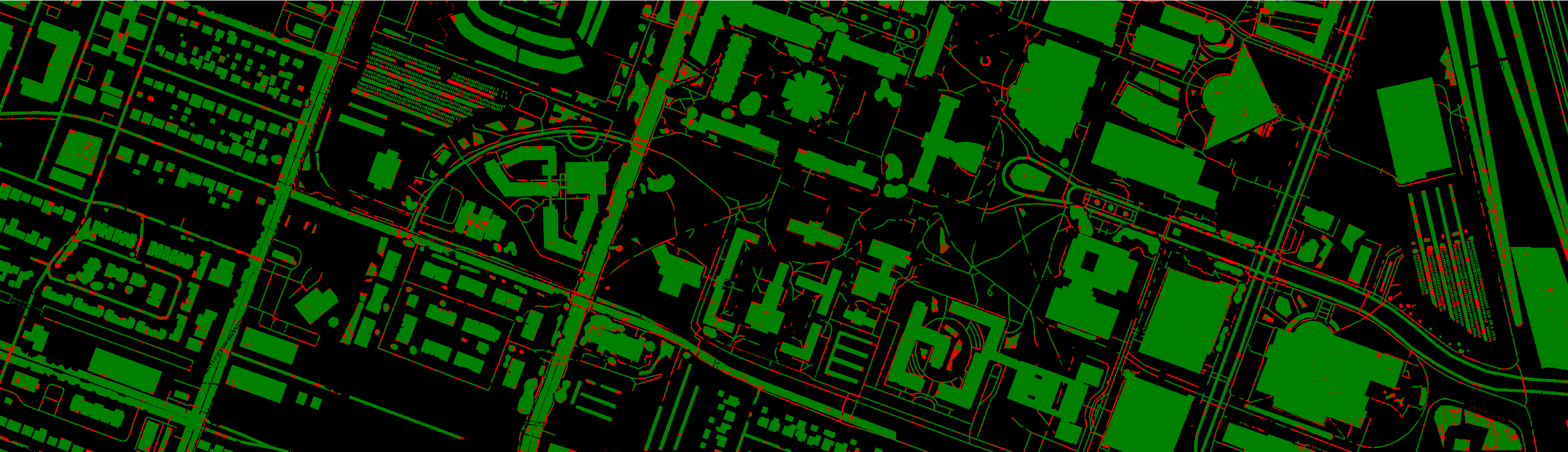}
            \caption{MCMTN}
        \end{subfigure}
        \begin{subfigure}{0.32\textwidth}
            \includegraphics[width=0.99\textwidth]{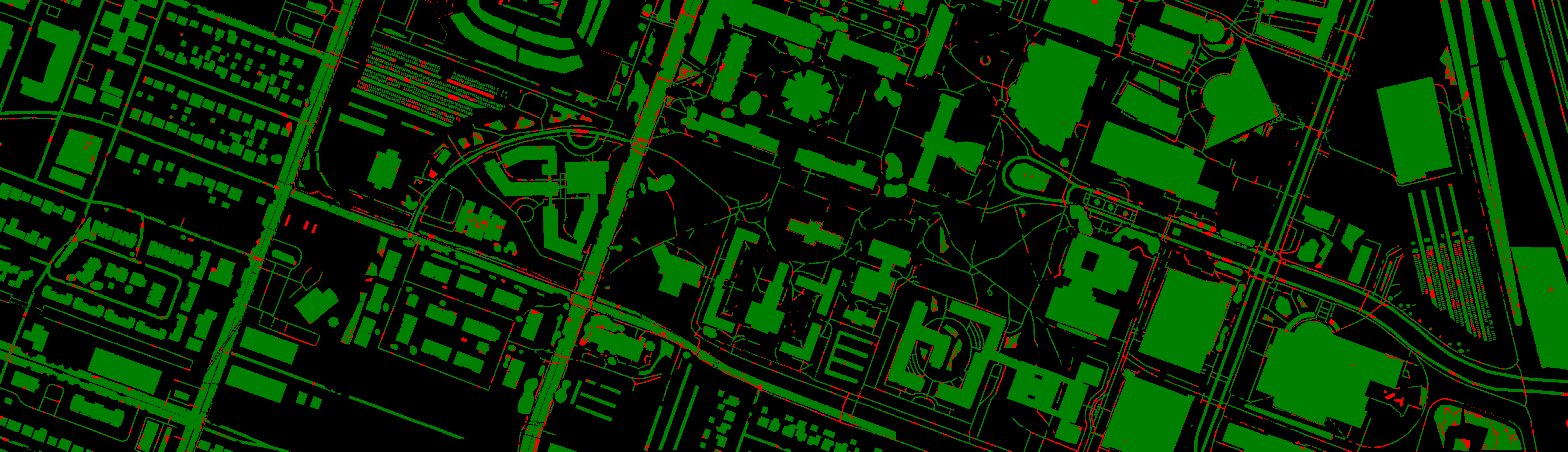}
            \caption{SMFAT}
        \end{subfigure}
        \begin{subfigure}{0.32\textwidth}
            \includegraphics[width=0.99\textwidth]{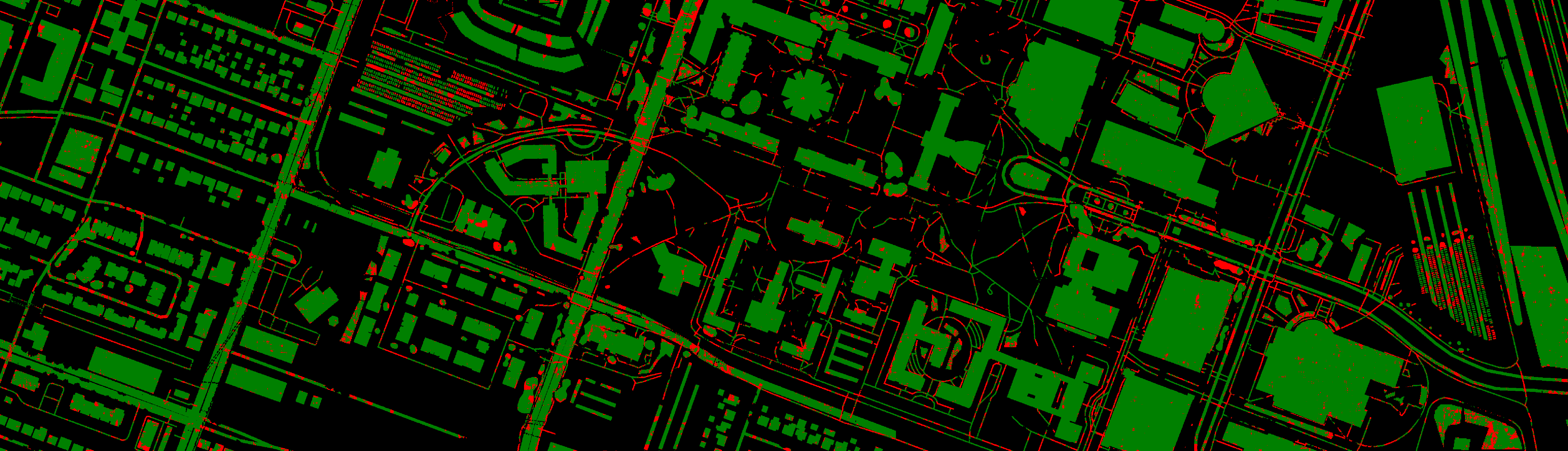}
            \caption{MSST}
        \end{subfigure}
        \begin{subfigure}{0.32\textwidth}
            \includegraphics[width=0.99\textwidth]{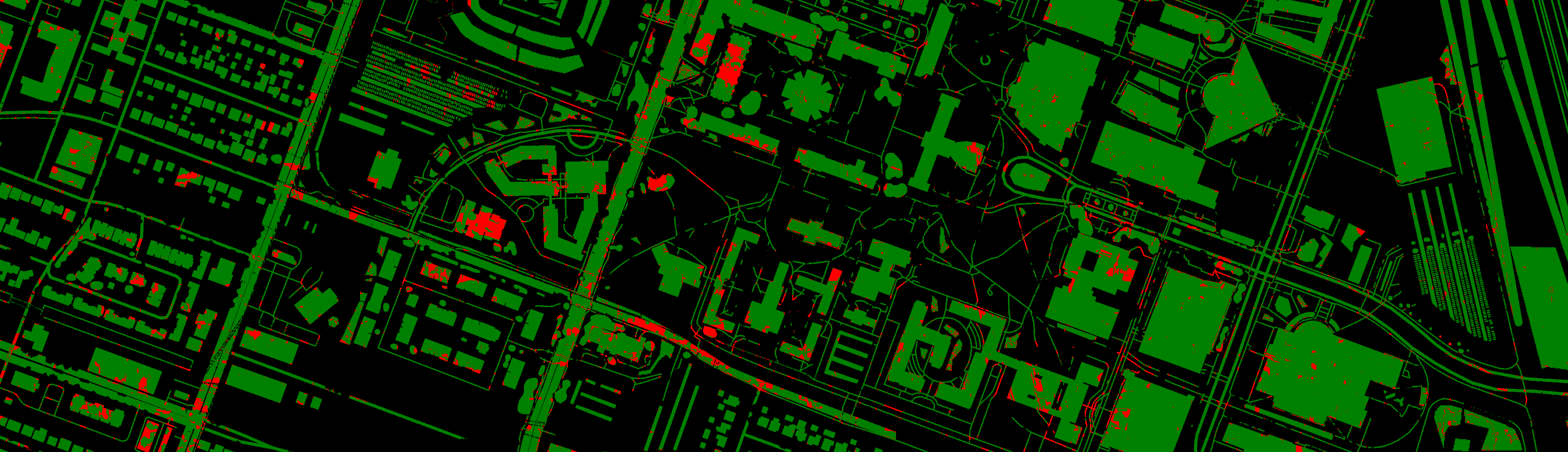}
            \caption{S2CIFT}
        \end{subfigure}
        \begin{subfigure}{0.32\textwidth}
            \includegraphics[width=0.99\textwidth]{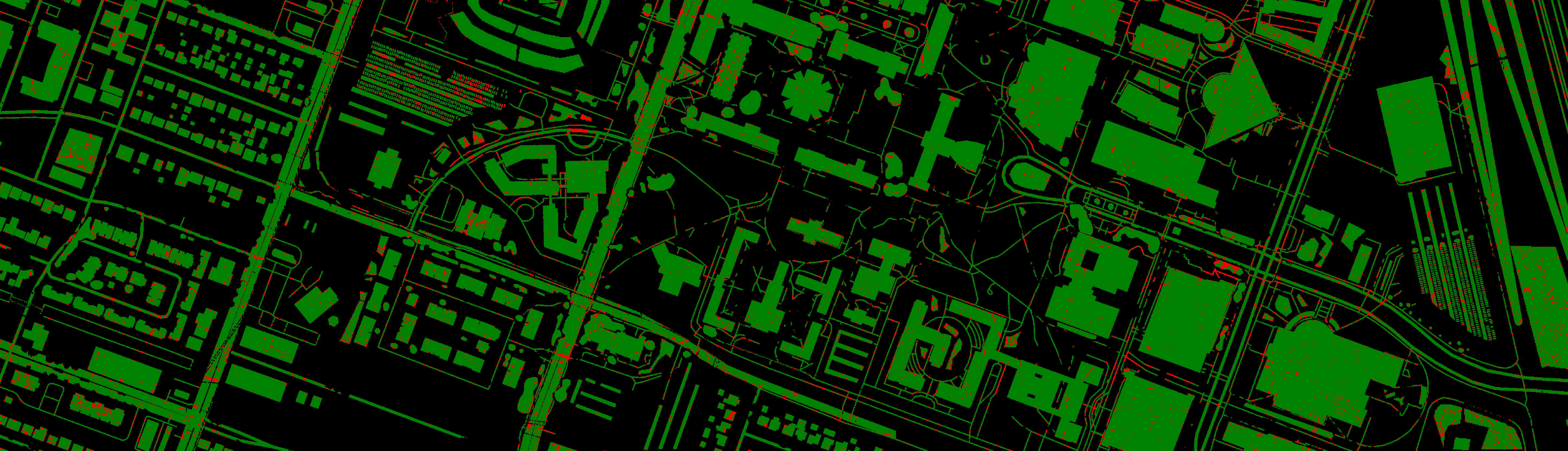}
            \caption{S2CAT}
        \end{subfigure}
    \caption{Qualitative comparison on the DFC18 dataset. The proposed LWM framework exhibits improved spatial consistency and reduced boundary confusion compared with existing transformer-based methods. Error maps reveal fewer fragmented predictions and more coherent class structures, indicating that latent trajectory modeling enhances representation robustness under spectrally similar categories.}
\label{fig:DFCComparison}
    \label{DFCGT}
\end{figure*}
\begin{figure*}[!htb]
    \centering
        \begin{subfigure}{0.32\textwidth}
            \includegraphics[width=0.99\textwidth]{Dioni/Dioni_GT.png}
            \caption{GT}
        \end{subfigure}
        \begin{subfigure}{0.32\textwidth}
            \includegraphics[width=0.99\textwidth]{Dioni/Dioni_Error.png}
            \caption{LWM}
        \end{subfigure}
        \begin{subfigure}{0.32\textwidth}
            \includegraphics[width=0.99\textwidth]{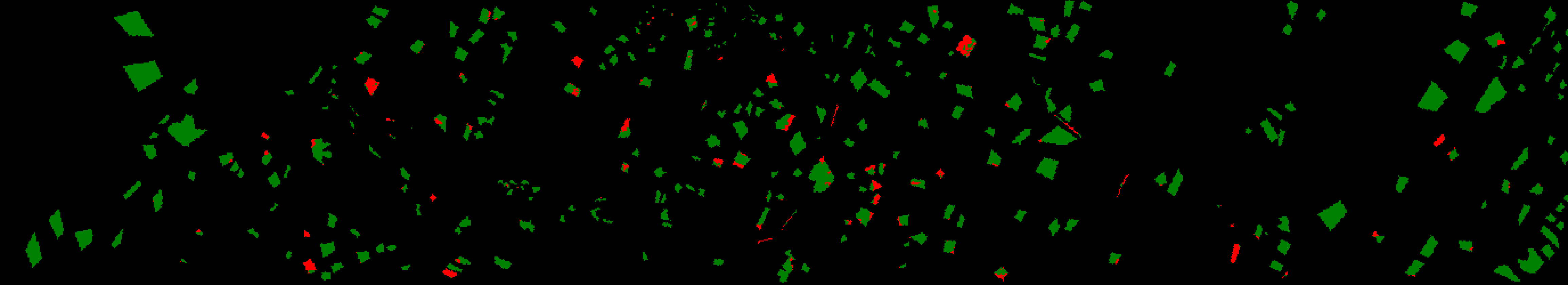}
            \caption{WF}
        \end{subfigure}
        \begin{subfigure}{0.32\textwidth}
            \includegraphics[width=0.99\textwidth]{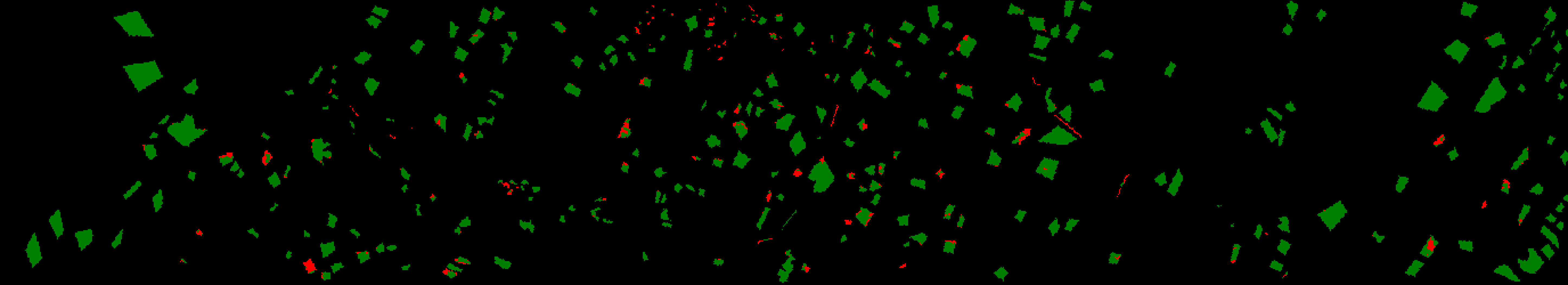}
            \caption{DF}
        \end{subfigure}
        \begin{subfigure}{0.32\textwidth}
            \includegraphics[width=0.99\textwidth]{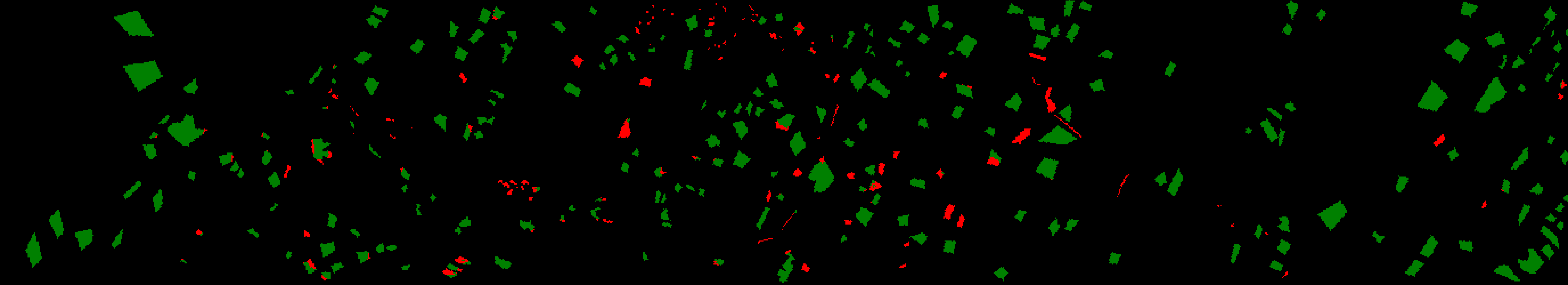}
            \caption{MCMTN}
        \end{subfigure}
        \begin{subfigure}{0.32\textwidth}
            \includegraphics[width=0.99\textwidth]{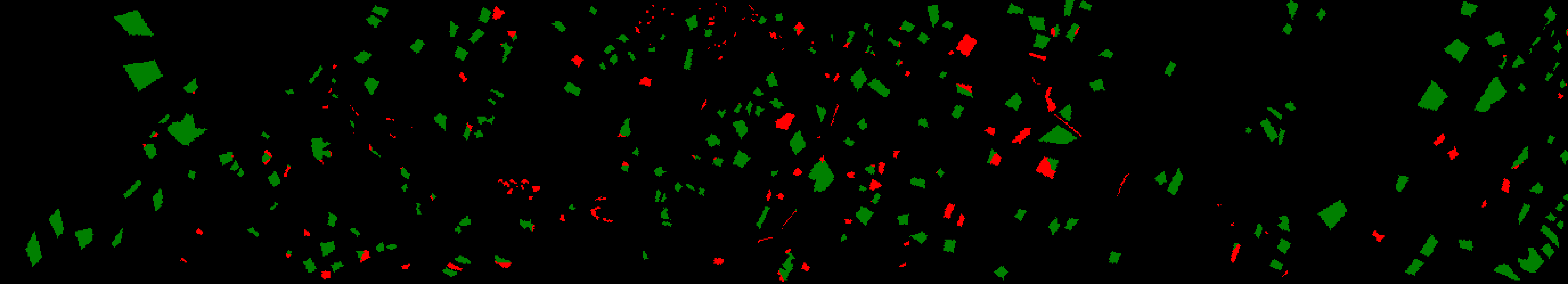}
            \caption{SMFAT}
        \end{subfigure}
        \begin{subfigure}{0.32\textwidth}
            \includegraphics[width=0.99\textwidth]{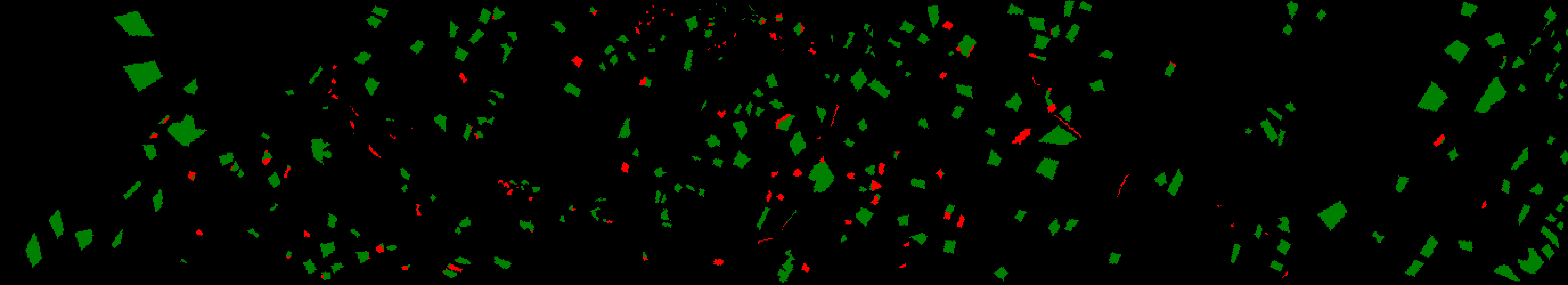}
            \caption{MSST}
        \end{subfigure}
        \begin{subfigure}{0.32\textwidth}
            \includegraphics[width=0.99\textwidth]{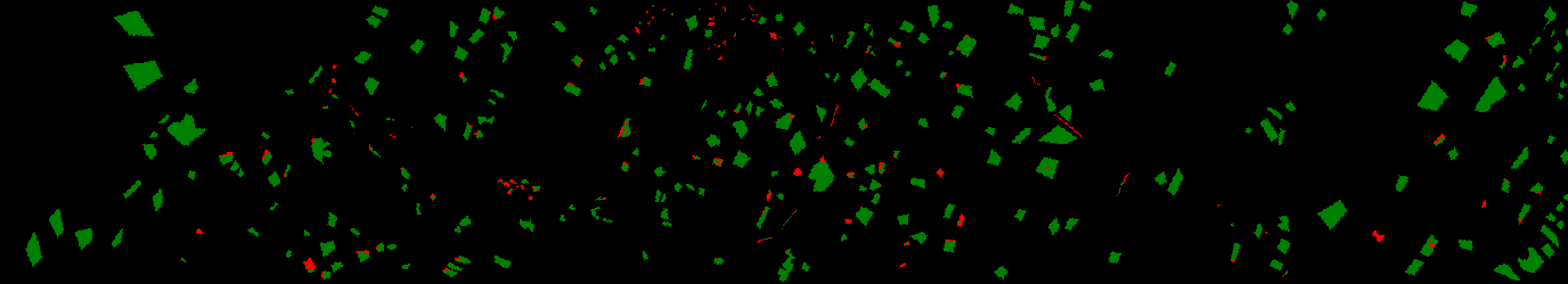}
            \caption{S2CIFT}
        \end{subfigure}
        \begin{subfigure}{0.32\textwidth}
            \includegraphics[width=0.99\textwidth]{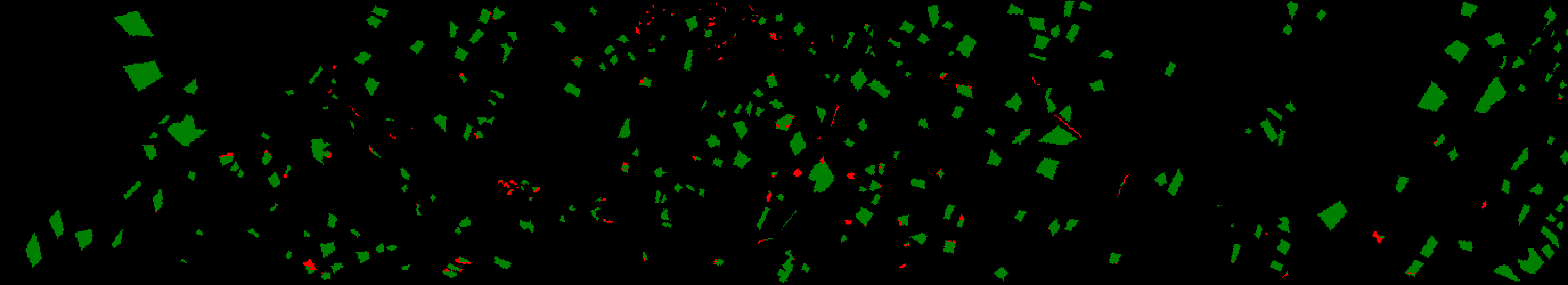}
            \caption{S2CAT}
        \end{subfigure}
    \caption{Qualitative comparison on the Dioni dataset.}
\label{fig:DioniComparison}
    \label{DIGT}
\end{figure*}
\begin{figure*}[!htb]
    \centering
        \begin{subfigure}{0.32\textwidth}
            \includegraphics[width=0.99\textwidth]{Loukia/Loukia_GT.png}
            \caption{GT}
        \end{subfigure}
        \begin{subfigure}{0.32\textwidth}
            \includegraphics[width=0.99\textwidth]{Loukia/Loukia_Error.png}
            \caption{LWM}
        \end{subfigure}
        \begin{subfigure}{0.32\textwidth}
            \includegraphics[width=0.99\textwidth]{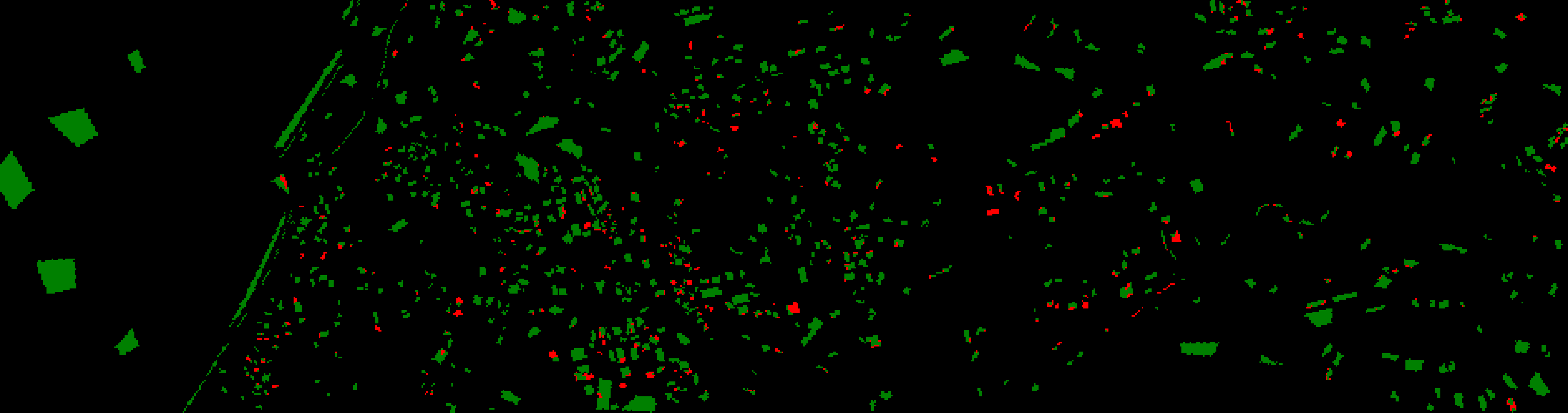}
            \caption{WF}
        \end{subfigure}
        \begin{subfigure}{0.32\textwidth}
            \includegraphics[width=0.99\textwidth]{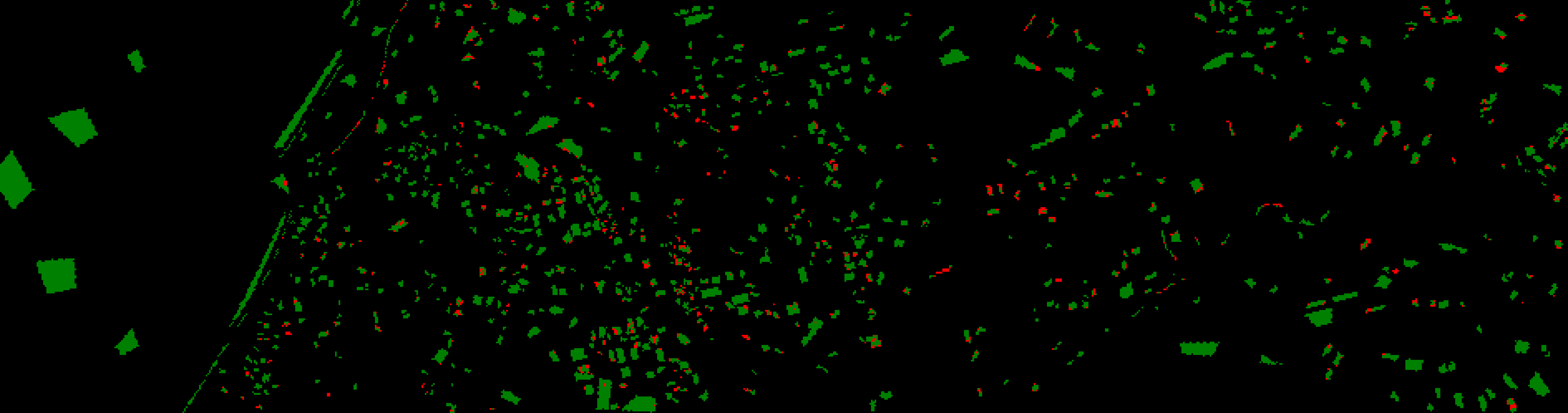}
            \caption{DF}
        \end{subfigure}
        \begin{subfigure}{0.32\textwidth}
            \includegraphics[width=0.99\textwidth]{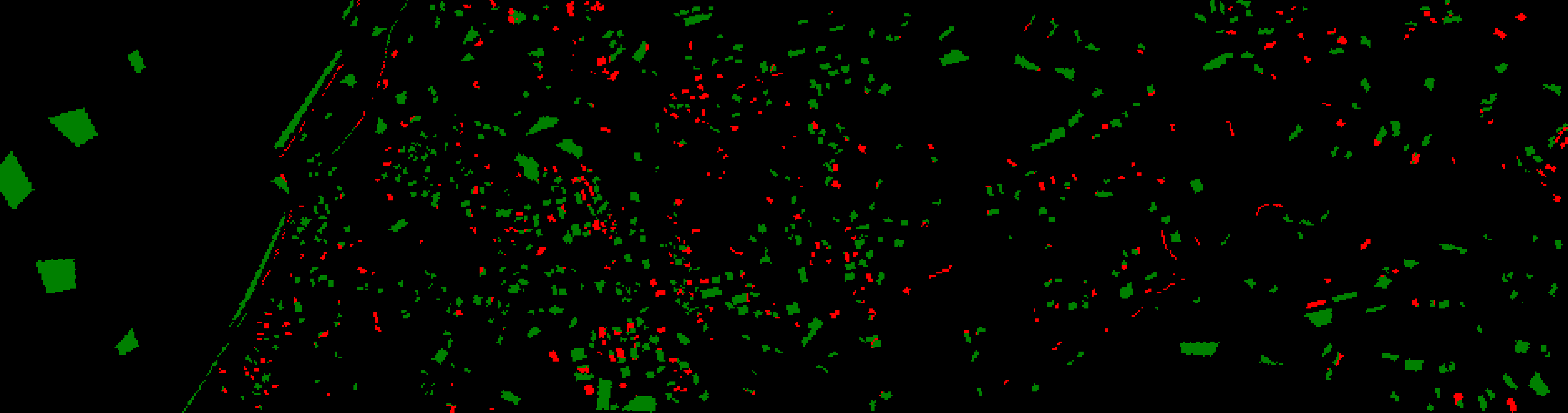}
            \caption{MCMTN}
        \end{subfigure}
        \begin{subfigure}{0.32\textwidth}
            \includegraphics[width=0.99\textwidth]{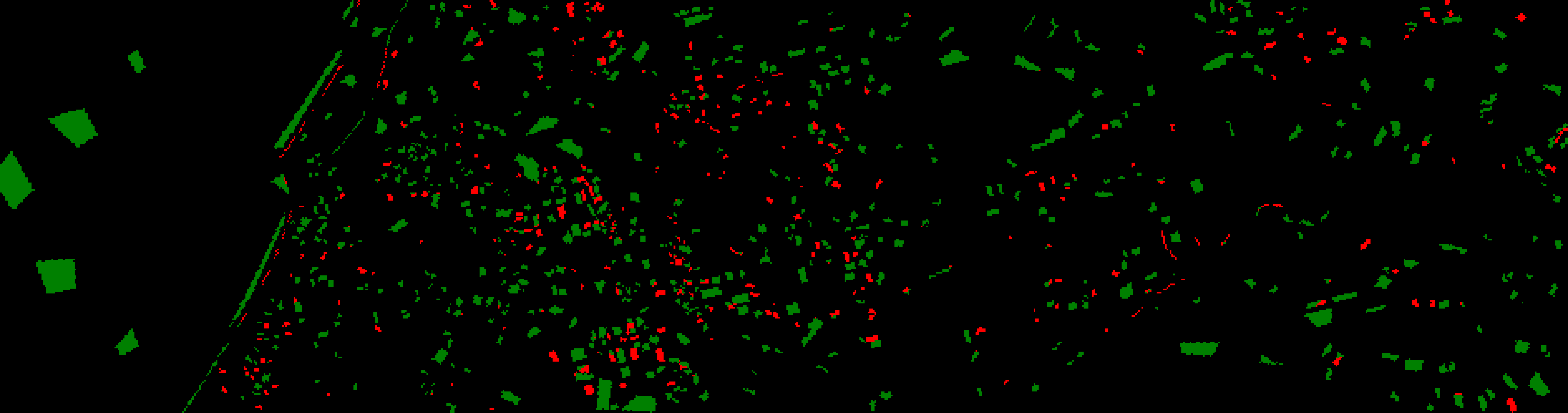}
            \caption{SMFAT}
        \end{subfigure}
        \begin{subfigure}{0.32\textwidth}
            \includegraphics[width=0.99\textwidth]{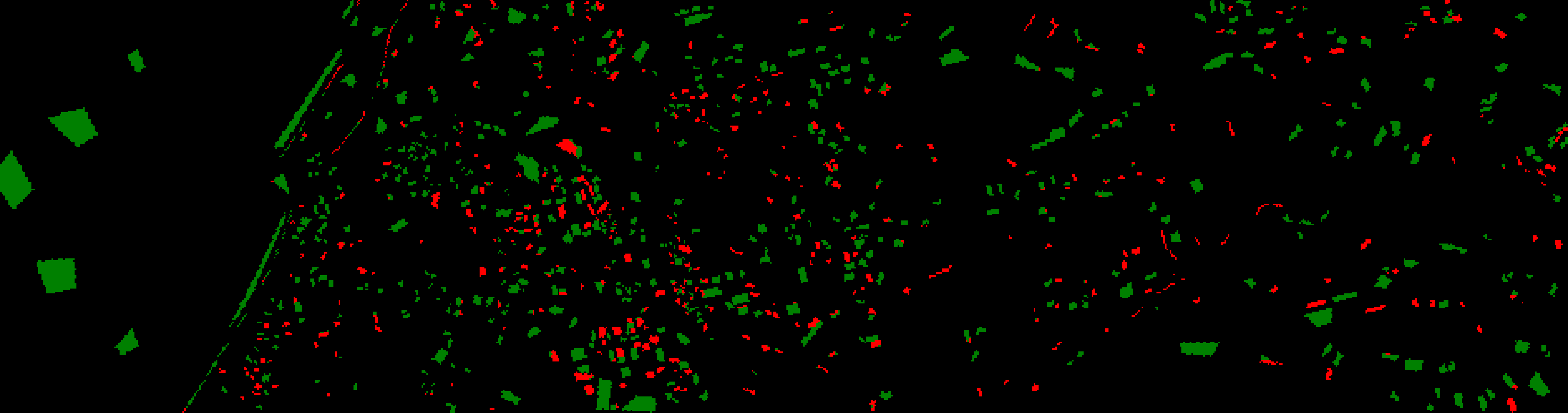}
            \caption{MSST}
        \end{subfigure}
        \begin{subfigure}{0.32\textwidth}
            \includegraphics[width=0.99\textwidth]{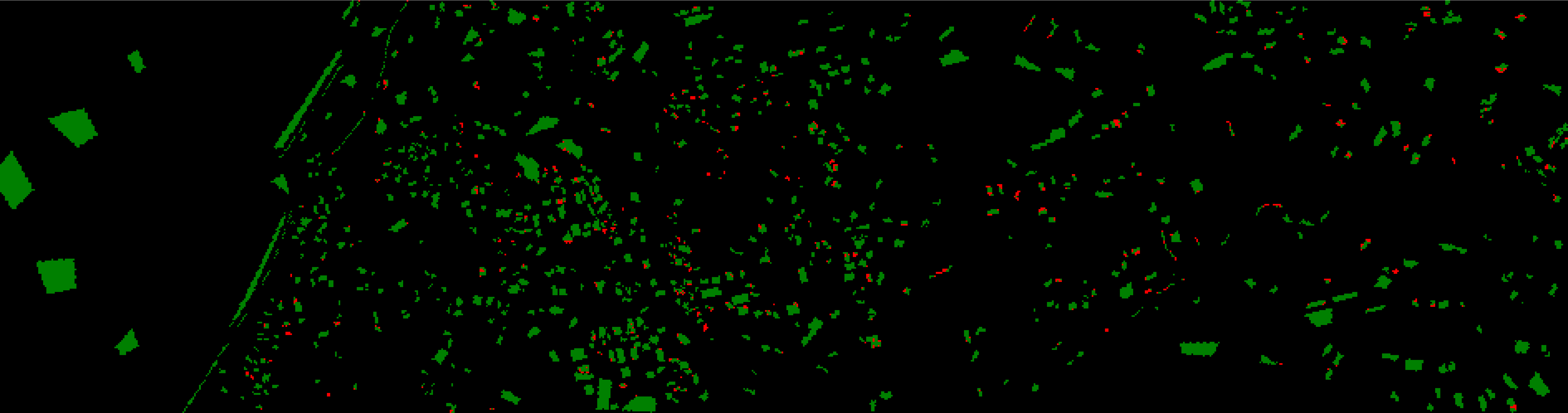}
            \caption{S2CIFT}
        \end{subfigure}
        \begin{subfigure}{0.32\textwidth}
            \includegraphics[width=0.99\textwidth]{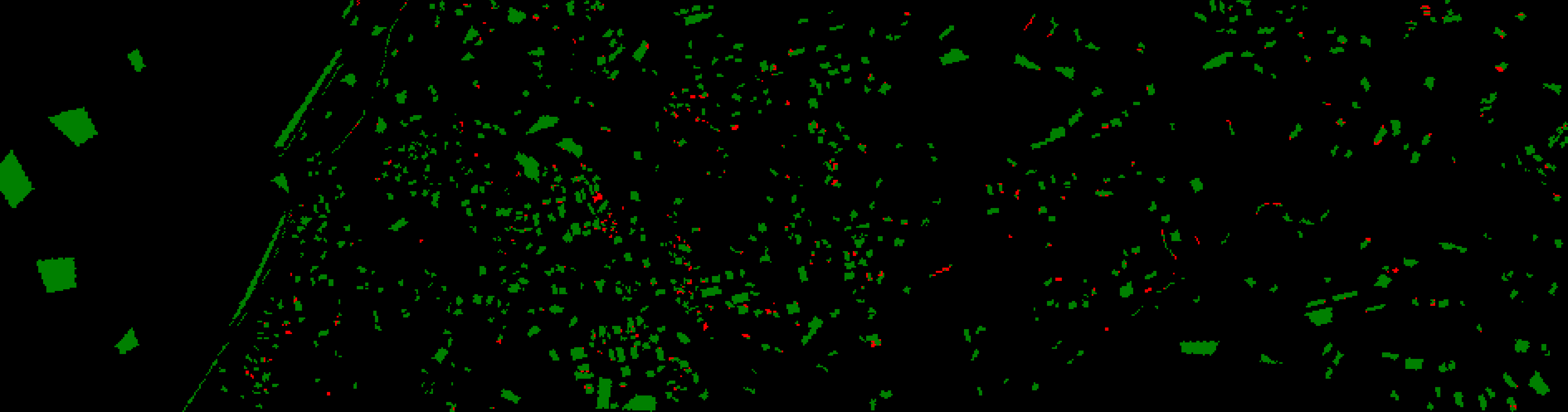}
            \caption{S2CAT}
        \end{subfigure}
    \caption{Qualitative comparison on the Loukia dataset.}
\label{fig:LoukiaComparison}
    \label{LKGT}
\end{figure*}
\begin{figure*}[!htb]
    \centering
        \begin{subfigure}{0.10\textwidth}
            \includegraphics[width=0.99\textwidth]{HH/HH_GT.png}
            \caption{GT}
        \end{subfigure}
        \begin{subfigure}{0.10\textwidth}
            \includegraphics[width=0.99\textwidth]{HH/HH_Error.png}
            \caption{LWM}
        \end{subfigure}
        \begin{subfigure}{0.10\textwidth}
            \includegraphics[width=0.99\textwidth]{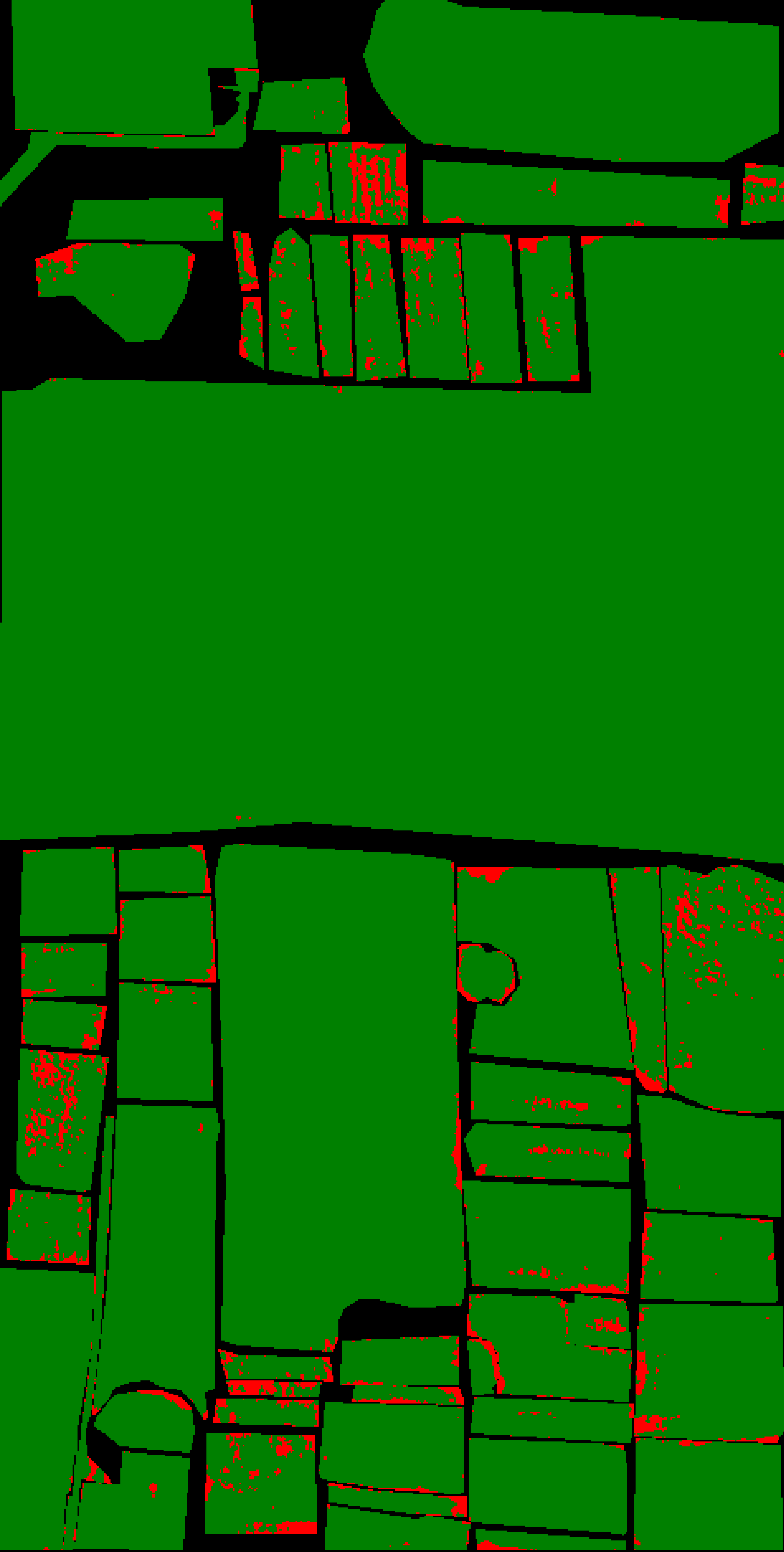}
            \caption{WF}
        \end{subfigure}
        \begin{subfigure}{0.10\textwidth}
            \includegraphics[width=0.99\textwidth]{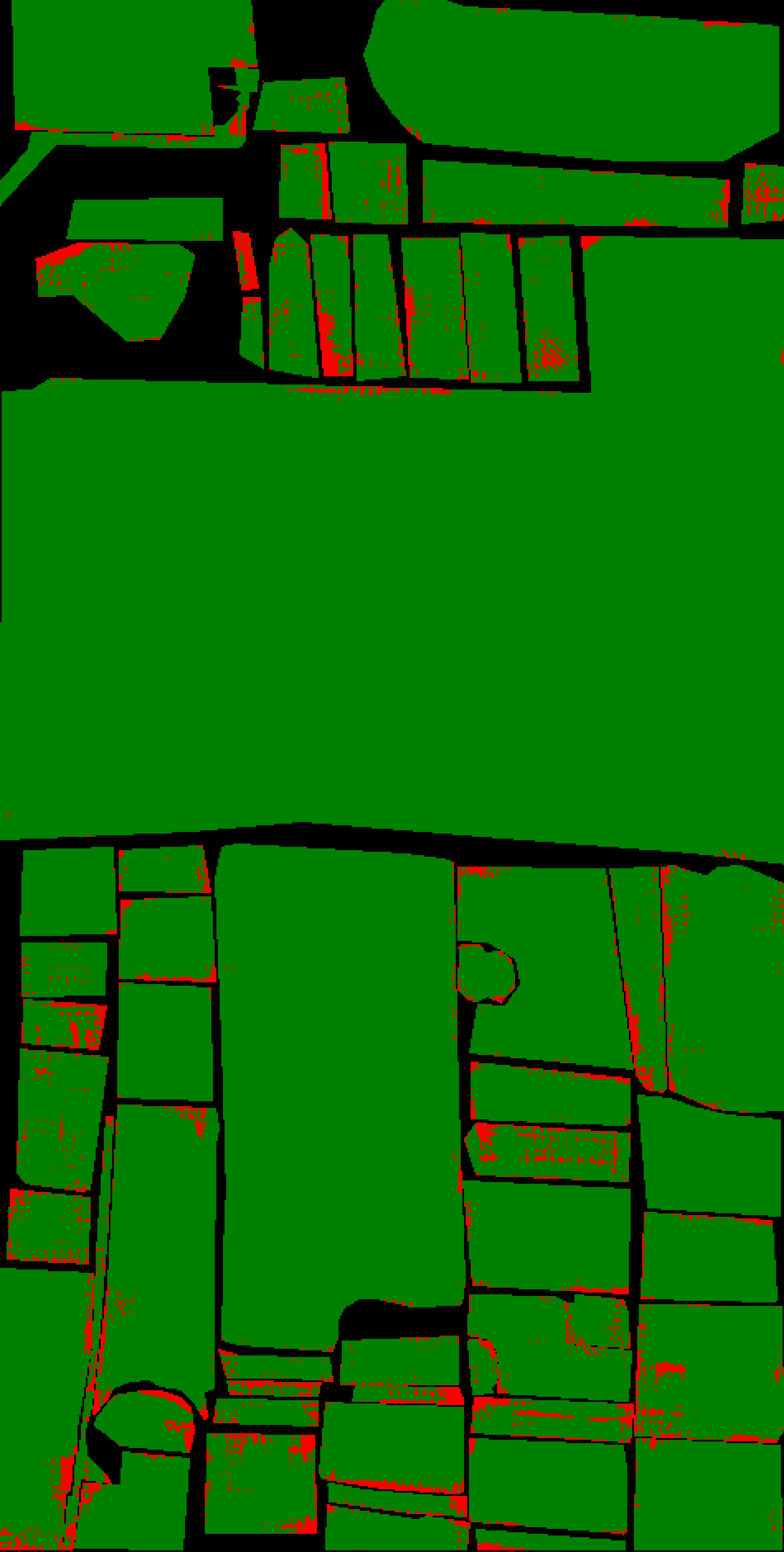}
            \caption{DF}
        \end{subfigure}
        \begin{subfigure}{0.10\textwidth}
            \includegraphics[width=0.99\textwidth]{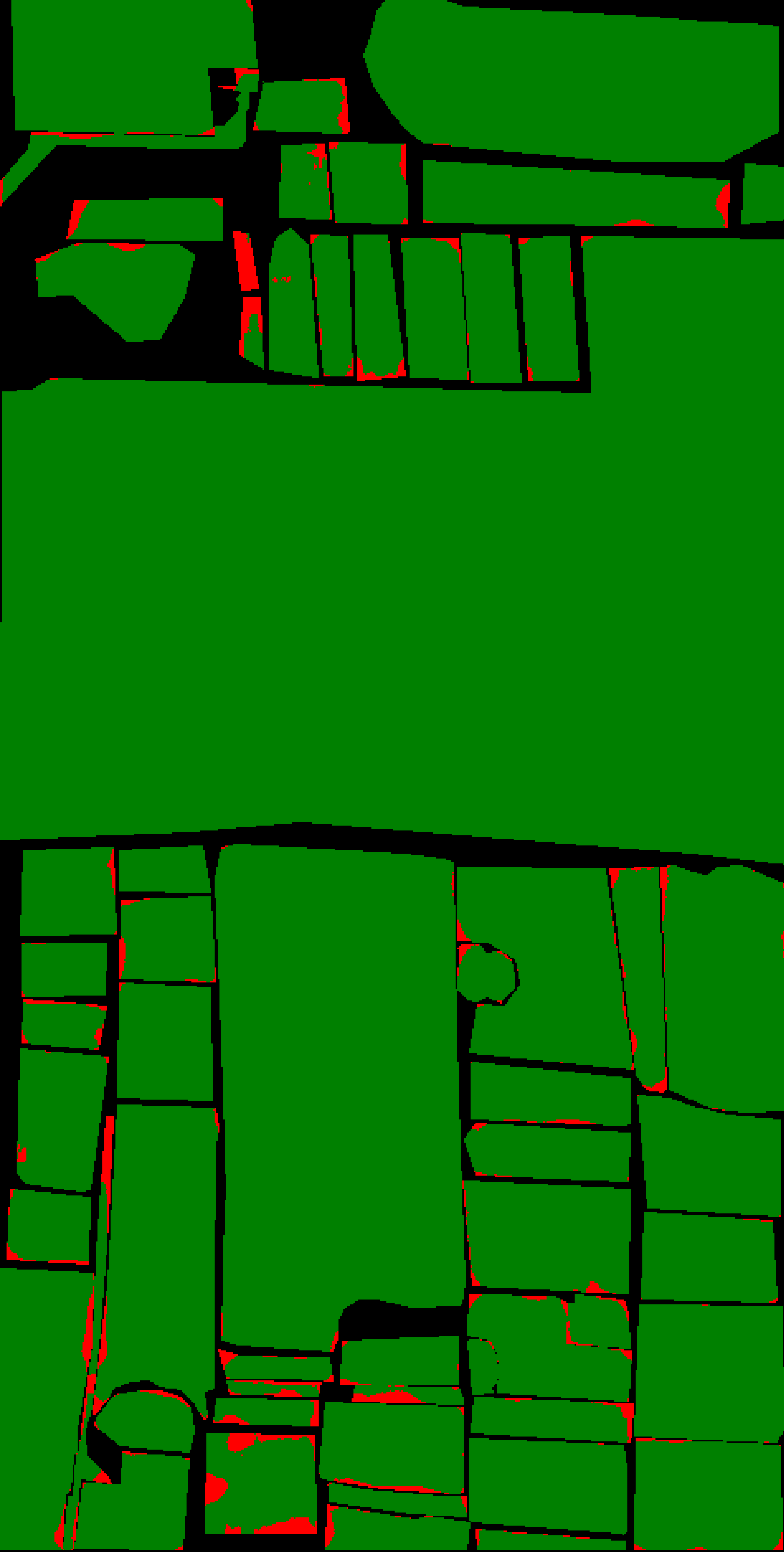}
            \caption{MCMTN}
        \end{subfigure}
        \begin{subfigure}{0.10\textwidth}
            \includegraphics[width=0.99\textwidth]{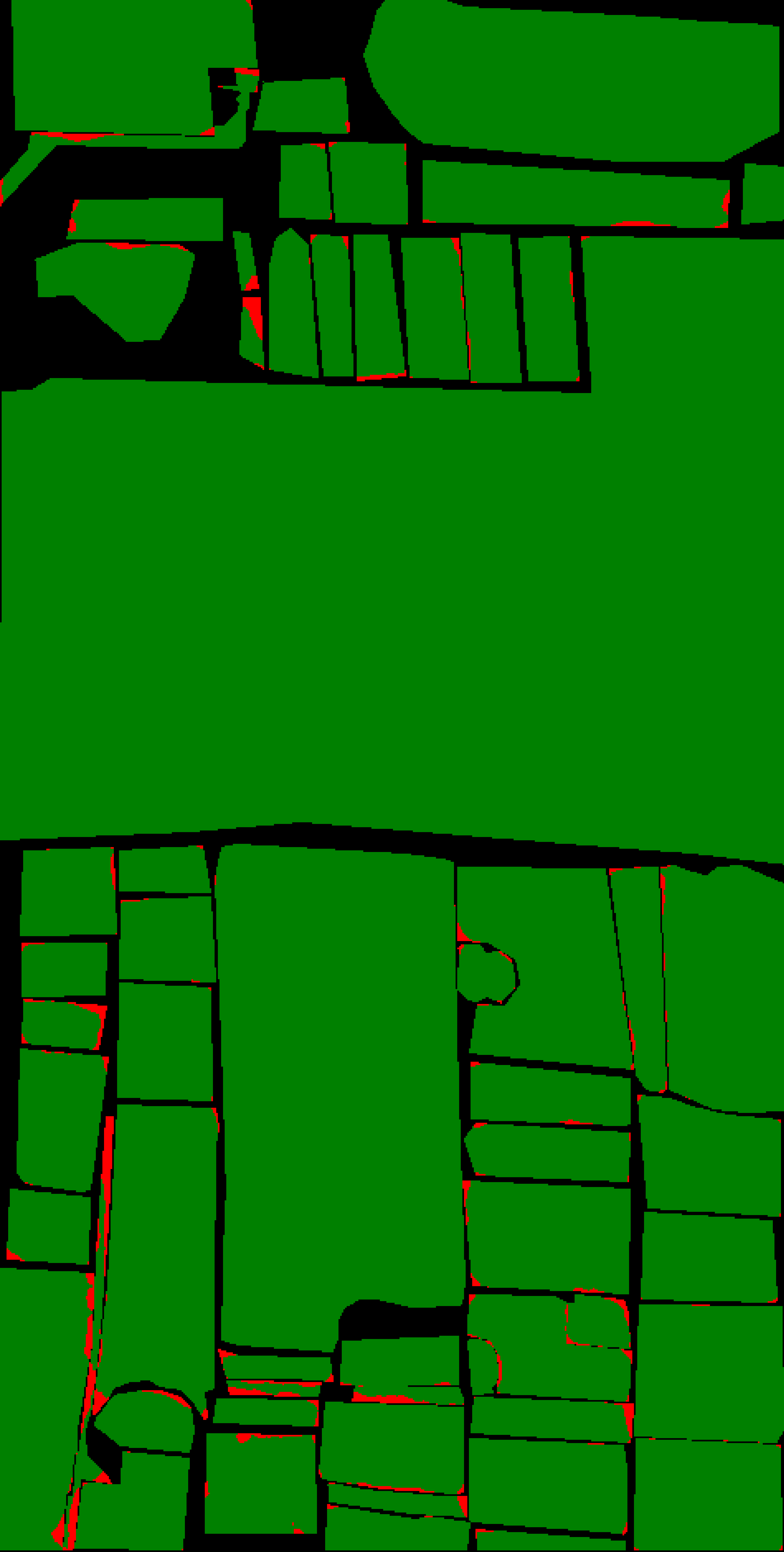}
            \caption{SMFAT}
        \end{subfigure}
        \begin{subfigure}{0.10\textwidth}
            \includegraphics[width=0.99\textwidth]{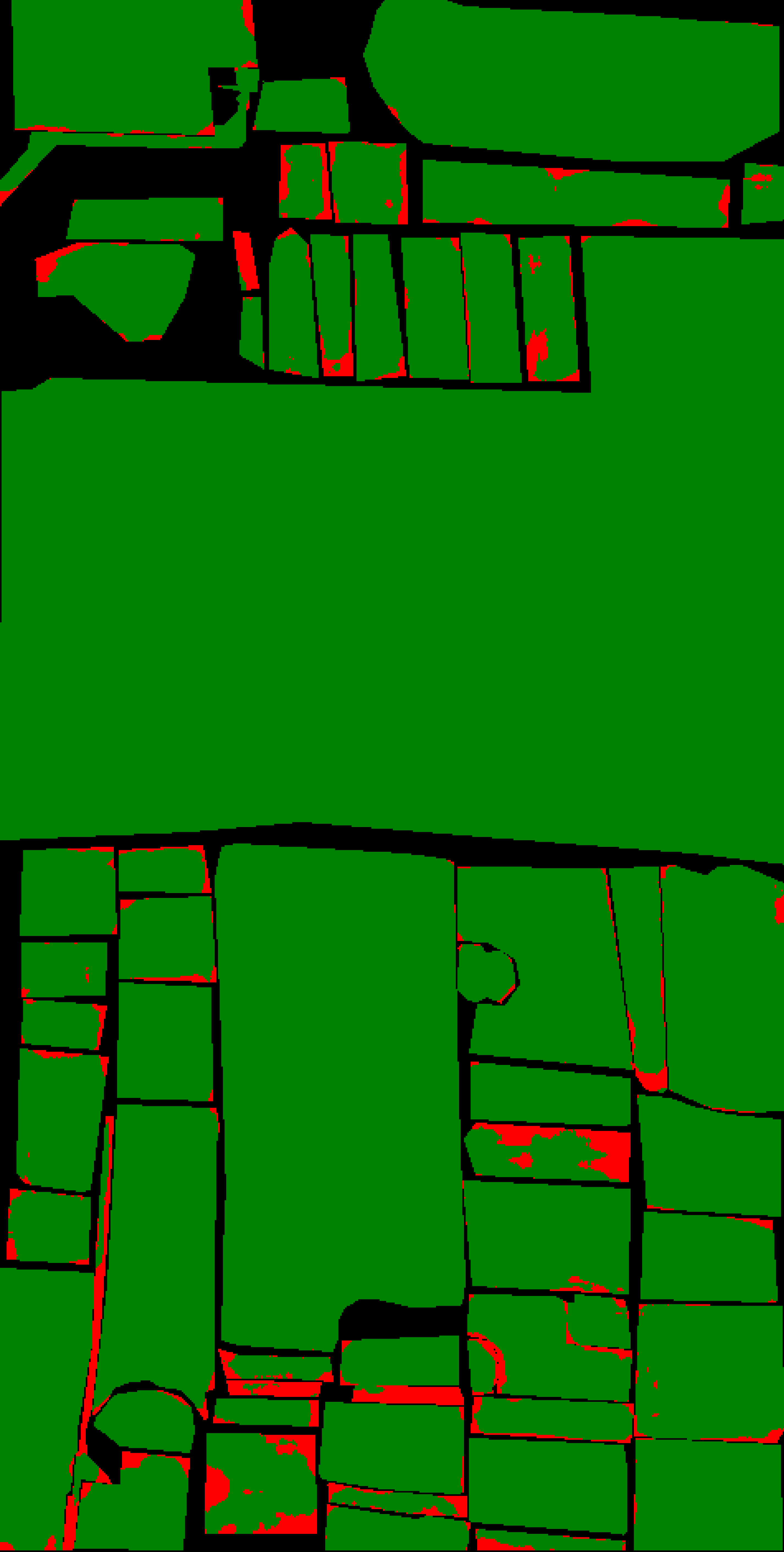}
            \caption{MSST}
        \end{subfigure}
        \begin{subfigure}{0.10\textwidth}
            \includegraphics[width=0.99\textwidth]{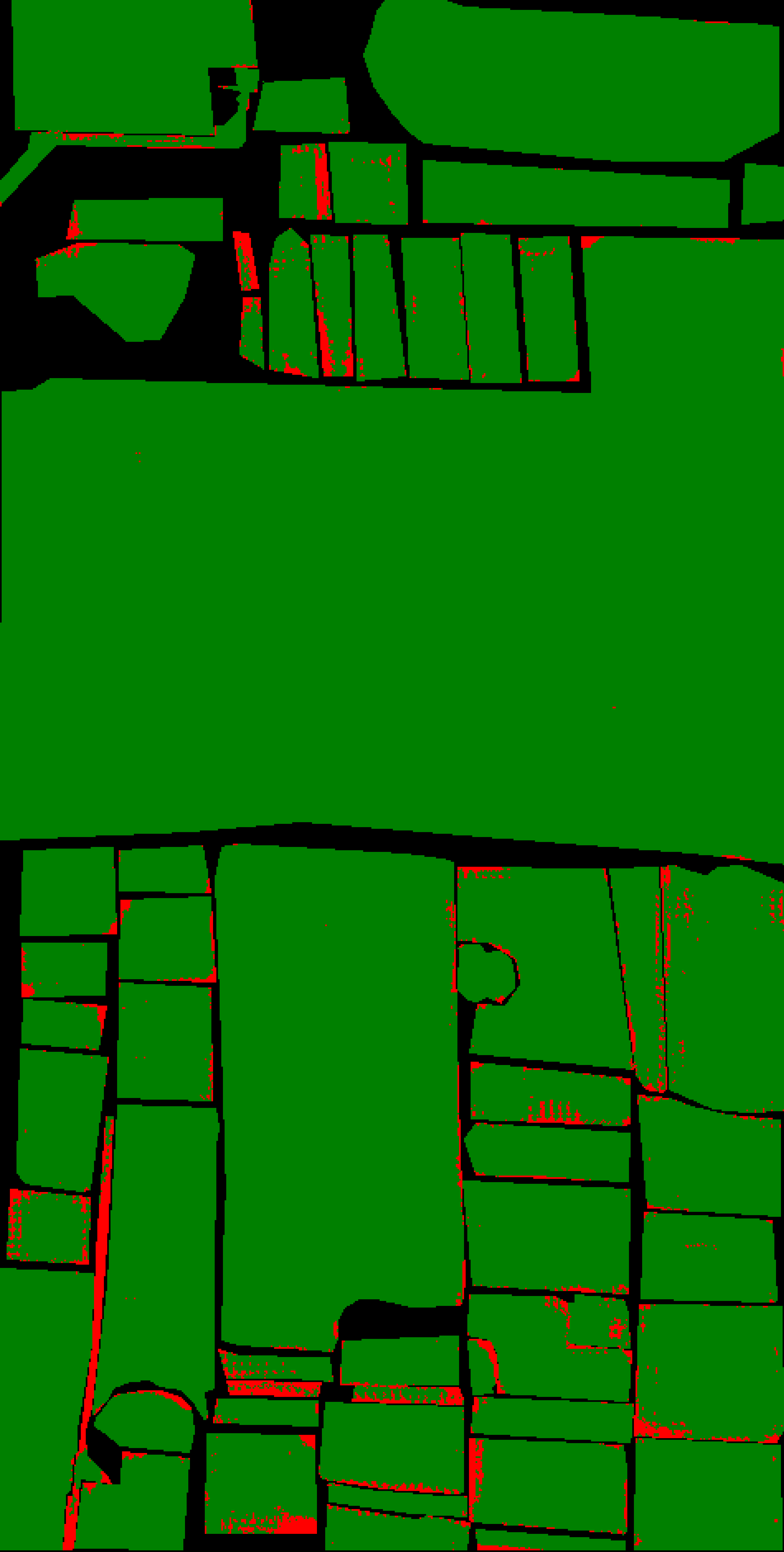}
            \caption{S2CIFT}
        \end{subfigure}
        \begin{subfigure}{0.10\textwidth}
            \includegraphics[width=0.99\textwidth]{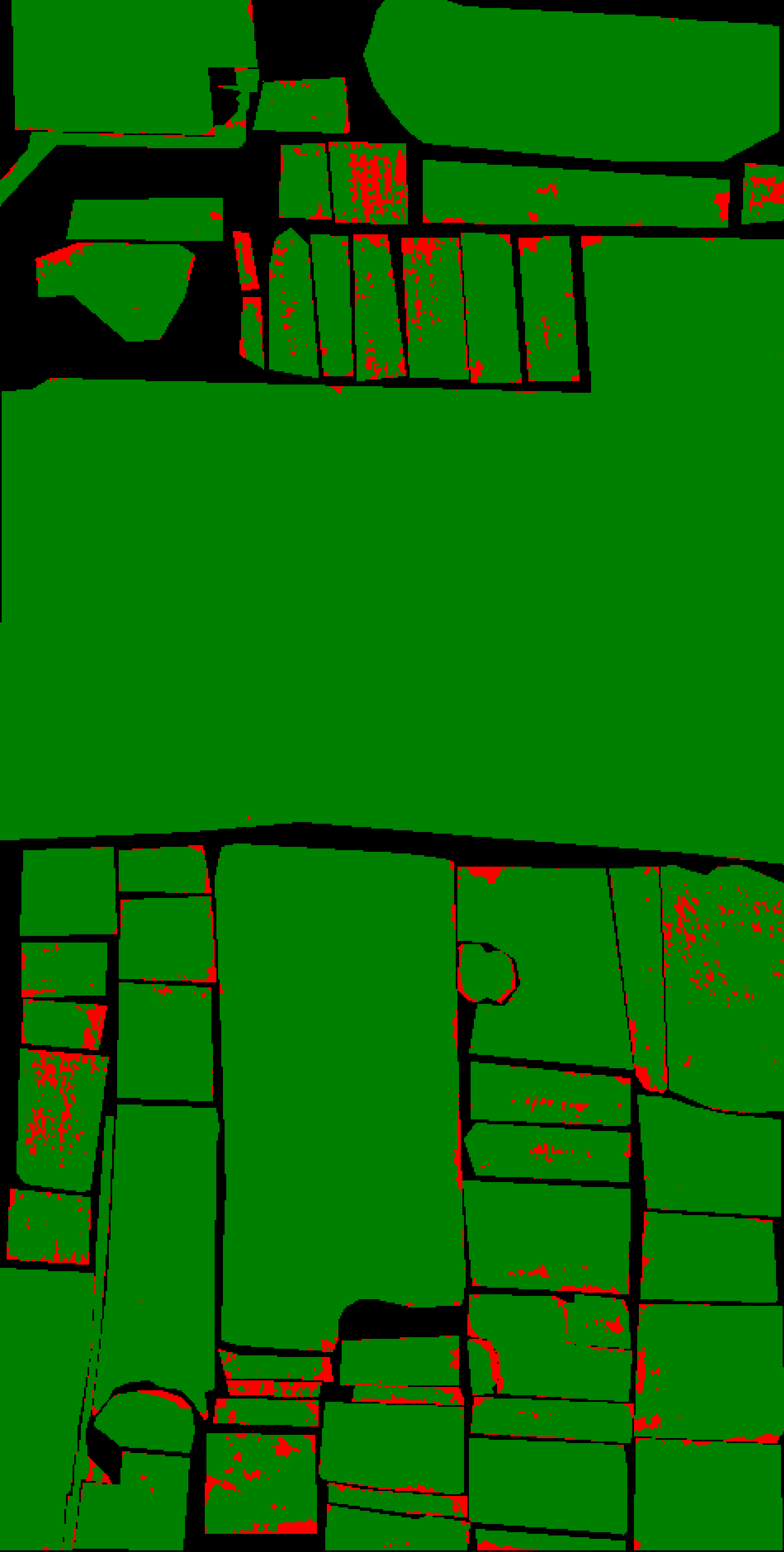}
            \caption{S2CAT}
        \end{subfigure}
    \caption{Qualitative comparison on the HH dataset.}
\label{fig:HHComparison}
    \label{HHGT}
\end{figure*}
\begin{figure*}[!htb]
    \centering
        \begin{subfigure}{0.10\textwidth}
            \includegraphics[width=0.99\textwidth]{TD/TD_GT.png}
            \caption{GT}
        \end{subfigure}
        \begin{subfigure}{0.10\textwidth}
            \includegraphics[width=0.99\textwidth]{TD/TD_Error.png}
            \caption{LWM}
        \end{subfigure}
        \begin{subfigure}{0.10\textwidth}
            \includegraphics[width=0.99\textwidth]{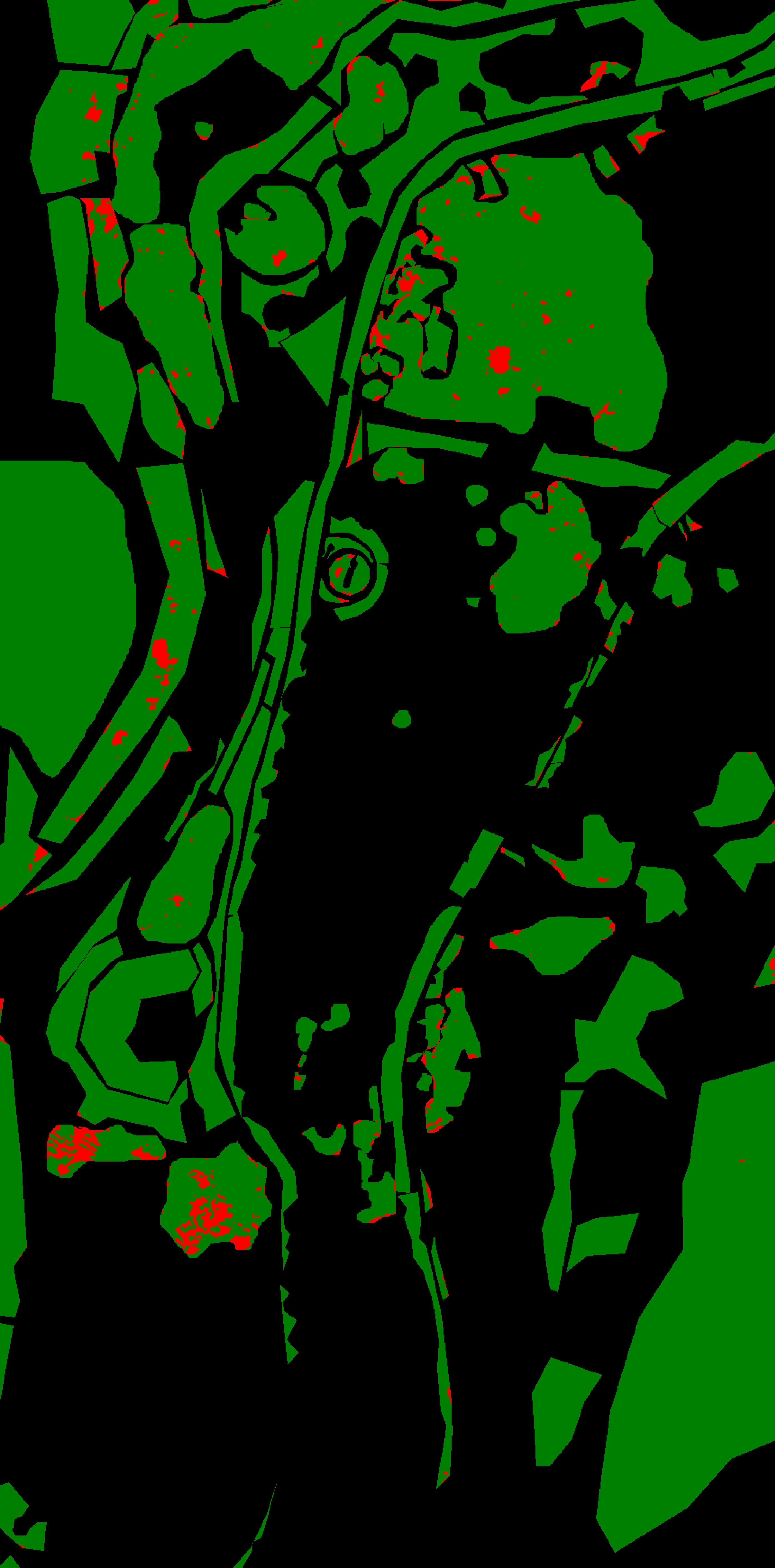}
            \caption{WF}
        \end{subfigure}
        \begin{subfigure}{0.10\textwidth}
            \includegraphics[width=0.99\textwidth]{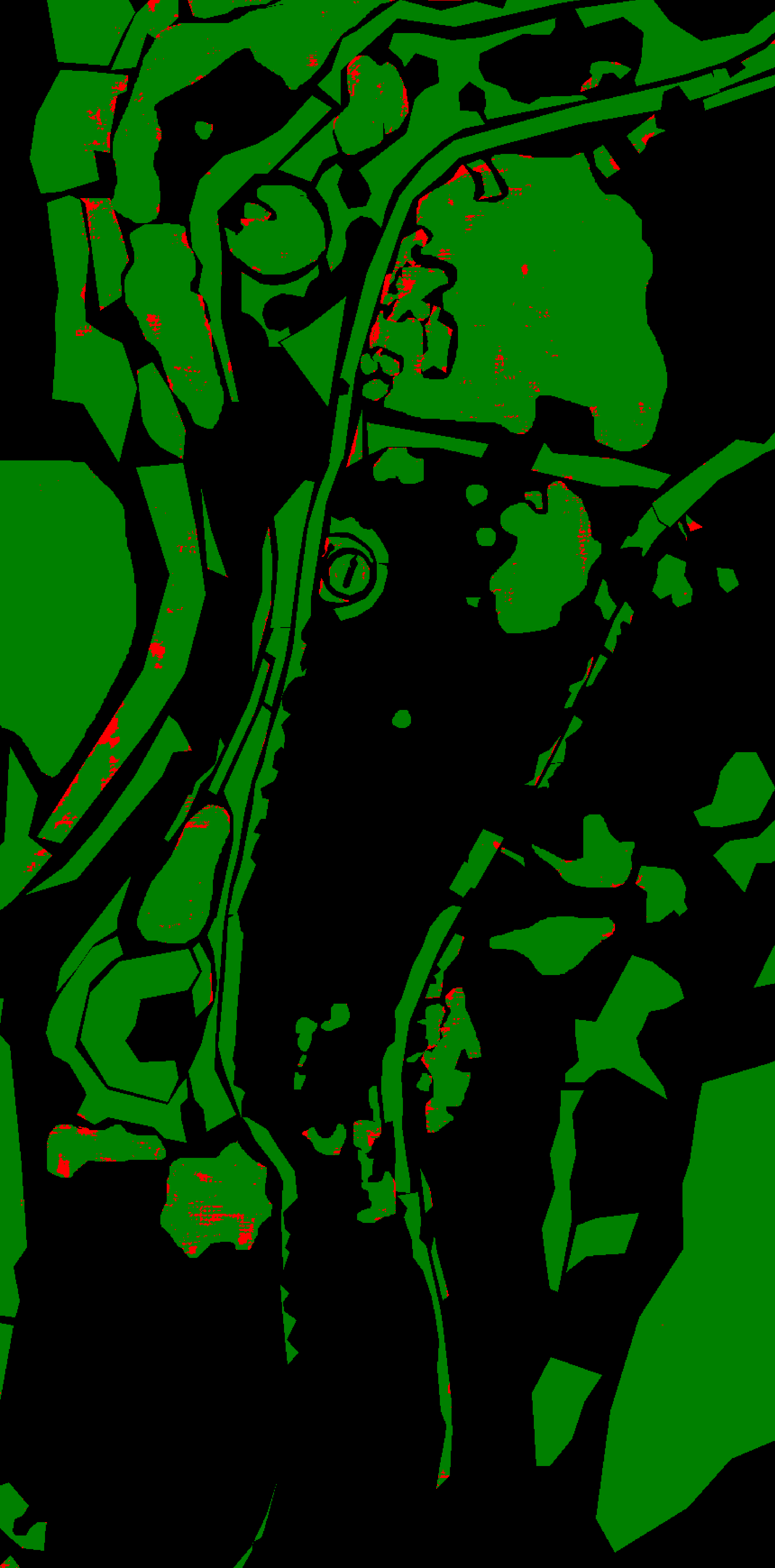}
            \caption{DF}
        \end{subfigure}
        \begin{subfigure}{0.10\textwidth}
            \includegraphics[width=0.99\textwidth]{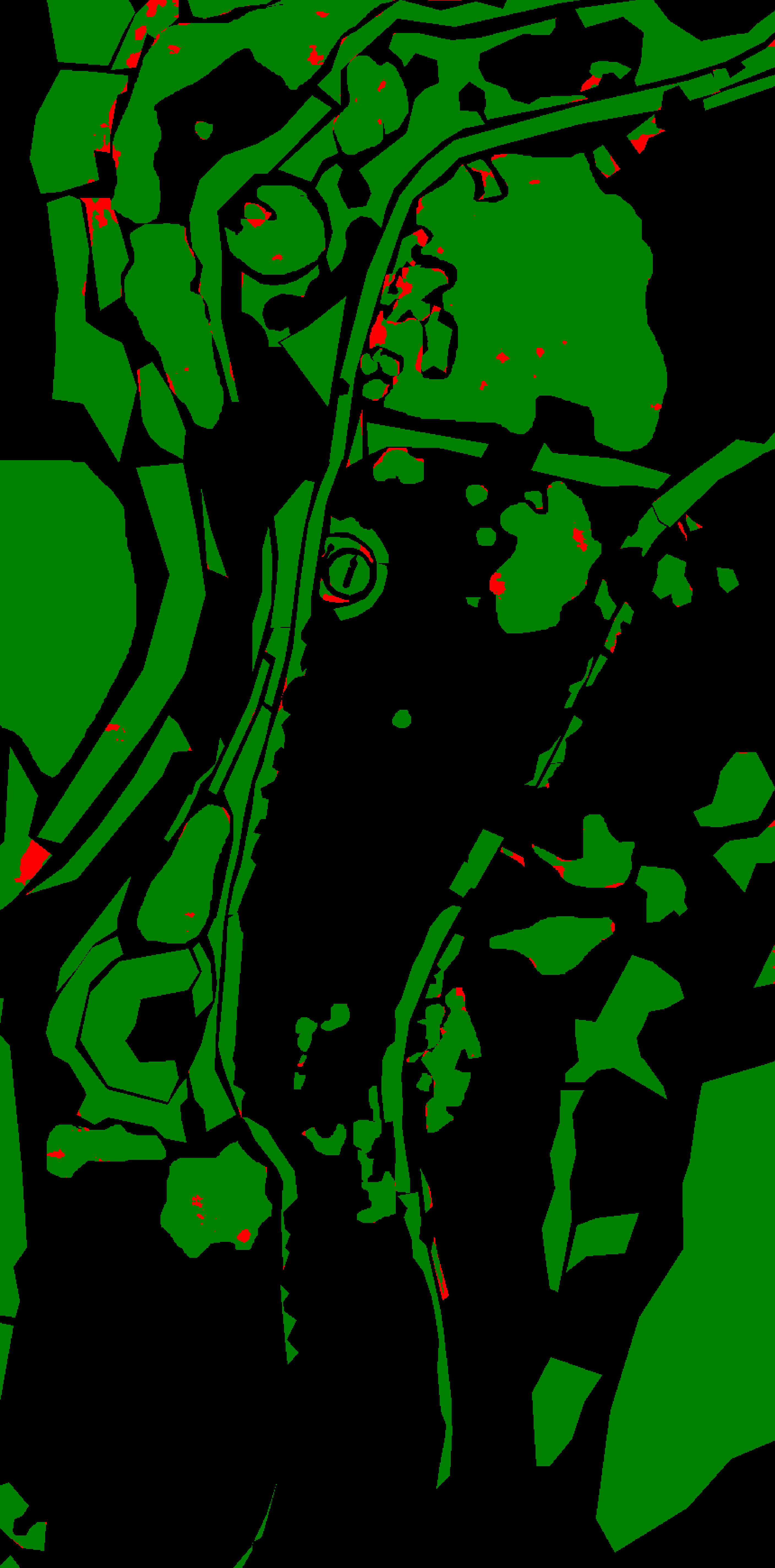}
            \caption{MCMTN}
        \end{subfigure}
        \begin{subfigure}{0.10\textwidth}
            \includegraphics[width=0.99\textwidth]{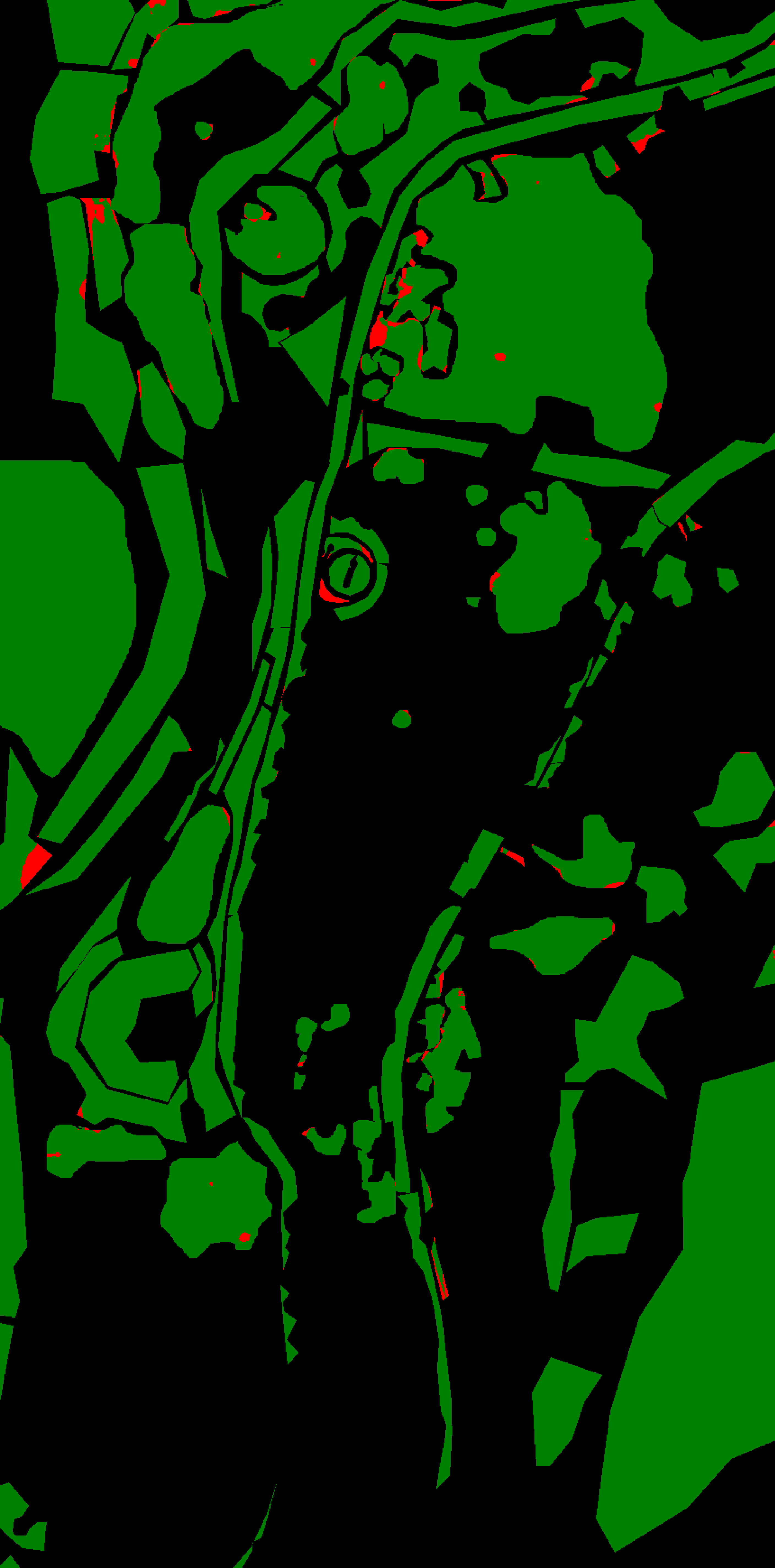}
            \caption{SMFAT}
        \end{subfigure}
        \begin{subfigure}{0.10\textwidth}
            \includegraphics[width=0.99\textwidth]{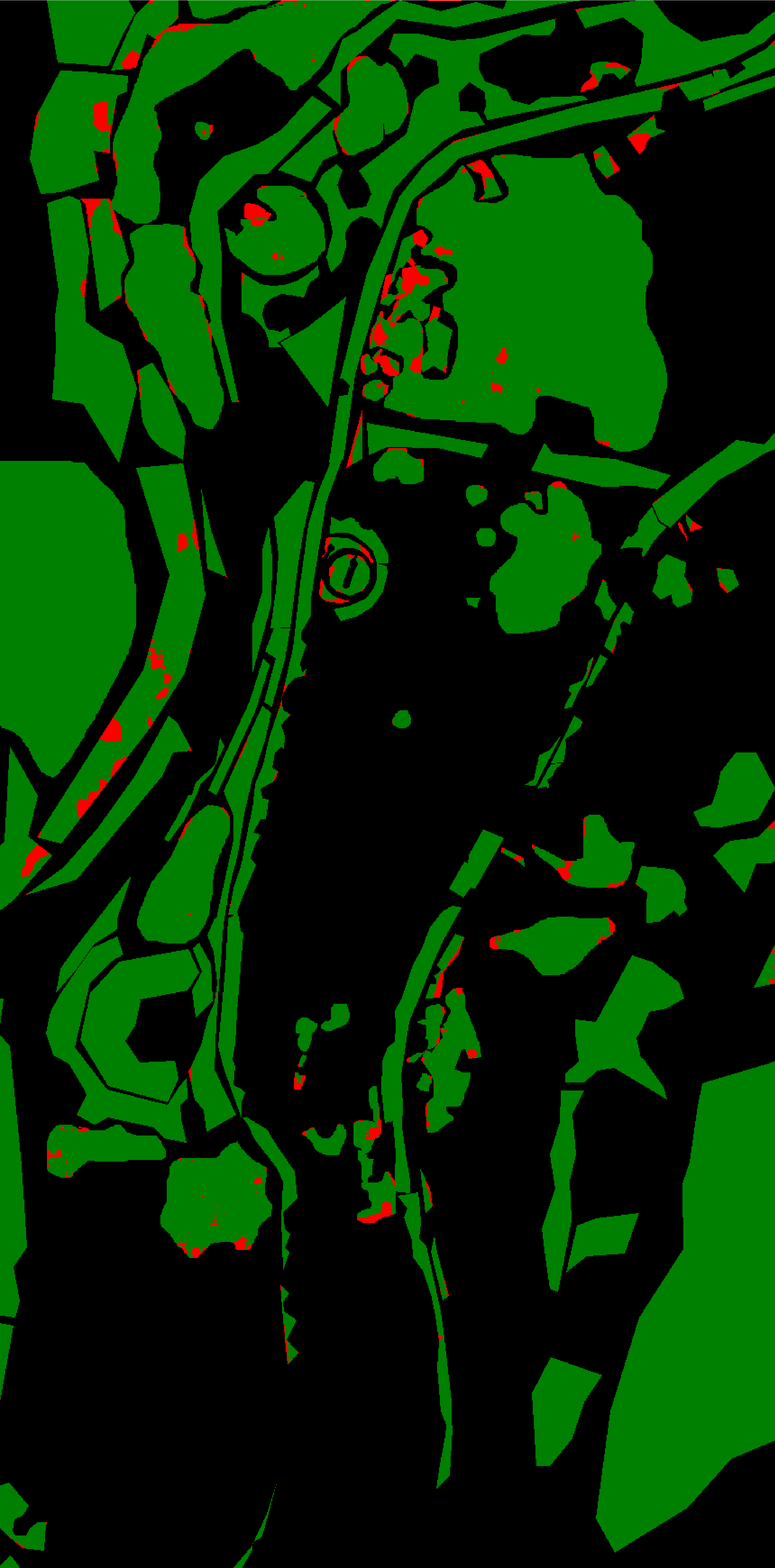}
            \caption{MSST}
        \end{subfigure}
        \begin{subfigure}{0.10\textwidth}
            \includegraphics[width=0.99\textwidth]{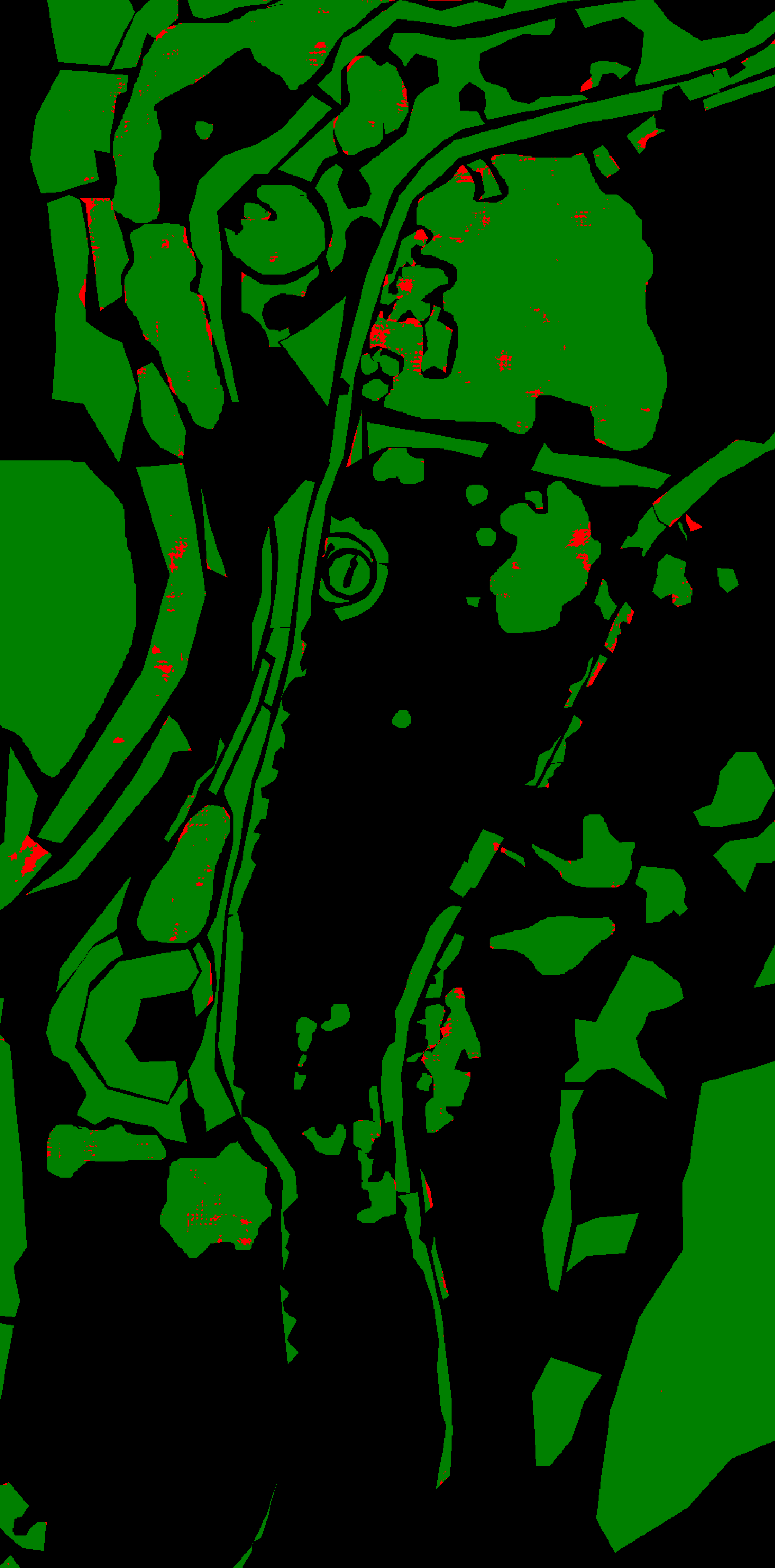}
            \caption{S2CIFT}
        \end{subfigure}
        \begin{subfigure}{0.10\textwidth}
            \includegraphics[width=0.99\textwidth]{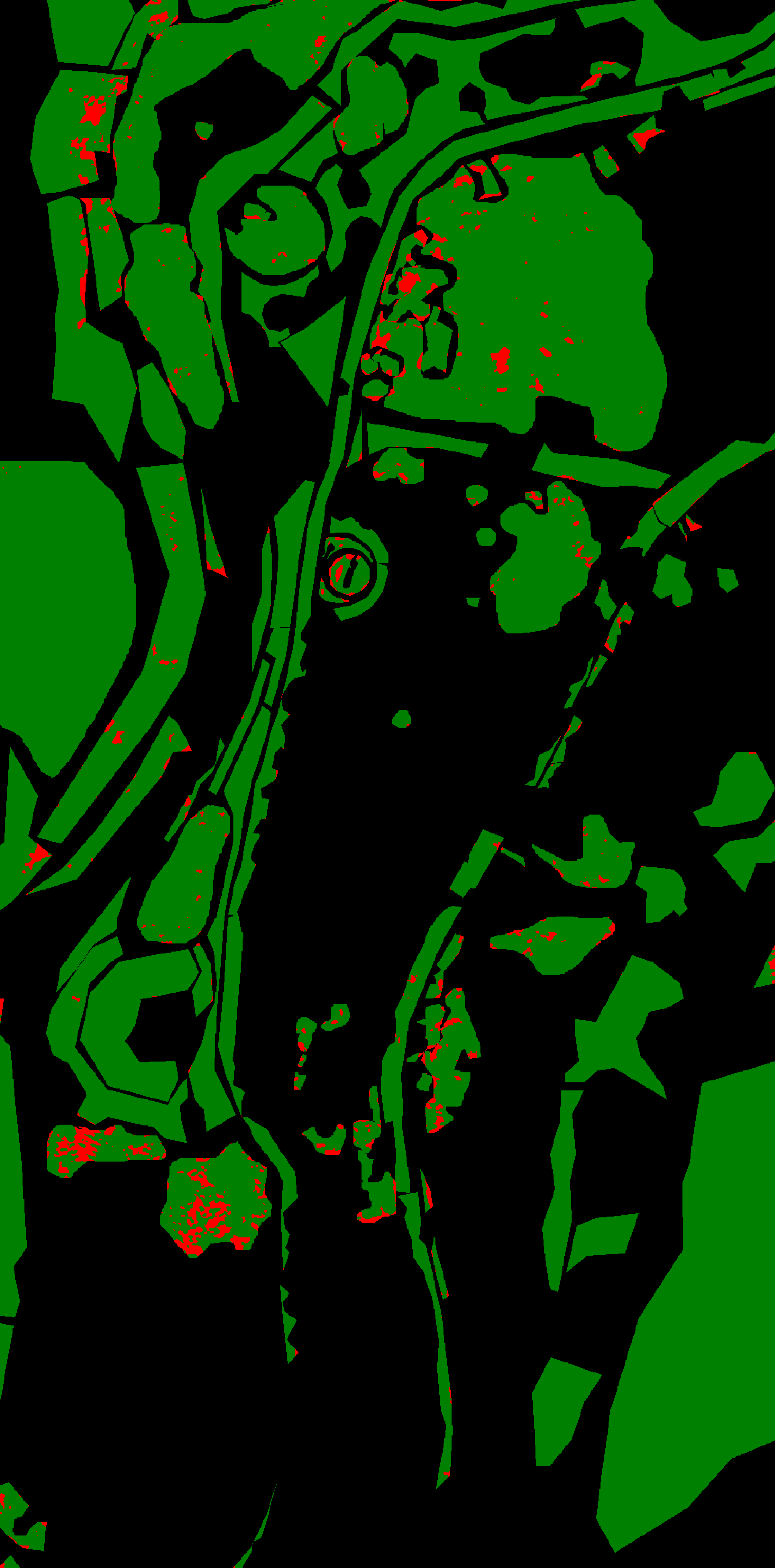}
            \caption{S2CAT}
        \end{subfigure}
    \caption{Qualitative comparison on the TD dataset.}
\label{fig:TDComparison}
    \label{TDGT}
\end{figure*}

\section{Comparison with State-of-the-Art}
\label{sec:comparison}

To evaluate the effectiveness of the proposed LWM framework, we compare it against several recent state-of-the-art transformer-based HSIC methods, including Spectral-Guided Multiscale Feature-Aware Transformer (SMFAT)~\cite{11277353}, WaveFormer (WF)~\cite{10399798}, Differential Spatial-Spectral Transformer (DF)~\cite{10955699}, Mid-Range Convolutional Modulated Transformer Network (MCMTN)~\cite{11219272}, Multiscale spatial-spectral Supertoken Aggregation Transformer (MSST)~\cite{11352814}, Spatial-Spectral Coupling and Interactive Fusion Transformer Network (S2CIFT)~\cite{11386932}, and Spatial-Spectral CNN-Transformer Network (S2CAT)~\cite{11359183}. All methods are evaluated under identical experimental settings. Quantitative comparisons are reported in Table~\ref{tab:Comparitive}, while qualitative error-map comparisons are presented in Figs.~\ref{DFCGT}-\ref{TDGT}.

\noindent\textbf{Quantitative Results Discussion.} The quantitative results in Table~\ref{tab:Comparitive} demonstrate that the proposed LWM framework consistently achieves the strongest overall performance. For the Dioni dataset, which contains substantial intra-class variability and complex spatial structures, LWM achieves an OA of 91.71\%, outperforming the strongest competing method by approximately 0.7-0.8\%. More importantly, the improvement in AA is considerably larger, demonstrating that the proposed framework improves classification performance not only for dominant categories but also for minority classes. A similar trend is observed on the Loukia dataset. Although several competing methods achieve competitive OA values, LWM consistently provides superior OA, AA, and $\kappa$ scores.

For the HH and TD datasets, which already exhibit extremely high baseline accuracies, LWM still achieves the best overall performance. On HH, the proposed framework exceeds all competing methods, demonstrating that latent trajectory reasoning remains beneficial even in near-saturated classification scenarios. On TD, LWM achieves an OA of 99.52\%, indicating exceptional discrimination capability under highly structured coastal and urban environments. The relatively small standard deviations further demonstrate the stability and robustness of the proposed learning framework.

\noindent\textbf{Qualitative Results Discussion.} The qualitative comparisons further support the quantitative findings. Across all evaluated datasets (Figs.~\ref{DFCGT}-\ref{TDGT}), the error maps generated by LWM contain noticeably fewer misclassified pixels than competing methods. The improvement is particularly evident near class boundaries, heterogeneous regions, and spectrally ambiguous areas, where traditional attention-based models frequently produce fragmented predictions. In contrast, LWM generates smoother and more spatially coherent classification maps while preserving fine structural details. Unlike conventional transformer architectures that perform classification using a single static representation, LWM augments representation learning with stochastic latent rollout reasoning. The latent trajectory provides multiple plausible future representations that are aggregated through the rollout planner, enabling the classifier to exploit structured predictive information before making a decision.

\section{Limitations and Failure Analysis}
\label{sec:limitations}

Although the proposed LWM framework demonstrates consistent improvements across multiple hyperspectral benchmarks, it is important to analyze its limitations and identify scenarios where latent rollout reasoning may provide limited benefit. Understanding such failure modes is essential for assessing the practical applicability of the framework and for guiding future research on latent predictive modeling for HSI analysis. In this section, we discuss computational considerations, modeling limitations, and representative failure cases observed during experimentation.

\subsubsection{\textbf{Modeling Limitations}} The proposed framework introduces additional computational complexity compared with conventional transformer-based classifiers. In addition to spectral, spatial, and fusion transformers, LWM performs multi-step latent rollouts and trajectory aggregation. Although the rollout horizon used in this work is relatively short ($T=3$), the computational cost grows approximately linearly with the rollout length. Consequently, very long rollout horizons may provide diminishing returns while increasing memory consumption and inference latency.

Another limitation stems from the absence of explicit physical dynamics. Unlike world models used in temporal environments, the latent transitions learned by LWM do not correspond to true temporal evolution. Instead, they represent structured transformations in representation space. While this formulation is effective for classification, the learned latent trajectories should not be interpreted as physically meaningful spectral or environmental changes. Consequently, the rollout process primarily acts as a predictive regularization mechanism rather than a physical simulation model.

Furthermore, the current implementation assumes a fixed patch size and local spatial context. Extremely large-scale structures or long-range spatial dependencies extending beyond the extracted patch may not be fully represented. Although transformer attention partially alleviates this limitation, global scene-level reasoning remains constrained by the local patch formulation adopted in most HSIC benchmarks.

Finally, the effectiveness of latent rollout reasoning appears to depend on dataset complexity. As observed in the branch-wise analysis, highly saturated datasets such as HH and TD exhibit extremely high agreement between the Base and Trajectory branches. In such cases, the predictive diversity between branches becomes very small, reducing the potential gains obtainable through latent rollout reasoning.

\subsubsection{\textbf{Failure Analysis}} Detailed inspection of classification maps, attention visualizations, and error maps reveals several recurring failure patterns. First, most misclassifications occur near class boundaries where neighboring categories exhibit similar spectral signatures. Such boundary regions often contain mixed pixels whose spectral responses represent combinations of multiple materials. Although latent rollout reasoning improves contextual understanding, the ambiguity inherent in mixed pixels remains challenging.

Second, errors are frequently observed among semantically related classes that possess strong spectral overlap. In agricultural scenes, crops at similar growth stages may produce nearly indistinguishable spectral responses, while in urban environments different materials can exhibit highly correlated reflectance patterns. These cases provide limited discriminative evidence even after latent trajectory aggregation.

Third, the latent rollout mechanism occasionally propagates uncertainty from an incorrect initial representation. Because rollout generation begins from the environment state $\mathbf{z}_0$, severe errors in the initial representation may lead to trajectories that remain consistently biased. This behavior is most likely to occur in highly corrupted regions, rare classes with limited training samples, or samples located near complex class transitions.

The branch correction analysis provides additional insight into these failure modes. Datasets such as Dioni and Loukia exhibit relatively large JSD and positive NG values, indicating that the trajectory branch successfully corrects a substantial number of Base-branch errors. In contrast, HH and TD show extremely high agreement rates and very small divergence values, suggesting that both branches converge to nearly identical predictions. In such scenarios, the latent rollout branch contributes limited corrective information because the underlying classification problem is already nearly solved by the static representation. Despite these limitations, the observed failure cases are largely concentrated in ambiguous boundary regions, spectrally overlapping categories, and highly challenging samples.

\section{Conclusions and Future Research Directions}
\label{sec:conclusion}

This paper introduced Stochastic Latent World Modeling (LWM), a trajectory-aware representation learning framework for hyperspectral image classification (HSIC). Unlike conventional approaches that rely on static spatial-spectral representations, the proposed framework models structured latent evolution through stochastic rollout reasoning in representation space. By constructing an environment state from fused spatial-spectral transformer features and propagating it through a stochastic latent transition process, LWM enables the classifier to exploit predictive latent structure beyond the information available in a single forward-pass representation.

Extensive experiments conducted on diverse hyperspectral benchmarks demonstrated the effectiveness of the proposed approach. The results showed that LWM consistently achieves superior classification performance compared with recent state-of-the-art methods while also improving calibration quality, robustness to spectral corruption and noise. Branch-wise analysis further confirmed that latent rollout reasoning contributes complementary predictive information beyond static transformer features. In addition, ablation studies verified the importance of each component of the framework.

Several research directions remain open for further investigation, as already discussed in Section \ref{sec:limitations}. First, the current rollout mechanism operates at a single spatial scale; extending LWM to hierarchical and multi-scale representations may further improve the modeling of complex scene structures. Second, incorporating adaptive rollout horizons and uncertainty-aware planning strategies could allow the model to dynamically adjust its latent reasoning process according to scene complexity. Third, integrating LWM with semi-supervised, self-supervised, and foundation-model paradigms may improve scalability under extremely limited supervision. Finally, extending the proposed framework to hyperspectral change detection, multimodal remote sensing fusion, domain adaptation, and spatiotemporal Earth observation analysis represents an exciting direction for future research.

\bibliographystyle{IEEEtran}
\bibliography{Sam}
\end{document}